\documentclass[acmsmall]{acmart}

\usepackage[mathletters]{ucs}
\usepackage[utf8x]{inputenc}
\usepackage[T1]{fontenc}    % use 8-bit T1 fonts
\usepackage[normalem]{ulem}
\usepackage{hyperref}       % hyperlinks
\usepackage{booktabs}       % professional-quality tables
\usepackage{colortbl} % Include the package
\usepackage{tabularx}
\usepackage{multicol}
\usepackage{multirow}
\usepackage{xcolor}
\usepackage{environ}
\usepackage{graphicx}
\usepackage[most]{tcolorbox}
\usepackage{doi}
\usepackage[framemethod=TikZ]{mdframed}
\usepackage[export]{adjustbox} % Required for valign option
\usepackage{pifont}		    % provides the checkmark and other symbols
\usepackage{calc} % Enables arithmetic operations with lengths
\definecolor{blackcolor}{RGB}{0, 0, 0} 
\definecolor{lightbluecolor}{RGB}{218, 232, 252} 
\definecolor{lightpurplecolor}{RGB}{225, 213, 231} 
\definecolor{lightredcolor}{RGB}{248, 206, 204} 
\definecolor{lightyellowcolor}{RGB}{255, 242, 204} 
\definecolor{lightorangecolor}{RGB}{255, 230, 204} 
\definecolor{lightgreencolor}{RGB}{214, 232, 213} 
\definecolor{lightgraycolor}{RGB}{247, 247, 247} 
\definecolor{darkgreencolor}{RGB}{0, 128, 0} 
\definecolor{linkblue}{RGB}{0, 0, 200} % Azul semelhante ao padrão de links

\hypersetup{
    colorlinks = true,
    urlcolor = linkblue
}

\newcommand{\myhref}[2][]{%
  \if\relax\detokenize{#1}\relax
    \href{https://github.com/nabormendonca/gpt-4v-enade-cs-2021/blob/main/questions/q#2-eng.md}{\textcolor{linkblue}{\underline{#2}}}%
  \else
    \href{https://github.com/nabormendonca/gpt-4v-enade-cs-2021/blob/main/questions/q#2-eng.md}{\textcolor{linkblue}{\underline{#1}}}%
  \fi
}

\newcommand{\figbox}[1]{
	\tcbox[left=2mm, right=2mm, boxsep=0mm, top=2mm, bottom=2mm, toptitle=0mm, bottomtitle=0mm, nobeforeafter, colback=white]{
		#1
	}
}

\newcommand{\boxfontsize}{\fontsize{8}{9}\selectfont\sffamily}

\newcommand{\imagewidth}{.65\linewidth}

\tcbset{onecolumnstyle/.style={
    before upper={
		\setlength{\parskip}{-4pt}
	},
	breakable, % allows the box to break over multiple pages
}}
\tcbset{twocolumnstyle/.style={
    before upper={
		\setlength{\parskip}{-2pt}
		\begin{multicols}{2}
	},
	after upper={\end{multicols}}
}}
\tcbset{myboxstyle/.style={
	fontupper=\boxfontsize,
	boxrule=1pt,
	colback=lightgraycolor,
	colframe=blackcolor,
	coltitle=blackcolor,
	fonttitle=\bfseries\boxfontsize,
	colbacktitle=lightgreencolor,
	enhanced,
	attach boxed title to top left={xshift=2mm, yshift=-2mm},
	left=2mm,
	right=2mm,
	top=2mm,
	bottom=2mm
}}

\newtcolorbox{questionbox}[1][]{
	myboxstyle,
	onecolumnstyle,
	colbacktitle=lightgreencolor,
	#1 % allows for the specification of additional options
}
\newtcolorbox{modelinputbox}[1][]{
	myboxstyle,
	onecolumnstyle,
	colbacktitle=lightbluecolor,
	#1 % allows for specification of additional options
}
\newtcolorbox{modeloutputbox}[1][]{
	myboxstyle,
	onecolumnstyle,
	colbacktitle=lightpurplecolor,
	#1 % allows for specification of additional options
}
\newtcolorbox{expertbox}[1][]{
	myboxstyle,
	onecolumnstyle,
	colbacktitle=lightorangecolor,
	#1 % allows for specification of additional options
}

\def\BibTeX{{\rm B\kern-.05em{\sc i\kern-.025em b}\kern-.08emT\kern-.1667em\lower.7ex\hbox{E}\kern-.125emX}}

\AtBeginDocument{%
  \providecommand\BibTeX{{%
    Bib\TeX}}}

\acmJournal{TOCE}

\begin{document}

\title{Evaluating ChatGPT-4 Vision on Brazil's National Undergraduate Computer Science Exam}

\author{Nabor C.~Mendon\c{c}a}
\email{nabor@unifor.br}
\orcid{0000-0002-2972-9423}
\affiliation{%
  \institution{Post-Graduate Program in Applied Informatics, University of Fortaleza}
  \streetaddress{Av. Washington Soares, 1321, Edson Queiroz}
  \city{Fortaleza}
  \state{Cear\'{a}}
  \country{Brazil}
  \postcode{60811-905}
}


\begin{abstract}
  The recent integration of visual capabilities into Large Language Models (LLMs) has the potential to play a pivotal role in science and technology education, where visual elements such as diagrams, charts, and tables are commonly used to improve the learning experience. This study investigates the performance of ChatGPT-4 Vision, OpenAI's most advanced visual model at the time the study was conducted, on the Bachelor in Computer Science section of Brazil's 2021 National Undergraduate Exam (ENADE). By presenting the model with the exam's open and multiple-choice questions in their original image format and allowing for reassessment in response to differing answer keys, we were able to evaluate the model's reasoning and self-reflecting capabilities in a large-scale academic assessment involving textual and visual content. ChatGPT-4 Vision significantly outperformed the average exam participant, positioning itself within the top 10 best score percentile. While it excelled in questions that incorporated visual elements, it also encountered challenges with question interpretation, logical reasoning, and visual acuity. A positive correlation between the model's performance in multiple-choice questions and the performance distribution of the human participants suggests multimodal LLMs can provide a useful tool for question testing and refinement. 
  However, the involvement of an independent expert panel to review cases of disagreement between the model and the answer key revealed some poorly constructed questions containing vague or ambiguous statements, calling attention to the critical need for improved question design in future exams. Our findings suggest that while ChatGPT-4 Vision shows promise in multimodal academic evaluations, human oversight remains crucial for verifying the model's accuracy and ensuring the fairness of high-stakes educational exams. 
  The paper's research materials are publicly available at \url{https://github.com/nabormendonca/gpt-4v-enade-cs-2021}.
\end{abstract}

\begin{CCSXML}
  <ccs2012>
    <concept>
      <concept_id>10010405.10010489.10010495</concept_id>
      <concept_desc>Applied computing~E-learning</concept_desc>
      <concept_significance>500</concept_significance>
    </concept>
    <concept>
      <concept_id>10010147.10010178.10010179</concept_id>
      <concept_desc>Computing methodologies~Natural language processing</concept_desc>
      <concept_significance>500</concept_significance>
    </concept>
    <concept>
      <concept_id>10010147.10010178.10010224</concept_id>
      <concept_desc>Computing methodologies~Computer vision</concept_desc>
      <concept_significance>500</concept_significance>
    </concept>
  </ccs2012>
\end{CCSXML}
  
\ccsdesc[500]{Applied computing~E-learning}
\ccsdesc[500]{Computing methodologies~Natural language processing}
\ccsdesc[500]{Computing methodologies~Computer vision}

\keywords{Multimodal Generative AI, ChatGPT-4 Vision, Educational Assessment, Computer Science Education}


\maketitle

\section{Introduction}\label{sec:introduction}

The advent of large language models (LLMs), such as OpenAI's ChatGPT~\citep{chatgpt} and Google's Gemini~\citep{gemini}, has ushered in a new era in natural language processing. These models exhibit human-like performance in tasks requiring complex reasoning, including abstract thinking, problem-solving across diverse knowledge domains, coding, and comprehending human emotions and motives~\citep{arxiv2023surveyLLMs,bubeck2023sparks}. The recent integration of visual capabilities into these models \citep{openai2023chatgpt} has the potential to play a pivotal role in advancing their use in educational settings, particularly in the science and technology fields, where visual elements such as diagrams, charts, and tables are commonly used to offer a more comprehensive and engaging learning experience~\citep{evagorou2015role}. 

Using LLMs like ChatGPT and Gemini in education presents novel opportunities and challenges for students, educators, and standardized exam creators \citep{li2023adapting,kasneci2023chatgpt}. These models offer students innovative tools for exam preparation, potentially reshaping study habits and learning methodologies \citep{dai2023can}. For educators, LLMs serve as a resource to enrich teaching strategies and help tailor exam content to better suit learning outcomes \citep{zhou2024the}. Meanwhile, standardized exam creators could utilize these AI tools to refine their questions, ensuring clarity, fairness, and appropriate difficulty levels. However, the use of these models also introduces concerns around accuracy, reliability, and potential for misuse, highlighting the need for careful evaluation of visual LLMs in academic settings \citep{meyer2023chatgpt,zhou2024the}. 

This study evaluates the accuracy of ChatGPT-4 Vision~\citep{openai2023gpt4v}, OpenAI's most advanced visual model at the time the study was conducted, in answering the Bachelor in Computer Science (BCS) section of Brazil's 2021 National Undergraduate Exam (ENADE) \citep{enade}. Unlike most prior evaluations of ChatGPT and other LLMs in academic assessments in Computer Science (e.g., \citep{joshi2023chatgpt,richards2024bob,jalil2023chatgpt,bordt2023chatgpt}) and other fields (e.g., \citep{nori2023capabilities,frenkel2023chatgpt,pursnani2023performance}), we presented all ENADE BCS exam questions to ChatGPT-4 Vision in their original image formats as seen by human participants. This method allowed a comprehensive evaluation of the model's multimodal reasoning abilities, challenging its linguistic and visual interpretation capabilities in a particular education domain. 

Additionally, we explored the model's capacity for self-reflection by prompting it to reassess its initial responses when presented with differing answer keys. When ChatGPT-4 Vision's answers diverged from the answer keys after reassessment, we assembled an independent panel of computer science researchers from various Brazilian universities to evaluate the accuracy of the model's responses. This additional layer of expert review provided insights into the model's reasoning accuracy and the exam questions' quality regarding clarity and objectivity.

We further correlated the model's performance with the difficulty and discrimination levels of the multiple-choice questions, as reported by ENADE organizers. This analysis aimed to provide insights into how the model's proficiency compared to that of the human participants, offering a more nuanced understanding of the model's capabilities and limitations in a real-world academic assessment context.

Our study aims to answer the following quantitative research questions:

\begin{description}
	\item[RQ1:] How well does ChatGPT-4 Vision perform in answering ENADE 2021 BCS open questions?
	\item[RQ2:] How well does ChatGPT-4 Vision perform in answering ENADE 2021 BCS multiple-choice questions?
	\item[RQ3:] How often did ChatGPT-4 change or maintain its response when confronted with a disagreeing official answer?
	\item[RQ4:] What is ChatGPT-4 Vision's ENADE 2021 BCS exam final score, and how does it compare with the exam participants?
\end{description}

In addition, we examined the reasoning, accuracy, and reflective analysis of ChatGPT-4 Vision, and the independent expert assessments to answer the following qualitative research question:

\begin{description}
	\item[RQ5:] What are the main challenges ChatGPT-4 Vision faced in answering ENADE 2021 BCS open and multiple-choice questions?
\end{description}

In summary, our contributions are as follows:

\begin{itemize}
	\item We have evaluated ChatGPT-4 Vision's multimodal reasoning abilities on the ENADE 2021 BCS exam, a large-scale computer science academic assessment test involving textual and visual content, providing all questions in image format.
	\item We have explored ChatGPT-4 Vision's capacity for self-reflection by allowing it to reassess its initial responses when confronted with differing answer keys.
	\item We have correlated ChatGPT-4 Vision's performance with the difficulty and discrimination levels of the exam's multiple-choice questions, offering insights into how the model's proficiency compares to that of the human participants.
	\item We have involved an independent panel of computer science researchers from various Brazilian universities to evaluate the accuracy of ChatGPT-4 Vision's responses that conflicted with the official answer, providing further insights into the model's limitations and the exam questions' design quality. 
\end{itemize}

Overall, our results reveal ChatGPT-4 Vision's impressive yet limited performance in reasoning about multimodal content in computer science academic assessments. In particular, ChatGPT-4 Vision significantly outperformed the average exam participant, achieving the top 10 best score percentile. However, the model faced several challenges with question interpretation, logical reasoning, and visual acuity. These findings underscore the continued need for human oversight in verifying LLMs's accuracy in solving complex reasoning tasks. Moreover, some disagreements between the model, the expert panel, and the official answer raise concerns about the quality of some exam questions. Given the significant impact of ENADE on Brazil's undergraduate academic rankings, our findings call attention to the critical need for improved question design and quality control in future ENADE exams to ensure a fair educational assessment process.

To facilitate the replication and extension of our study, we have made our research materials publicly available at \url{https://github.com/nabormendonca/gpt-4v-enade-cs-2021}. This repository contains the exam questions, response standards, answer keys, ChatGPT-4's full conversations, and experts' evaluations, in Portuguese and English.

The remainder of the paper is organized as follows. Section~\ref{sec:related-work} discusses related work in LLM evaluations in logical reasoning tasks and educational assessments in computer science and other domains. Section~\ref{sec:backgroung} provides background information on ENADE and the 2021 BCS exam. Section~\ref{sec:method} details the methodology used in our study, including our evaluation procedure and experimental setup. Section~\ref{sec:results} presents the results of our study, and Section~\ref{sec:discussion} discusses its implications, insights, and limitations. Finally, Section~\ref{sec:conclusion} concludes the paper with a summary of the main contributions and suggestions for future research.

\section{Related Work}
\label{sec:related-work}

Our work touches on several areas of related research, including LLM evaluations in general logical reasoning tasks \citep{bubeck2023sparks,liu2023evaluating,mao2024gpteval,espejel2023gpt}, using LLMs in educational assessments in computer science \citep{joshi2024chatgpt,richards2024bob,jalil2023chatgpt,bordt2023chatgpt,mahon2023no} and other domains \citep{nori2023capabilities,toyama2023performance,frenkel2023chatgpt,pursnani2023performance}, and recent advances in Vision-Language Models (VLMs) \citep{zhang2024far,mayne2024-improvingGPT,wu2024visualization,zhang2024mathverse}. Also of interest are related works that evaluated LLMs' effectiveness in answering Brazil's ENEM \citep{nunes2023evaluating,pires2023evaluating,santos2023benchmarking-cbic}---a national multidisciplinary high-school-level exam widely used as an admission test in Brazilian universities---and ENADE \citep{almeida2024sabia}. We briefly discuss these studies below to provide a comprehensive backdrop against which the contributions of our work can be contextualized and compared.

\subsection{LLMs Evaluation in Logical Reasoning}

\citet{bubeck2023sparks} showed that GPT-4 exhibited significant logical reasoning capabilities in various domains, including professional and academic benchmarks, where it performs at the human level. The study also highlighted the model's struggles with planning and creative problem-solving due to its reliance on a sequential prediction paradigm. Similarly, \citet{liu2023evaluating} focused on logical reasoning in reading comprehension and natural language inference, finding that despite strong performances on known datasets, GPT-4's effectiveness diminishes with new, out-of-distribution data. This observation points to a crucial gap in GPT-4's capacity to handle unfamiliar data types and logical inferences beyond its training data. 
\citet{mao2024gpteval} provided a more holistic view by surveying assessments across various capabilities of ChatGPT and GPT-4, indicating strong language understanding and generation skills but also limitations in tasks requiring multi-step reasoning and domain-specific knowledge. 
Finally, \citet{espejel2023gpt} compared reasoning abilities across LLMs including GPT-3.5, GPT-4, and BARD, finding that while GPT-4 generally outperforms its predecessors and competitors, all models show limitations in inductive, mathematical, and multi-step reasoning tasks. 

Building on these foundational studies, our study further explores the capabilities of ChatGPT-4 Vision by evaluating its performance on the ENADE 2021 BCS exam, which includes complex reasoning tasks in a multimodal context. This extends the current understanding of LLMs' capabilities by testing them on more challenging, real-world academic assessments.

\subsection{LLMs Evaluation in Computer Science Assessments}

\citet{joshi2024chatgpt} evaluated the performance of ChatGPT-3.5 across a variety of computer science exercises, including true/false, multiple-choice, short and long answers, and coding-related questions. The study demonstrated ChatGPT-3.5's inconsistent accuracy across subjects and question types, particularly in numerical-based problems. Similarly, \citet{richards2024bob} provide an in-depth analysis of ChatGPT-3.5's performance in a university computer science curriculum. The study involved a blind marking exercise in which the ChatGPT-generated answers were directly compared with those of human students. The findings showed that barring two outliers, every ChatGPT-3.5 answer achieved at least a passing grade. The study also highlights that while ChatGPT-3.5 is capable of producing adequate solutions in most cases, its performance varies across subjects and question types, with ChatGPT-3.5's effectiveness diminishing as the complexity of the questions increases. 
\citet{jalil2023chatgpt} explored ChatGPT-3.5's performance in answering 31 questions from a popular software testing textbook. The study found that ChatGPT-3.5 provided correct or partially correct answers and explanations for over half of the questions tested, showcasing the model's potential as an educational tool. However, the study also noted ChatGPT's inconsistent accuracy and poor self-assessment, especially in questions that involved both coding and conceptual understanding.
\citet{bordt2023chatgpt} investigated ChatGPT-3.5 and 4's performance in an undergraduate computer science exam focusing on algorithms and data structures. ChatGPT-3.5 narrowly passed the exam with 20.5 out of 40 points, a result that was slightly below the performance of the average student. ChatGPT-4 scored about 17\% higher, reaching the average student's level. The study underlined the model's challenges with exercises requiring a detailed understanding of algorithmic processes, particularly involving structured output that was not in pseudo-code format. 
Finally, \citet{mahon2023no} investigated the capabilities of ChatGPT-4 (no vision) on the UK A-Level and Irish Leaving Certificate computer science examinations. They found that ChatGPT-4 achieved high marks on both exams but struggled with questions involving symbols or images and cascading questions where initial inaccuracies led to further errors. This work underscores the strengths and weaknesses of generative AI in educational assessments and has implications for exam design and marking schemes.

Our research builds on these findings by not only evaluating ChatGPT-4 Vision on a comprehensive academic assessment in Computer Science but also examining its performance relative to question difficulty and discrimination levels. This provides a more nuanced understanding of the model's capabilities and limitations compared to the human participants, which is extremely important in educational contexts.

\subsection{LLMs Evaluation in Other Education Domains}

\citet{nori2023capabilities} presents a comprehensive evaluation of ChatGPT-4's performance on the United States Medical Licensing Examination (USMLE) and the MultiMedQA medical benchmark. The study found that GPT-4, without any specialized prompt crafting, surpassed the passing score on the USMLE by over 20 points and outperformed previous models like ChatGPT-3.5 and specialized medical models. The study also explored ChatGPT-4's calibration, crucial in high-stakes domains like medicine, and found it to be significantly better than its predecessor. Similarly, \citet{toyama2023performance} investigated the performance of ChatGPT-3.5, ChatGPT-4, and Bard on the official board examination of the Japan Radiology Society. The study found that ChatGPT-4 significantly outperformed ChatGPT-3.5 and Bard by over 20 percentage points, achieving an accuracy of 65\% in the exam's multiple-choice questions. Both studies underscore the potential utility of AI models in assisting professionals in high-stakes domains like medicine, while also emphasizing the need for cautious application due to the models' limitations in accuracy and safety-critical tasks.

In the field of engineering, \citet{frenkel2023chatgpt}  examined ChatGPT-3.5 and 4's performance on the United States Fundamentals of Engineering (FE) Mechanical Engineering Exam. The study found that ChatGPT-4 significantly outperforms ChatGPT-3.5, achieving 76\% correct answers compared to 51\% by GPT-3.5. A similar study was conducted by \citet{pursnani2023performance} where they evaluated GPT-3.5-Legacy, GPT-3.5-Turbo, and GPT-4's performance across different sections of the FE Environmental Exam. The study also found that GPT-4 outperformed GPT-3.5-Legacy and GPT-3.5-Turbo by a significant margin, achieving 75\% accuracy with slightly refined prompts. However, both studies confirmed previous findings about ChatGPT's error types and pitfalls, noting its inconsistency across question types and subjects and its tendency to confidently produce incorrect answers. 

Our study contributes to this body of work by evaluating ChatGPT-4 Vision's performance on a high-stakes academic exam (ENADE), involving both textual and visual content. By allowing the model to reassess its responses with differing answer keys, we explore its self-reflective capabilities, a novel aspect not covered in previous education-centered studies.
	
\subsection{Recent Advances in VLMs}

\citet{zhang2024far} conducted a comprehensive evaluation of state-of-the-art VLMs, including GPT-4 Vision, on tasks requiring visual deductive reasoning using Raven's Progressive Matrices (RPMs). Their findings revealed significant challenges in models' abilities to perform multi-hop relational and deductive reasoning based solely on visual cues, highlighting perception issues and difficulties in understanding abstract patterns. \citet{mayne2024-improvingGPT} proposed improvements to GPT-4 Vision's performance on visual reasoning tasks through enhanced prompt engineering techniques. By incorporating specific instructions in prompts, such as paying attention to details and describing items with positional specificity, Mayne demonstrated substantial improvements in the model's accuracy on visual puzzles. \citet{wu2024visualization} introduced the Visualization-of-Thought prompting technique, which aims to elicit spatial reasoning in LLMs by visualizing their reasoning traces. This technique enhances spatial reasoning tasks, such as natural language navigation, visual navigation, and visual tiling, by guiding subsequent reasoning steps through mental image visualization. Finally, \citet{zhang2024mathverse} presented {\sc MathVerse}, a specialized benchmark designed to evaluate the visual mathematical reasoning skills of Multi-modal Large Language Models (MLLMs). {\sc MathVerse} includes a diverse set of 2,612 visual math problems across three primary areas: plane geometry, solid geometry, and functions. The study found that most existing MLLMs struggle to understand math diagrams and often rely more on textual questions than visual inputs, emphasizing the need for better visual encoding capabilities.

Our work differs from these related studies in several ways but also stands to benefit significantly from their insights and methodologies. While \citet{zhang2024far} provide a comprehensive evaluation of VLMs on visual deductive reasoning tasks, our focus is on assessing ChatGPT-4 Vision's performance in the context of the ENADE exam, which involves a broader range of multimodal reasoning tasks beyond visual deductive reasoning. The visual-enhanced prompt engineering techniques proposed by \citet{mayne2024-improvingGPT} and \citet{wu2024visualization} could be particularly beneficial in improving ChatGPT-4 Vision's handling of spatial reasoning tasks within the ENADE exam, guiding the model through intermediate steps via mental image visualization. Furthermore, our work could benefit from the development of a {\sc MathVerse}-like fine-grained multimodal benchmark based on ENADE questions with visual content to better evaluate and understand ChatGPT-4 Vision's capabilities in real-world educational assessments. 

\subsection{LLMs Evaluation in Brazil's ENEM and ENADE}

\citet{nunes2023evaluating} studied GPT-3.5 and GPT-4's performance on answering multiple-choice ENEM questions, using different prompt strategies, including Chain-of-Thought (CoT). GPT-4, especially with CoT, achieved up to 87\% accuracy on the 2022 exam, significantly outperforming GPT-3.5. 
\citet{pires2023evaluating} also examined the performance of ChatGPT-3.5 and 4, the latter with and without Vision capabilities, on the 2022 and 2023 editions of ENEM. The study showed a clear advancement in performance from ChatGPT-3.5 to ChatGPT-4, with GPT-4 models significantly outperforming their predecessors. Specifically, ChatGPT-4 (no Vision) and ChatGPT-4 Vision achieved accuracies of about 73\% and 79\% in the 2022 exam, and 80\% and 86\% in the 2023 exam, respectively. Interestingly, when provided with human-generated captions transcribing the questions' visual content, ChatGPT-4 with no Vision capabilities improved its accuracies to 81\% and 91\% in the 2022 and 2023 exams, respectively, outperforming ChatGPT-4 Vision with images and no captions.
\citet{santos2023benchmarking-cbic} investigated the performance of quantized open-source LLaMA-based models on a custom-built benchmark containing over one thousand ENEM questions, showing that the best models achieved an accuracy of 46\% for the original texts of the Portuguese questions and 49\% on their English translations. Despite the modest computational requirements of these open-source models, allowing them to run on low-end hardware, their limited accuracy compared to that of closed models indicates significant room for improvement. 
Finally, in a recent study, \citet{almeida2024sabia} introduced Sabi{\'a}-2, a new family of LLMs specifically trained on Portuguese texts and evaluated its performance on a diverse range of exams, including a multidisciplinary ENADE benchmark containing only multiple-choice questions with no images from the 2022 and 2023 exams. The study revealed that Sabi{\'a}-2, particularly the Sabi{\'a}-2 Medium model, matched or surpassed GPT-4 Turbo's performance in several ENADE disciplines and significantly outperformed GPT-3.5 in all but one discipline. The model's lowest scores were achieved in domains related to engineering and economics, with a particularly lower score compared to GPT-4 Turbo in Computer Engineering.

Our work extends this area of research by evaluating ChatGPT-4 Vision's performance on the ENADE BCS exam, which includes both open and multiple-choice multimodal questions. Additionally, by involving an independent panel of computer science researchers, we provide a more rigorous assessment of the model's accuracy and limitations in answering ENADE questions, in addition to the questions' overall quality.

\section{Background}
\label{sec:backgroung}

\subsection{ENADE}

Established in 2004 by the Brazilian Ministry of Education, the \textit{Exame Nacional de Desempenho dos Estudantes} (ENADE) is a federal government exam that assesses the academic performance of undergraduate students in Brazil, fostering an environment of continuous improvement in educational standards~\citep{enade}. ENADE conducts annual assessments targeting students who are nearing the completion of their academic programs. The exam committee defines the knowledge areas to be evaluated each year, with individual areas having their graduating students assessed every three years.

The exam covers various disciplines---from computer science and mathematics to history and philosophy---and encompasses bachelor, teaching, and technology degrees. The exam consists of two primary components: the first, the \textit{general education component}, common across various disciplines, assesses competencies and skills developed by students and their general knowledge relating to Brazilian and global realities. The second component, the \textit{specific knowledge component}, is tailored to each discipline, evaluating the mastery of knowledge and skills expected for the discipline's professional profile \citep{enade}. Our study focuses on the specific knowledge component of the ENADE 2021 BCS exam \citep{enade-relatorio-sintese-cc}.

\subsection{2021 Bachelor in Computer Science Exam}

ENADE's BCS exam evaluates core computer science principles and concepts such as algorithms, programming, operating systems, software engineering, artificial intelligence, and distributed systems. At the time of writing, the BCS exam had last been conducted in November 2021 with 10,119 participants from over 300 institutions across all five regions of Brazil~\citep{enade-relatorio-sintese-cc}. 

\subsubsection{Questions and Statistics}
\label{sec:questions-stats}

The 2021 BCS exam comprises three open questions and 27 multiple-choice questions. Tables~\ref{tab:enade-cs-open-questions} and \ref{tab:enade-cs-mc-questions} present the number, subject, modality (text or visual), and reasoning strategy (requiring direct or indirect answers) for the open and multiple-choice questions, respectively. Additionally, Table~\ref{tab:enade-cs-open-questions} lists the mean, median, and standard deviation scores (out of 100) for the open questions. Table~\ref{tab:enade-cs-mc-questions} includes additional statistics for the multiple-choice questions, such as the percentage of correct answers, facility level, point-biserial correlation, and discrimination level. Two questions, 29 and 33 (highlighted in red in Table~\ref{tab:enade-cs-mc-questions}), were invalidated after the exam, possibly due to poor construction. Note that questions 01 and 02, as well as questions 06--08, are not listed in these tables as they are general knowledge questions not related to computer science.

The facility level classifies the percentage of correct answers for each multiple-choice question into a hierarchy of levels, from Very Easy ($\geq 0.86$) to Very Hard ($\leq 0.15$), as outlined in Table~\ref{tab:enade-cs-facility-levels}. The discrimination level classifies the point-biserial correlation \citep{bonett2020point} between participants' performance on each question and their overall performance in the exam into another hierarchy of levels, from Very Good ($\geq 0.40$) to Weak ($\leq 0.19$), as shown in Table~\ref{tab:enade-cs-discrimination-levels}. The point-biserial correlation is commonly used in academic exams to measure the relationship between a student's overall exam performance (a continuous variable) and their response to a specific question (a dichotomous variable, correct or incorrect) \citep{rezigalla2022item}. A higher point-biserial correlation for a question indicates effective discrimination between students who score high overall and those who score low. Questions with very low or negative point-biserial correlation values may be seen as excessively difficult or potentially flawed, as they do not align with the general performance trends of the examinees and could be considered for exclusion from scoring. 

\begin{table}[t]
	\caption{ENADE 2021 BCS open questions' features and scores (out of 100).}
	\centering
	\footnotesize
	\begin{tabular}{@{}cp{3.5cm}ccrrr@{}}
		\toprule
		& & & \multirow{2}{*}{\shortstack[c]{Reasoning\\Strategy}} & \multirow{2}{*}{\shortstack[c]{Mean \\ Score}} & \multirow{2}{*}{\shortstack[c]{Median \\ Score}} & \multirow{2}{*}{\shortstack[c]{Standard \\ Deviation}} \\
		No. & \multicolumn{1}{c}{Subject} & Modality & & & \\ 
		\midrule
		\myhref{03} & Theory of Computing     & Visual &   Direct & 29.3 & 25.0 & 34.4 \\
		\myhref{04} & Computer Architecture   & Visual &   Direct &  9.8 &  0.0 & 22.8 \\
		\myhref{05} & Algorithms              & Visual &   Direct &  6.6 &  0.0 & 17.7 \\
		\midrule
		\multicolumn{4}{l}{\textbf{Agregated score}} & \textbf{15.2} & \textbf{8.3} & \textbf{25.0}\\ 
		\bottomrule
	\end{tabular}
	\label{tab:enade-cs-open-questions}
\end{table}

\begin{table}[t]
	\caption{ENADE 2021 BCS multiple-choice questions' features and statistics. Questions invalidated by the ENADE committee after the exam are highlighted in light red. Questions excluded from the final score due to statistical discrepancies are highlighted in light yellow.}
	\centering
	\footnotesize
	\begin{tabular}{@{}cp{3.5cm}ccrcrc@{}}
        \toprule 
		& & & \multirow{2}{*}{\shortstack[c]{Reasoning\\Strategy}} & \multirow{2}{*}{\shortstack[c]{\% Correct\\Answers}} & \multirow{2}{*}{\shortstack[c]{Facility\\Level}} & \multirow{2}{*}{\shortstack[c]{Point-Biserial\\Correlation}} & \multirow{2}{*}{\shortstack[c]{Discrim.\\Level}} \\
		No. & \multicolumn{1}{c}{Subject} & Modality & & & & & \\
        \midrule
        \myhref{09} & Operating Systems       & Text & Direct & 0.41 & Medium & 0.24 & Medium \\
        \myhref{10} & Programming             & Text & Direct & 0.40 &  Hard & 0.40 & Very Good \\
        \myhref{11} & Artificial Intelligence & Visual & Indirect & 0.60 &  Easy & 0.47 & Very Good \\
		\rowcolor{lightyellowcolor}
        \myhref{12} & Comp. in Society        & Text & Indirect & 0.14 &  Very Hard & 0.04 & Weak \\
		\rowcolor{lightyellowcolor}
        \myhref{13} & Software Engineering    & Text & Direct & 0.20 &  Hard & -0.04 & Weak \\
        \myhref{14} & Computer Architecture   & Text & Indirect & 0.25 & Hard & 0.26 & Medium \\
        \myhref{15} & Software Engineering    & Text & Indirect & 0.51 & Medium & 0.29 & Medium \\
        \myhref{16} & Computer Architecture   & Visual & Direct & 0.63 & Easy & 0.37 & Good \\
		\rowcolor{lightyellowcolor}
        \myhref{17} & Operating Systems       & Visual & Indirect & 0.12 & Very Hard & -0.01 & Weak \\
        \myhref{18} & Artificial Intelligence & Text & Indirect & 0.42 & Medium & 0.31 & Good \\
        \myhref{19} & Human-Computer Inter.   & Text & Indirect & 0.38 & Hard & 0.26 & Medium \\
        \myhref{20} & Programming             & Text & Indirect & 0.35 & Hard & 0.53 & Very Good \\
		\rowcolor{lightyellowcolor}
        \myhref{21} & Computer Networks       & Visual & Direct & 0.10 & Very Hard & -0.08 & Weak \\
        \myhref{22} & Information Systems     & Visual & Direct & 0.53 & Medium & 0.38 & Good \\
        \myhref{23} & Programming             & Visual & Direct & 0.45 & Medium & 0.28 & Medium \\
        \myhref{24} & Computer Security       & Text & Indirect & 0.28 & Hard & 0.33 & Good \\
		\rowcolor{lightyellowcolor}
        \myhref{25} & Distributed Systems     & Text & Indirect & 0.19 & Hard & -0.05 & Weak \\
        \myhref{26} & Web Development         & Text & Indirect & 0.15 & Very Hard & 0.23 & Medium \\
        \myhref{27} & Performance Analysis    & Visual & Direct & 0.51 & Medium & 0.45 & Very Good \\
        \myhref{28} & Computer Architecture   & Visual & Indirect & 0.34 & Hard & 0.34 & Good \\
		\rowcolor{lightredcolor}
        \myhref{29} & Image Processing        & Visual & Indirect & & & & \\
        \myhref{30} & Compilers               & Text & Indirect & 0.45 & Medium & 0.37 & Good \\
        \myhref{31} & Theory of Computing     & Visual & Indirect & 0.23 & Hard & 0.37 & Good \\
        %\rowcolor{gray!25}
		\rowcolor{lightyellowcolor}
        \myhref{32} & Algorithms              & Text & Indirect & 0.17 & Hard & 0.02 & Weak \\
        %\rowcolor{gray!60}
		\rowcolor{lightredcolor}
        \myhref{33} & Programming             & Text & Direct & & & & \\
        \myhref{34} & Graph Theory     & Visual & Direct & 0.23 & Hard & 0.20 & Medium \\
		\myhref{35} & Distributed Systems     & Visual & Direct & 0.51 & Medium & 0.48 & Very Good \\
		\bottomrule
	\end{tabular}
	\label{tab:enade-cs-mc-questions}
\end{table}

\begin{table}[t]
	\caption{ENADE 2021 BCS classification of multiple-choice questions into facility levels.}
	\centering
	\footnotesize
	\begin{tabular}{@{}cc@{}}
		\toprule
		\% Correct Answers & Facility Level \\
		\midrule
		$\geq 0.86$ & Very Easy \\
		$0.61$ to $0.85$ & Easy \\
		$0.41$ to $0.60$ & Medium \\
		$0.16$ to $0.40$ & Hard \\
		$\leq 0.15$ & Very hard \\
		\bottomrule
	\end{tabular}
	\label{tab:enade-cs-facility-levels}
\end{table}

\begin{table}[t]
	\caption{ENADE 2021 BCS classification of multiple-choice questions into discrimination levels.}
	\centering
	\footnotesize
	\begin{tabular}{@{}cc@{}}
		\toprule
		Point-Biserial Correlation & Discrimination Level \\
		\midrule
		$\geq 0.40$ & Very Good \\
		$0.30$ to $0.39$ & Good \\
		$0.20$ to $0.29$ & Medium \\
		$\leq 0.19$ & Weak \\
		\bottomrule
	\end{tabular}
	\label{tab:enade-cs-discrimination-levels}
\end{table}

The ENADE 2021 committee utilized the calculated facility and discrimination levels to exclude six multiple-choice questions from the final score, namely questions 12, 13, 17, 21, 25, and 32 (highlighted in light yellow in Table~\ref{tab:enade-cs-mc-questions}) \citep{enade-relatorio-sintese-cc}. These questions were classified as Hard or Very Hard at the facility level and Weak at the discrimination level.

Appendix~\ref{app:conversations} presents the image and English transcription of the three open questions and a sample of six multiple-choice questions of the ENADE 2021 BCS exam.
\footnote{In the PDF version of the paper, all question numbers in Tables~\ref{tab:enade-cs-open-questions} and \ref{tab:enade-cs-mc-questions} are hyperlinked to their corresponding page in the paper's companion repository, where readers can view the questions' original images, ChatGPT-4 prompts and responses, and the independent expert evaluations (when available) in both English and Portuguese.}

\subsubsection{Scores}

\begin{table}[t]
	\caption{ENADE 2021 BCS specific knowledge component scores.}
	\centering
	\footnotesize
	\begin{tabular}{@{}p{3.5cm}rrrr@{}}
		\toprule
		& & \multirow{2}{*}{\shortstack[c]{Exam \\ Mean}} & \multirow{2}{*}{\shortstack[c]{Exam \\ Median}} & \multirow{2}{*}{\shortstack[c]{Standard \\ Deviation}} \\
		Section & Weight & & & \\ 
		\midrule
		Open questions            & 0.15 & 15.2 &  8.3 & 19.5 \\
		Multiple-choice questions & 0.85 & 40.5 & 42.1 & 16.5 \\
		\midrule
		\textbf{Final  (weighed) score}               &      & \textbf{36.7} & \textbf{35.8} & \textbf{15.8} \\
		\bottomrule
	\end{tabular}
	\label{tab:enade-cs-scores}
\end{table}

ENADE 2021 calculates the final score of the specific knowledge component by combining the scores of the open and multiple-choice questions, with the former weighting 15\% of the final score and the latter weighting 85\% \citep{enade-relatorio-sintese-cc}. The open question score is calculated as the average score of the three open questions. The multiple-choice question score is calculated as the proportion of correct responses to the remaining 19 out of the initial 27 multiple-choice questions. The final score is then converted to a scale of 0 to 100, with 0 being the lowest possible score and 100 being the highest. 

Table~\ref{tab:enade-cs-scores} presents the numbers for the open questions, multiple-choice questions, and final scores for the 2021 BCS exam. The mean and median scores for the open questions were 15.2 and 8.3, respectively, with a standard deviation of 19.5. The mean and median scores for the multiple-choice questions were 40.5 and 42.1, respectively, with a standard deviation of 16.5. Lastly, the mean and median scores for the final score were 36.7 and 35.8, respectively, with a standard deviation of 15.8. These numbers indicate that the exam's open questions were particularly challenging for the participants. In contrast, the multiple-choice questions were relatively more accessible, even though the average score in this section was still below the 50\% threshold. This result aligns with the observation that nearly half of the 19 scored multiple-choice questions were classified as Hard or Very Hard, as shown in Table~\ref{tab:enade-cs-mc-questions}.

\section{Study Method}
\label{sec:method}

Our study evaluates ChatGPT-4 Vision's performance on the specific knowledge component of the ENADE 2021 BCS exam. To this end, we provided the model with the exam's questions in image format by taking a high-resolution screen capture of each question's page from the exam's official PDF document. Since the exam's questions and response standard are in Portuguese, we conducted the evaluation experiment in Portuguese, including the model's prompts and responses, and the expert assessments of conflicting answers. We translated all materials into English for this paper. Although we have not replicated the entire experiment using the English translations of the questions, our experience with ChatGPT-Vision and its successors is that these OpenAI models are as good at answering questions in Portuguese as they are in English.

\subsection{ChatGPT Versions Used in the Study}

We used ChaGPT-4 Vision \citep{openai2023gpt4v} to answer ENADE's 2021 BCS exam questions and ChatGPT-4 Turbo \citep{openai2023chatgpt-turbo}, to reflect on ChaGPT-4 Vision's performance on each question. The reason we have not used ChatGPT-4 Turbo from the start is that it was only released after we completed the question-answering phase of the research using ChatGPT-4 Vision.

ChatGPT-4 Vision, released in September 2023, expands upon the text-based capabilities of ChatGPT-4 by incorporating image analysis and understanding \citep{openai2023gpt4v}. It features a context window of 32k tokens and an updated knowledge base until January 2022. Since the ENADE 2021 exam was administered on November 14th, 2021, and the exam key and response standard were made publicly available on March 18th, 2022, it is reasonable to assume ChatGPT-4 Vision's knowledge base did not include any information from the exam.

ChatGPT-4 Turbo, which was launched in November 2023, features a significantly larger context window (128k tokens) compared to GPT-4, allowing it to handle complex and lengthy prompts with greater accuracy and coherence \citep{openai2023chatgpt-turbo}. Additionally, ChatGPT-4 Turbo incorporates Retrieval Augmented Generation (RAG) \citep{lewis2020retrieval}, which integrates external information to enhance the model's responses. Furthermore, ChatGPT-4 Turbo boasts DALL-E~3 integration for text-to-image generation, text-to-speech capabilities with multiple voice options, and an updated knowledge base extending to April 2023. 

\subsection{Evaluation Procedure}

\begin{figure}[t]
	\centering
	\figbox{\includegraphics[width=0.9\textwidth]{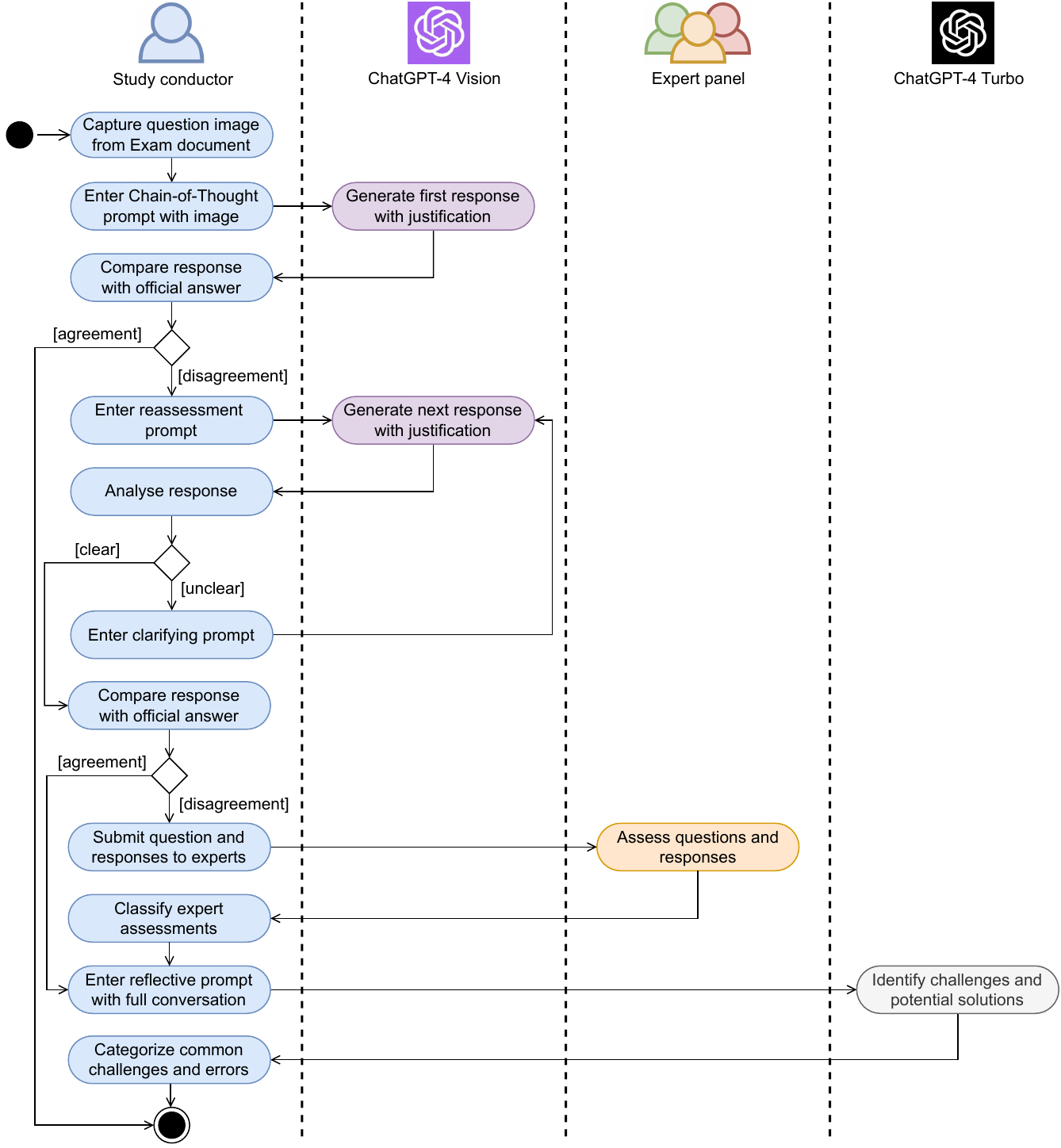}}
	% \fbox{}
	\caption{Question-answering evaluation procedure.}
	\label{fig:enade2021-cs-method}
\end{figure}

Our evaluation procedure for each exam question involved a multi-step process, as illustrated in the activity diagram shown in Figure~\ref{fig:enade2021-cs-method}. Initially, we captured an image of the question from the exam's PDF file. We then provided this image along with a simple (zero-shot) Chain-of-Thought (CoT) prompt \citep{wei2022chain} to ChatGPT-4 Vision, asking the model to answer the question depicted in the image and to detail its reasoning step by step. This prompting approach was chosen because it enables the model to demonstrate its ability to understand and solve complex problems without prior specific training on the exam's content, relying instead on its general pre-trained knowledge base. Moreover, by incorporating CoT prompts, the evaluation process gains transparency and insight into the model's reasoning, allowing the assessment not just of the correctness of the final answers but also of the logical processes used to arrive at them. This is especially valuable in an educational context, where understanding the steps of problem-solving is as important as the solution itself. Appendix~\ref{app:prompt-templates} shows our study's zero-shot CoT prompt templates for the ENADE 2021 BCS open and multiple-choice exam questions. 

After providing ChatGPT-4 Vision with the appropriate prompt, we compared the model's generated response against the exam's official answer standard (for open questions) or answer key (for multiple-choice questions). When the model's response did not align with the official answer, we prompted the model to reassess its response in light of the official answer and self-reflect on its possible reasoning mistakes. Essentially, our reassessment prompts inform the model that its initial answer is inconsistent with the official answer and instruct it to ponder whether to maintain or change its initial answer given the new information. Appendix~\ref{app:prompt-templates} shows our reassessment prompt templates for open and multiple-choice questions. We interactively analyzed the model's subsequent responses and prompted it for further clarification when the model's decision on the correct answer was unclear or incoherent.

Following the clarification stage, if discrepancies persisted between ChatGPT-4 Vision's responses and the official answer, the conflicting answers were escalated to an independent expert panel for review. The panel comprised two or more computer science researchers from multiple Brazilian universities with known expertise on the question's subject. We provided each panel member with the question, official answer, and ChatGPT-4 Vision's responses. They were individually asked to assess the correctness of the model's responses and provide their reasoning for their decisions without knowing the identity and opinions of their peers. The panel members were also asked to indicate if the question was unclear or had multiple or no possible correct answers. We then compiled and classified the panel's assessments as conclusive or inconclusive. Conclusive assessments were those where all panel members agreed on the question's correct answer. Inconclusive assessments were those where the panel members disagreed on the correctness of the question's answer or indicated that the question was unclear or had no possible answer. 

In the final evaluation step, we provided ChatGPT-4 Turbo with the question image, ChatGPT-4 Vision's prompts and responses, and the expert assessments (when available) for each question in which the model did not initially agree with the official response and prompted it to reflect on the entire question-answering process. In particular, we asked the model to identify the possible challenges the AI faced in answering the question, the cause of its errors, and whether these errors could be mitigated by using more sophisticated prompt engineering techniques or would be inherent to the limitations of current LLMs. This final reflective assessment aimed to provide further insights into the AI's reasoning and decision-making processes, contributing to a deeper understanding of the model's multimodal capabilities and limitations. 

\subsection{Question Scoring}

We scored ChatGPT-4 Vision's responses to the three open questions using the exam's official answer standard as a reference. We assigned scores from 0 to 100 to each response, where 0 indicated an utterly incorrect response, and 100 indicated an entirely correct response. We scored partially correct responses proportionally, based on the number of correct items or assertions they contained. For example, a response with two correct assertions out of three would be scored 66.6. 

We scored ChatGPT-4 Vision's responses to the multiple-choice questions proportionally to the number of responses that agreed with the exam's answer key. For reference, we scored all 25 valid multiple-choice questions, including the six questions the exam committee excluded from the final score. However, we only consider the 19 valid non-excluded questions for computing the model's final score.

We only scored the model's initial response to each question and ignored its subsequent reassessments or the expert panel's reviews in determining the score value. For example, if the model initially answered a question incorrectly but later changed its response after reassessment, we only considered the initial incorrect response for scoring purposes.

\subsection{Experimental Setup}

During the question-answering sessions, we manually interacted with both ChatGPT-4 Vision and ChatGPT-4 Turbo using their respective web interfaces. We did not use an automated script for this task because ChatGPT-4 Vision's model was not available via API at the time the experiment was conducted. We used the models' default settings with no parameter fine-tuning. We saved all the model's prompts and responses and the expert assessments as text files in their original Portuguese version. We translated them into English using the GPT-4 Turbo API via a customized Python script with minor manual revisions and improvements. We also translated the exam's official answer standard for open questions into English using the same script. 

\section{Results}
\label{sec:results}

We now present the results for each of the five research questions outlined in Section~\ref{sec:introduction}. We first analyze ChatGPT-4 Vision's performance in answering the ENADE 2021 BCS exam questions, focusing on the open and multiple-choice questions. We then examine the model's response changes when confronted with disagreeing official answers and compare its final score with the exam participants. Finally, we discuss the main challenges ChatGPT-4 Vision faced in answering the exam questions and reflect on its reasoning and decision-making processes.

\subsection{RQ1: Performance in Open Questions}

\begin{table}[t]
	\caption{Human participants and ChatGPT-4 Vision's scores for open questions.}
	\centering
	\footnotesize
	\begin{tabular}{@{}cccrrrrrr@{}}
		\toprule
		& & & \multicolumn{3}{c}{Human Participants} & \multicolumn{3}{c}{ChatGPT-4 Vision} \\
		\cmidrule[0.5pt](r{0.5em}){4-6} \cmidrule[0.5pt](l{0.5em}){7-9}
		& & \multirow{2}{*}{\shortstack[c]{Reason.\\Strategy}} & \multirow{2}{*}{\shortstack[c]{Mean \\ Score}} & \multirow{2}{*}{\shortstack[c]{Median \\ Score}} & \multirow{2}{*}{\shortstack[c]{Standard \\ Deviation}} & & \multirow{2}{*}{\shortstack[c]{Imp./\\Mean}} & \multirow{2}{*}{\shortstack[c]{Top\\Perc.}} \\
		No. & Modality & & & & & Score & & \\ 
		\midrule
		\myhref{03} & Visual &   Direct & 29.3 & 25.0 & 34.4 & 62.5 & 2.1 & 22 \\
		\myhref{04} & Visual &   Direct &  9.8 &  0.0 & 22.8 & 50.0 & 5.1 & 17 \\
		\myhref{05} & Visual &   Direct &  6.6 &  0.0 & 17.7 & 50.0 & 7.6 & 11 \\
		\midrule
		\multicolumn{2}{r}{\textbf{Agregated score}} & & \textbf{15.2} & \textbf{8.3} & \textbf{15.8} & \textbf{54.2} & \textbf{3.6} & \textbf{8} \\ 
		\bottomrule
	\end{tabular}
	\label{tab:enade-cs-open-questions-chatgpt-4-scores}
\end{table}

Table~\ref{tab:enade-cs-open-questions-chatgpt-4-scores} shows ChatGPT-4 Vision's score (out of 100), improvement over mean, and top percentile relative to the participants' mean, median, and standard deviation scores for the three open questions. The improvement over mean metric for ChatGPT-4 Vision is calculated by comparing the model's score to the mean score of the participants, specifically by dividing the model's score by the mean participant score. The table shows that ChatGPT-4 Vision failed to provide entirely correct answers to all three questions, with an average score of 54.2. Nevertheless, the model still performed much better than most participants in all three questions, significantly more so for questions 04 and 05, where the model's score was five to seven times higher than the participants' mean scores. These results place the model's scores within the top 10-20 best score percentiles for all three open questions, with its average score being within the top 8 best average score percentile.

\subsection{RQ2: Performance in Multiple-Choice Questions}

\begin{table}[t]
	\caption{ChatGPT-4 Vision's results for multiple-choice questions.}
	\centering
	\footnotesize
	\begin{tabular}{@{}p{1.5cm}ccccrrrrr@{}}
		\toprule
		& & & & & & \multicolumn{4}{c}{ChatGPT-4 Vision Performance} \\
		\cmidrule[0.5pt](l){7-10}
		& & \multirow{2}{*}{\shortstack[c]{Reason.\\Strategy}} & \multirow{2}{*}{\shortstack[c]{Facility\\Level}} & \multirow{2}{*}{\shortstack[c]{Disc.\\Level}} & & & & & \multirow{2}{*}{\shortstack[c]{Score\\(\% Corr.)}}\\
		Scope & Modality & & & & Total & Corr. & Incorr. & Confl. & \\ 
		\midrule
		\multirow{2}{*}{\shortstack[l]{Scored\\questions}} & \textbf{all} & \textbf{all} & \textbf{all}  & \textbf{all} & \textbf{19} & \textbf{13} & \textbf{5} & \textbf{1} & \textbf{68.4}\\
		\cmidrule[0.5pt](l){2-10}
		& Text & all & all & all & 11 & 7 & 3 & 1 & 63.6\\
		& Visual & all & all & all & 8 & 6 & 2 & 0 & 75.0\\
		\cmidrule[0.5pt](l){2-10}
		& all & Direct & all & all & 8 & 6 & 2 & 0 & 75.0\\
		& all & Indirect & all & all & 11 & 7 & 3 & 1 & 63.6\\
		\cmidrule[0.5pt](l){2-10}
		& all & all & Easy & all & 2 & 2 & 0 & 0 & 100.0\\
		& all & all & Medium & all & 8 & 6 & 1 & 1 & 75.0\\
		& all & all & Hard & all & 8 & 5 & 3 & 0 & 62.5\\
		& all & all & Very Hard & all & 1 & 0 & 1 & 0 & 0.0\\
		\cmidrule[0.5pt](l){2-10}
		& all & all & all & Very Good & 5 & 5 & 0 & 0 & 100.0\\
		& all & all & all & Good & 7 & 5 & 1 & 1 & 71.4\\
		& all & all & all & Medium & 7 & 3 & 4 & 0 & 42.9\\
		\midrule
		\multirow{2}{*}{\shortstack[l]{Excluded\\questions}} & \textbf{all} & \textbf{all} & \textbf{all} & \textbf{Weak} & \textbf{6} & \textbf{2} & \textbf{2} & \textbf{2} & \textbf{33.3}\\
		\cmidrule[0.5pt](l){2-10}
		& Text & all & all & Weak & 4 & 1 & 1 & 2 & 25.0\\
		& Visual & all & all & Weak & 2 & 1 & 1 & 0 & 50.0\\
		\cmidrule[0.5pt](l){2-10}
		& all & Direct & all & Weak & 2 & 0 & 1 & 1 & 0.0\\
		& all & Indirect & all & Weak & 4 & 2 & 1 & 1 & 50.0\\
		\cmidrule[0.5pt](l){2-10}
		& all & all & Hard & Weak & 3 & 1 & 0 & 2 & 33.3\\
		& all & all & Very Hard & Weak & 3 & 1 & 2 & 0 & 33.0\\
		% \midrule
		% \multirow{2}{*}{\shortstack[l]{Valid\\questions}} & Text & all & all & all & 15 & 8 & 4 & 3 & \textbf{53.3}\\
		% & Visual & all & all & all & 10 & 7 & 3 & 0 & \textbf{70.0}\\
		% \cmidrule[0.5pt](l){2-10}
		% & all & Direct & all & all & 10 & 6 & 3 & 1 & \textbf{60.0}\\
		% & all & Indirect & all & all & 15 & 9 & 4 & 2 & \textbf{60.0}\\
		% \cmidrule[0.5pt](l){2-10}
		% & all & all & Easy & all & 2 & 2 & 0 & 0 & \textbf{100.0}\\
		% & all & all & Medium & all & 8 & 6 & 1 & 1 & \textbf{75.0}\\
		% & all & all & Hard & all & 11 & 6 & 3 & 2 & \textbf{54.5}\\
		% & all & all & Very Hard & all & 4 & 1 & 3 & 0 & \textbf{25.0}\\
		% \cmidrule[0.5pt](l){2-10}
		% & all & all & all & Very Good & 5 & 5 & 0 & 0 & \textbf{100.0}\\
		% & all & all & all & Good & 7 & 5 & 1 & 1 & \textbf{71.4}\\
		% & all & all & all & Medium & 7 & 3 & 4 & 0 & \textbf{42.9}\\
		% & all & all & all & Weak & 6 & 2 & 2 & 2 & \textbf{33.3}\\
		% \cmidrule[0.5pt](l){2-10}
		% & \textbf{all} & \textbf{all} & \textbf{all} & \textbf{all} & \textbf{25} & \textbf{15} & \textbf{7} & \textbf{3} & \textbf{60.0}\\
		\bottomrule
	\end{tabular}
	\label{tab:enade-cs-mc-questions-chatgpt-4-scores}
\end{table}

Table~\ref{tab:enade-cs-mc-questions-chatgpt-4-scores} shows ChatGPT-4 Vision's performance for all valid (scored and excluded) multiple-choice questions. The model's results are shown for all question types of each category and also grouped row-wise by modality, reasoning strategy, facility level, and discrimination level. 

ChatGPT-4 Vision's overall score, considering all 19 scored questions, was 68.4, with 13 correct answers, five incorrect answers, and one conflicting answer where the experts disagreed on the correct answer. The model answered correctly 75.0\% of questions with visual elements and 63.6\% of text-only questions. The model showed a similar scoring pattern for questions requiring direct and indirect answers, with 75.0\% and 63.6\% correct answers, respectively. These results indicate that the model performed slightly better in questions with visual elements and direct answers than in text-only questions or those requiring multi-step answers. 

The model's performance degraded as the questions' facility level increased from Easy to Very Hard, with 100.0\%, 75.0\%, 62.5\%, and 0.0\% correct answers, respectively. We observed a similar degradation pattern in the model's performance when grouping questions by discrimination level, where it answered correctly 100.0\%, 71.4\%, and 42.9\% of questions classified as Very Good, Good, and Medium, respectively. 

The model's performance in the six excluded questions was significantly lower than in the scored questions, with only 33.3\% correct answers compared to 68.4\% in the first category. Even when grouping the excluded questions by modality, reasoning strategy, and facility and discrimination levels, the model did not get more than half of the questions correct in either group. These results suggest that ChatGPT-4 Vision's performance in the multiple-choice questions is generally aligned with the overall performance of the exam's participants, with the model scoring higher in questions classified as Medium to Easy at the facility level and Medium to Very Good at the discrimination level. 

We applied Spearman's rank correlation coefficient \citep{zar2005spearman} to test this hypothesis by calculating the correlation between the AI model's performance and each question's facility and discrimination levels. This method was chosen for its robustness in handling non-parametric data and its effectiveness in measuring the strength and direction of association between two ranked variables. The AI model's answers were numerically coded to facilitate the analysis: correct answers as 1 and 0 otherwise. The facility level and the discrimination level were numerically represented by the percentage of correct answers and the Point-Biserial correlation values, respectively.

Our Spearman's rank correlation analysis indicated a coefficient of 0.555 ($p = 0.004$) between the AI model's performance and the facility level. The analysis also yielded a Spearman's correlation coefficient of 0.584 ($p = 0.0022$) when comparing the AI model's performance with the Point-Biserial Correlation values. These results reveal statistically significant positive correlations ($p < 0.05$) between the AI model's performance and the exam questions' facility and discrimination levels, as evidenced by the $p$-values reported. Thus, the correlation results support our claim that ChatpGPT-4 Vision is more adept at answering questions that are both easier for students and more discriminative regarding student performance. 

\subsection{RQ3: Performance in Questions with Disagreeing Answers}

\begin{table}[t]
	\caption{ChatGPT-4 Vision's responses and expert panel assessments for multiple-choice questions with initially disagreeing responses. Questions excluded from the final score due to a weak point-biserial correlation are highlighted in light yellow. Questions where the expert panel disagreed on the correct answer are highlighted in boldface.}
	\centering
	\footnotesize
	\begin{tabular}{@{}cccccccc@{}}
    \toprule
		& & \multicolumn{2}{c}{ChatGPT-4 Vision} & \multicolumn{3}{c}{Independent Expert Panel} \\
		\cmidrule[0.5pt](r){3-4} \cmidrule[0.5pt]{5-7}
		& \multirow{2}{*}{\shortstack[c]{Answer\\Key}} & \multirow{2}{*}{\shortstack[c]{Initial\\Response}} & \multirow{2}{*}{\shortstack[c]{Revised\\Response}} & & & & \multirow{2}{*}{\shortstack[c]{Final\\Assessment}}\\
		No. & & & & Expert ~\#1 & Expert~\#2 & Expert~\#3 & \\  
        \midrule
        \myhref{09} & E & C & C & E & E & & Incorrect \\
 		\rowcolor{lightyellowcolor}
        \myhref{12} & B & D & B & & & & Incorrect \\
 		\rowcolor{lightyellowcolor}
        \textbf{\myhref{13}} & \textbf{E} &\textbf{ D} & \textbf{D} & \textbf{D} & \textbf{Invalid} & \textbf{E} & \textbf{Inconclusive} \\
        \myhref{14} & C & B & B & C & C & C & Incorrect \\
        \textbf{\myhref{18}} & \textbf{E} & \textbf{D} & \textbf{D} & \textbf{Invalid} & \textbf{A} & \textbf{E} & \textbf{Inconclusive} \\
		\rowcolor{lightyellowcolor}
        \myhref{21} & E & Invalid & E & & & & Incorrect \\
		\rowcolor{lightyellowcolor}
        \textbf{\myhref{25}} & \textbf{B} & \textbf{E} & \textbf{D} & \textbf{Invalid} & \textbf{B} & \textbf{Invalid} & \textbf{Inconclusive} \\
        \myhref{26} & A & E & A & & & & Incorrect \\
        \myhref{31} & E & D & Invalid & E & E & & Incorrect \\
        \myhref{34} & C & Invalid & C & & & & Incorrect \\
		\bottomrule
	\end{tabular}
	\label{tab:enade-cs-mc-questions-responses-reviews}
\end{table}

Table~\ref{tab:enade-cs-mc-questions-responses-reviews} shows ChatGPT-4 Vision's initial and revised responses and the expert panel's assessments (when available) for the ten multiple-choice questions where the model initially disagreed with the official answer. This table reveals that the model initially disagreed and subsequently agreed with the official answer key after reassessment in four questions (12, 21, 26, and 34). These four questions were classified as Hard (21) or Very Hard (12, 26, and 34), with two (12 and 21) being excluded from the final score due to a Weak discrimination level (see Table~\ref{tab:enade-cs-mc-questions}). Notably, for questions 21 and 34, ChatGPT-4 Vision initially suggested invalidation but revised its stance to align with the official answer upon reassessment.

Out of the six questions where the model's disagreement with the answer key persisted after reassessment, the expert panel unanimously agreed with the answer key in three questions (09, 14, and 31) and did not agree on the correct answer for the remaining three questions (13, 18, 25). The three questions where the expert panel agreed on the correct answer were classified from Medium (09) to Hard (14 and 31) at the facility level and from Medium (09 and 14) to Good (31) at the discrimination level. On the other hand, two of the three questions where the expert panel disagreed on the correct answer (13 and 25) were classified as Hard with a Weak discrimination level, thus being among the six excluded from the final score. Question 18 is the exception from this group since it was classified as Medium at the facility level and Good at the discrimination level. This discrepancy means that the AI model and the expert panel struggled to provide a definitive answer to a question that a significant fraction of the exam's participants did not find particularly difficult.

The results in Table~\ref{tab:enade-cs-mc-questions-responses-reviews} also reinforce the claim that the model's performance tends to reflect that of the human participants in the multiple-choice section of the exam, with both the model and the expert panel struggling to provide a definitive answer to some questions classified as Hard or Very Hard with a Weak discrimination level. 

\subsection{RQ4: Final Score and Relative Performance}

\begin{table}[t]
	\caption{ChatGPT-4 Vision's partial and final scores for the open and multiple-choice questions of the ENADE 2021 BCS exam.}
	\centering
	\footnotesize
	\begin{tabular}{@{}p{3.5cm}rrrrrrr@{}}
		\toprule
		& & & & & \multicolumn{3}{c}{ChatGPT-4 Vision} \\
		\cmidrule[0.5pt]{6-8}
		& & \multirow{2}{*}{\shortstack[c]{Exam \\ Mean}} & \multirow{2}{*}{\shortstack[c]{Exam \\ Median}} & \multirow{2}{*}{\shortstack[c]{Standard \\ Deviation}} & & \multirow{2}{*}{\shortstack[c]{Imp./\\Mean}} & \multirow{2}{*}{\shortstack[c]{Top\\Perc.}}\\
		Scope & Weight & & & & Score & & \\ 
		\midrule
		Open questions            & 0.15 & 15.2 &  8.3 & 19.5 & 54.2 & 3.6 & 8 \\
		Multiple-choice questions & 0.85 & 40.5 & 42.1 & 16.5 & 68.4 & 1.7 & \\
		\midrule
		\textbf{Final (weighed) score}               &      & \textbf{36.7} & \textbf{35.8} & \textbf{15.8} & \textbf{66.3} & \textbf{1.8} & \textbf{10} \\
		\bottomrule
	\end{tabular}
	\label{tab:enade-cs-chatgpt-4-final-score}
\end{table}

Table~\ref{tab:enade-cs-chatgpt-4-final-score} shows ChatGPT-4 Vision's partial and final scores for the specific knowledge component of the ENADE 2021 BCS exam. Considering both open and multiple-choice questions, the model's final (weighed) score was 66.3. This score, representing an 80\% improvement over the participants' mean final score of 36.7, places ChatGPT-4 Vision within the top 10 percentile of exam scores \citep{enade-relatorio-sintese-cc}. This result indicates that ChatGPT-4 Vision performed significantly better than the average human participant in the ENADE 2021 BCS exam.

\subsection{RQ5: Challenges in Answering Open and Multiple-Choice Questions}

\begin{table}[t]
	\caption{Challenges ChatGPT-4 Vision faced in answering ENADE 2021 BCS questions.}
	\centering
	\footnotesize
	\begin{tabularx}{\linewidth}{@{}p{2cm}p{2.4cm}Xp{3cm}@{}}
		\toprule
		% & & & \multirow{2}{*}{\shortstack[c]{Open\\Questions}} & \multirow{2}{*}{\shortstack[c]{Multiple-Choice\\Questions}}\\	
		\multicolumn{1}{c}{Challenge} & \multicolumn{1}{c}{Error Type} & \multicolumn{1}{c}{Example} & \multicolumn{1}{c}{Questions with Error} \\ 
		\midrule
		Question \newline Interpretation & Inconsistent \newline Responses & The AI model produces inconsistent responses due to vague or ambiguous statements in the question. & \myhref{13}, \myhref{18}, \myhref{25}  \\
		\cmidrule{2-4}
		& Non-Compliance \newline with Guidelines & The AI model ignores specific directives or constraints, resulting in an incorrect response that does not align with the question's guidelines.  & \myhref{31} \\
		\midrule
		Logical \newline Reasoning & Evaluation \newline Leniency & The AI model identifies yet overlooks ``minor'' flaws, validating a false or partly correct assertion as true. & \myhref{12}, \myhref{14}, \myhref{21} \\
		\cmidrule{2-4}
		& Insufficient \newline Domain \newline Knowledge & The AI model applies an incorrect or incomplete premise, causing a false or partly correct assertion to be misidentified as true. & \myhref{09}, \myhref{13}, \myhref{14}, \myhref{26} \\
		\cmidrule{2-4}
		& Incorrect \newline Algorithmic \newline Reasoning & The AI model fails to simulate the execution of an algorithm accurately. & \myhref{05}, \myhref{31}  \\
		\cmidrule{2-4}
		& Incorrect \newline Multi-Step \newline Reasoning & The AI model correctly evaluates the validity of individual assertions but then chooses a ``true'' answer that includes or depends upon one or more false assertions. & \myhref{03}, \myhref{12}, \myhref{18}, \myhref{25}, \myhref{26} \\
		\midrule
		Visual  \newline Acuity & Misidentification \newline of Visual \newline Elements & The AI model misidentifies one or more labels in a table or diagram. & \myhref{03}, \myhref{31}, \myhref{34}  \\
		\cmidrule{2-4}
		& Lack of Domain-\newline Specific Visual \newline Output & The AI model cannot generate a response using the required visual formalism. & \myhref{04} \\
		\bottomrule
	\end{tabularx}
	\label{tab:enade-cs-chatgpt-4-challenges}
\end{table}

Based on our analysis of ChatGPT-4 Vision's initially incorrect responses, the expert assessments (when available), and ChatGPT-4 Turbo's responses, we identified three challenges categories: \textit{Question Interpretation}, \textit{Logical Reasoning}, and \textit{Visual Acuity}. We further classified the most common errors generated by the model when facing each of these challenges into eight error types, as described and illustrated in Table~\ref{tab:enade-cs-chatgpt-4-challenges}. Question interpretation errors are caused by the model's inability to correctly interpret the questions' statements. Logical reasoning errors arise from the model's flawed reasoning processes when answering the assertions in the questions. Finally, visual acuity errors are caused by the model's inability to correctly identify and process the questions' visual elements. These challenge categories and error types are not mutually exclusive, as the manifestation of errors related to one challenge may lead to the manifestation of other errors related to another challenge. For example, the model's misidentification of a visual element in a question may trickle down to an incorrect multi-step reasoning process.

The last column of Table~\ref{tab:enade-cs-chatgpt-4-challenges} reveals that the most common error types generated by ChatGPT-4 Vision in answering the ENADE 2021 BCS questions were related to logical reasoning, particularly incorrect multi-step reasoning (in five questions), insufficient domain knowledge (in four questions), and evaluation leniency (in three questions). Inconsistent responses and misidentification of visual elements (in three questions each) were also common error types. Other less frequent error types included incorrect algorithmic reasoning and non-compliance with guidelines (in two questions), and lack of domain-specific visual output (in one question). Appendix~\ref{app:conversations} shows the model's conversations for the three open and six multiple-choice questions illustrating most of these challenges categories and error types. The model's full conversations for all the questions are available in the paper's companion repository.

These results provide several insights into ChatGPT-4 Vision's multimodal reasoning capabilities. For example, the model's frequent incorrect multi-step reasoning and insufficient domain knowledge suggest its ability to reason about complex, multi-step problems in computer science is still limited. Moreover, the model's multiple evaluation leniency issues indicate its ability to evaluate the validity of assertions is not yet as strict as that of the human experts. Finally, the model's tendency to misidentify visual elements and lack of domain-specific visual output show that its visual acuity is not yet as developed as its text-based reasoning capabilities. This limitation is particularly critical given the frequent use of visual data in academic assessments in computer science and other science and engineering fields.

Another relevant result was the model's struggle with vague or ambiguous statements, particularly in questions where the expert panel did not agree on the correct answer, such as in questions 13, 18, and 25. This result suggests that some of the exam's questions may have been poorly constructed, leading to confusion and disagreement among the model and the human experts. Indeed, some experts explicitly mentioned the low quality of some ENADE questions, with one expert being even more forthright when assessing the quality of one particular question: ``I'm sad that such a poorly formulated question was used to evaluate computer science students.''

It is worth noting that these challenges did not manifest uniformly across the open and multiple-choice questions. For example, the model's issues with incorrect multi-step reasoning were more frequent in the multiple-choice questions with indirect answers. At the same time, the lack of domain-specific visual output is only an issue for open questions requiring visual responses. Moreover, some challenges were not uniformly distributed across questions with similar characteristics. For example, the model's issues with misidentifying visual elements did not occur in all questions containing figures or diagrams. These results suggest that the model's challenges are highly context-dependent and may require tailored solutions to address them effectively.

\section{Discussion}\label{sec:discussion}

Given the results presented in the previous section, we discuss the implications of ChatGPT-4 Vision's performance in the ENADE 2021 BCS exam and the insights gained from the model's challenges in answering the exam's questions. We also discuss the limitations of our study and future research directions.

\subsection{Implications of ChatGPT-4 Vision's Performance}

The performance of ChatGPT-4 Vision on the ENADE 2021 BCS exam highlights significant advancements in AI's ability to engage with complex, multimodal academic content. The AI's overall score, comparable to that of the top-scoring human participants, particularly its superior performance in open questions and its competitive edge in multiple-choice questions with visual elements, underscores the potential of multimodal AI in educational settings. These results indicate the evolving capabilities of AI systems to process and analyze information in a manner previously thought to be exclusive to human cognition \citep{zhai2024generative}. Such capabilities could be harnessed to support various educational tasks, such as developing advanced tutoring systems, aiding educational assessment, and providing personalized learning experiences \citep{li2023adapting}. The model's ability to reason about complex, multimodal problems in computer science is particularly promising, given the growing demand for AI systems that can understand and reason about multimodal data in this education domain \citep{zhou2024the}.

\subsection{Insights from ChatGPT-4 Vision's Challenges}

The challenges faced by ChatGPT-4 Vision, particularly in logical reasoning, question interpretation, and visual acuity, provide valuable insights into the current limitations of AI in fully understanding and responding to complex multimodal academic problems. The model's difficulties with incorrect multi-step reasoning and insufficient domain knowledge reinforce similar findings reported by prior studies (e.g., \citep{bubeck2023sparks,mao2023gpteval,espejel2023gpt}). Nevertheless, these insights provide valuable guidance for future improvements to the model's reasoning capabilities, perhaps through more sophisticated training datasets or advanced reasoning algorithms \citep{tyen2024llms}. The model's challenges with visual acuity also highlight the importance of developing AI systems that can accurately interpret and integrate multi-modal inputs \citep{zhao2023mmicl}. Furthermore, the model and expert panel's struggle with vague or ambiguous statements could inform educational assessment design, helping improve the clarity and precision of questions during exam elaboration. Finally, the finding that the model's performance is correlated with the difficulty and discriminative power of the exam's multiple-choice questions suggests that ChatGPT-4 Vision's performance could be used as a proxy for question quality and complexity, thus providing a valuable tool for exam creators to test and refine their questions before administering the exam.

\subsection{Study Limitations and Future Directions}

Our study has certain limitations that are important to acknowledge. First, the reliance on zero-shot CoT prompts may have limited the AI's ability to fully engage with the multi-step reasoning required by some exam questions, potentially underestimating its true reasoning capabilities. Future studies should experiment with advanced prompting techniques like Self-Consistency \citep{wang2023selfconsistency}, Tree-of-Thoughts \citep{yao2023tree}, and Visualization-of-Thought \citep{wu2024visualization}, which may enable the AI to navigate complex reasoning sequences more effectively, potentially revealing deeper insights into its problem-solving abilities.
The fact that ChatGPT-4 Vision was able to reassess and correct its initial response to several questions during our experiment suggests that new prompting techniques tailored for the exam's format and style have the potential to further enhance the model's accuracy.
Second, our evaluation protocol involved asking the AI model to answer each question only once. This design choice limits our ability to analyze the consistency of the AI's performance or its variability in responding to the same question across multiple attempts. A more comprehensive evaluation could involve multiple attempts to answer each question, allowing for a more nuanced understanding of the AI's performance and the potential for improvement over time \citep{chen2023chatgpts}. 
Third, our study did not compare ChaGPT-4 Vision's ENADE 2021 BCS exam performance with other multimodal generative AI models like Google's recently released Gemini family of models \citep{team2023gemini}. A systematic comparative analysis involving other multimodal generative AI models could provide valuable insights into the unique capabilities and limitations of each model in handling educational content. This approach could also help identify areas for cross-model improvements and innovations \citep{geminivsgpt4v}.
Finally, our study did not incorporate traditional exam validation methods, such as pre-assessment moderation, where human experts review and validate exam questions to ensure their quality, fairness, and alignment with educational objectives~\cite{bloxham2016s}. Incorporating such traditional methods could provide a valuable benchmark against which the AI's performance can be measured. Future research could explore this comparison to better understand the potential and limitations of AI in educational assessments.

\section{Conclusion}\label{sec:conclusion}

ChatGPT-4 Vision's performance on Brazil's ENADE 2021 Bachelor in Computer Science exam offers a compelling glimpse into the future of generative AI technologies in educational assessment. ChatGPT-4 Vision not only matched but, in some cases, exceeded the top human scores, particularly in handling questions with visual components and complex problem-solving scenarios. While showcasing the model's impressive multimodal reasoning capabilities, the results particularly underline the need for improvements in the AI's handling of complex visual data interpretations and its ability to apply domain-specific knowledge accurately. Future research should focus on refining multimodal generative AI models to enhance their interpretive, reasoning, and vision capabilities. This includes exploring adaptive prompt techniques that dynamically adjust based on the AI's response quality, aiming to steer AI models more effectively through complex reasoning pathways. Additionally, employing AI models in simulated testing environments can help identify biases, ambiguities, or inconsistencies in exam questions, thereby enhancing their fairness and effectiveness. Successfully addressing these challenges will not only advance AI's capabilities in educational assessments but also pave the way for broader applications in personalized learning, automated tutoring systems, and interactive educational content.

\section{Acknowledgments}

The author is grateful to the following computer science researchers in Brazil who kindly agreed to participate in this study: Alfredo Goldman (University of São Paulo), Aline Marins Paes Carvalho (Federal Fluminense University), Altigran Soares da Silva (Federal University of Amazonas), Avelino Francisco Zorzo (Pontifical Catholic University of Rio Grande do Sul), Carla Taciana Lima Lourenco Silva Schuenemann (Federal University of Pernambuco), Carlos Alberto Kamienski (Federal University of ABC), Celso Alberto Saibel Santos (Federal University of Espírito Santo), Francisco Vilar Brasileiro (Federal University of Campina Grande), Itana Maria de Souza Gimenes (State University of Maringá), João Fernando Lima Alcântara (Federal University of Ceará), Leopoldo Motta Teixeira (Federal University of Pernambuco), Luiz Fernando Bittencourt (State University of Campinas), Nelson Souto Rosa (Federal University of Pernambuco), Ricardo Marcondes Marcacini (University of São Paulo), Raimundo da Silva Barreto (Federal University of Amazonas), and Tayana Uchôa Conte (Federal University of Amazonas).

The generative AI tools ChatGPT-4, GitHub Copilot, and Gemini Advanced were used to prepare this paper, including revising and improving several sections of the manuscript, creating and formatting tables and figures, and generating and executing code for data analysis.

\bibliographystyle{ACM-Reference-Format}
\bibliography{references}


\begin{thebibliography}{51}

%%% the Web address.

\ifx \showCODEN    \undefined \def \showCODEN     #1{\unskip}     \fi
\ifx \showDOI      \undefined \def \showDOI       #1{#1}\fi
\ifx \showISBNx    \undefined \def \showISBNx     #1{\unskip}     \fi
\ifx \showISBNxiii \undefined \def \showISBNxiii  #1{\unskip}     \fi
\ifx \showISSN     \undefined \def \showISSN      #1{\unskip}     \fi
\ifx \showLCCN     \undefined \def \showLCCN      #1{\unskip}     \fi
\ifx \shownote     \undefined \def \shownote      #1{#1}          \fi
\ifx \showarticletitle \undefined \def \showarticletitle #1{#1}   \fi
\ifx \showURL      \undefined \def \showURL       {\relax}        \fi
% invisible to TeX
\providecommand\bibfield[2]{#2}
\providecommand\bibinfo[2]{#2}
\providecommand\natexlab[1]{#1}
\providecommand\showeprint[2][]{arXiv:#2}

\bibitem[Almeida et~al\mbox{.}(2024)]%
        {almeida2024sabia}
\bibfield{author}{\bibinfo{person}{Thales~Sales Almeida}, \bibinfo{person}{Hugo
  Abonizio}, \bibinfo{person}{Rodrigo Nogueira}, {and} \bibinfo{person}{Ramon
  Pires}.} \bibinfo{year}{2024}\natexlab{}.
\newblock \showarticletitle{{Sabi{\'a}-2: A New Generation of Portuguese Large
  Language Models}}.
\newblock \bibinfo{journal}{\emph{arXiv preprint arXiv:2403.09887}}
  (\bibinfo{year}{2024}).
\newblock


\bibitem[Bloxham et~al\mbox{.}(2016)]%
        {bloxham2016s}
\bibfield{author}{\bibinfo{person}{Sue Bloxham}, \bibinfo{person}{Clair
  Hughes}, {and} \bibinfo{person}{Lenore Adie}.}
  \bibinfo{year}{2016}\natexlab{}.
\newblock \showarticletitle{{What's the point of moderation? A discussion of
  the purposes achieved through contemporary moderation practices}}.
\newblock \bibinfo{journal}{\emph{Assessment \& Evaluation in Higher
  Education}} \bibinfo{volume}{41}, \bibinfo{number}{4} (\bibinfo{year}{2016}),
  \bibinfo{pages}{638--653}.
\newblock


\bibitem[Bonett(2020)]%
        {bonett2020point}
\bibfield{author}{\bibinfo{person}{Douglas~G Bonett}.}
  \bibinfo{year}{2020}\natexlab{}.
\newblock \showarticletitle{Point-biserial correlation: Interval estimation,
  hypothesis testing, meta-analysis, and sample size determination}.
\newblock \bibinfo{journal}{\emph{Brit. J. Math. Statist. Psych.}}
  \bibinfo{volume}{73} (\bibinfo{year}{2020}), \bibinfo{pages}{113--144}.
\newblock


\bibitem[Bordt and von Luxburg(2023)]%
        {bordt2023chatgpt}
\bibfield{author}{\bibinfo{person}{Sebastian Bordt} {and}
  \bibinfo{person}{Ulrike von Luxburg}.} \bibinfo{year}{2023}\natexlab{}.
\newblock \showarticletitle{{ChatGPT Participates in a Computer Science Exam}}.
\newblock \bibinfo{journal}{\emph{arXiv preprint arXiv:2303.09461}}
  (\bibinfo{year}{2023}).
\newblock


\bibitem[Bubeck et~al\mbox{.}(2023)]%
        {bubeck2023sparks}
\bibfield{author}{\bibinfo{person}{S{\'e}bastien Bubeck},
  \bibinfo{person}{Varun Chandrasekaran}, \bibinfo{person}{Ronen Eldan},
  \bibinfo{person}{Johannes Gehrke}, \bibinfo{person}{Eric Horvitz},
  \bibinfo{person}{Ece Kamar}, \bibinfo{person}{Peter Lee},
  \bibinfo{person}{Yin~Tat Lee}, \bibinfo{person}{Yuanzhi Li},
  \bibinfo{person}{Scott Lundberg}, {et~al\mbox{.}}}
  \bibinfo{year}{2023}\natexlab{}.
\newblock \showarticletitle{{Sparks of Artificial General Intelligence: Early
  experiments with GPT-4}}.
\newblock \bibinfo{journal}{\emph{arXiv preprint arXiv:2303.12712}}
  (\bibinfo{year}{2023}).
\newblock


\bibitem[Chen et~al\mbox{.}(2023)]%
        {chen2023chatgpts}
\bibfield{author}{\bibinfo{person}{Lingjiao Chen}, \bibinfo{person}{Matei
  Zaharia}, {and} \bibinfo{person}{James Zou}.}
  \bibinfo{year}{2023}\natexlab{}.
\newblock \showarticletitle{{How is ChatGPT's behavior changing over time?}}
\newblock \bibinfo{journal}{\emph{arXiv preprint arXiv:2307.09009}}
  (\bibinfo{year}{2023}).
\newblock


\bibitem[Dai et~al\mbox{.}(2023)]%
        {dai2023can}
\bibfield{author}{\bibinfo{person}{Wei Dai}, \bibinfo{person}{Jionghao Lin},
  \bibinfo{person}{Hua Jin}, \bibinfo{person}{Tongguang Li},
  \bibinfo{person}{Yi-Shan Tsai}, \bibinfo{person}{Dragan Ga{\v{s}}evi{\'c}},
  {and} \bibinfo{person}{Guanliang Chen}.} \bibinfo{year}{2023}\natexlab{}.
\newblock \showarticletitle{{Can Large Language Models Provide Feedback to
  Students? A Case Study on ChatGPT}}. In \bibinfo{booktitle}{\emph{2023 IEEE
  International Conference on Advanced Learning Technologies (ICALT)}}. IEEE,
  \bibinfo{pages}{323--325}.
\newblock


\bibitem[Espejel et~al\mbox{.}(2023)]%
        {espejel2023gpt}
\bibfield{author}{\bibinfo{person}{Jessica~L{\'o}pez Espejel},
  \bibinfo{person}{El~Hassane Ettifouri}, \bibinfo{person}{Mahaman
  Sanoussi~Yahaya Alassan}, \bibinfo{person}{El~Mehdi Chouham}, {and}
  \bibinfo{person}{Walid Dahhane}.} \bibinfo{year}{2023}\natexlab{}.
\newblock \showarticletitle{{GPT-3.5, GPT-4, or BARD? Evaluating LLMs reasoning
  ability in zero-shot setting and performance boosting through prompts}}.
\newblock \bibinfo{journal}{\emph{Natural Language Processing Journal}}
  \bibinfo{volume}{5} (\bibinfo{year}{2023}), \bibinfo{pages}{100032}.
\newblock


\bibitem[Evagorou et~al\mbox{.}(2015)]%
        {evagorou2015role}
\bibfield{author}{\bibinfo{person}{Maria Evagorou}, \bibinfo{person}{Sibel
  Erduran}, {and} \bibinfo{person}{Terhi M{\"a}ntyl{\"a}}.}
  \bibinfo{year}{2015}\natexlab{}.
\newblock \showarticletitle{The role of visual representations in scientific
  practices: from conceptual understanding and knowledge generation to 'seeing'
  how science works}.
\newblock \bibinfo{journal}{\emph{International journal of STEM education}}
  \bibinfo{volume}{2}, \bibinfo{number}{1} (\bibinfo{year}{2015}),
  \bibinfo{pages}{1--13}.
\newblock


\bibitem[Frenkel and Emara(2023)]%
        {frenkel2023chatgpt}
\bibfield{author}{\bibinfo{person}{Matthew Frenkel} {and}
  \bibinfo{person}{Hebah Emara}.} \bibinfo{year}{2023}\natexlab{}.
\newblock \showarticletitle{{ChatGPT \& Mechanical Engineering: Examining
  performance on the FE Mechanical Engineering and Undergraduate Exams}}.
\newblock \bibinfo{journal}{\emph{arXiv preprint arXiv:2309.15866}}
  (\bibinfo{year}{2023}).
\newblock


\bibitem[{Gemini Team}(2023)]%
        {team2023gemini}
\bibfield{author}{\bibinfo{person}{{Gemini Team}}.}
  \bibinfo{year}{2023}\natexlab{}.
\newblock \showarticletitle{{Gemini: A Family of Highly Capable Multimodal
  Models}}.
\newblock \bibinfo{journal}{\emph{arXiv preprint arXiv:2312.11805}}
  (\bibinfo{year}{2023}).
\newblock


\bibitem[Google(2023)]%
        {gemini}
\bibfield{author}{\bibinfo{person}{Google}.} \bibinfo{year}{2023}\natexlab{}.
\newblock \bibinfo{title}{{Gemini}}.
\newblock \bibinfo{howpublished}{\url{https://gemini.google.com/}}.
\newblock


\bibitem[{Inep}(2022)]%
        {enade-relatorio-sintese-cc}
\bibfield{author}{\bibinfo{person}{{Inep}}.} \bibinfo{year}{2022}\natexlab{}.
\newblock \bibinfo{title}{{Relatório Síntese de Área: Ciência da
  Computação (Bacharelado / Licencitatura)}}.
\newblock
  \bibinfo{howpublished}{\url{https://download.inep.gov.br/educacao_superior/enade/relatorio_sintese/2021/Enade_2021_Relatorios_Sintese_Area_Ciencia_Computacao.pdf}}.
\newblock


\bibitem[{Inep}(2023)]%
        {enade}
\bibfield{author}{\bibinfo{person}{{Inep}}.} \bibinfo{year}{2023}\natexlab{}.
\newblock \bibinfo{title}{{Exame Nacional de Desempenho dos Estudantes
  (ENADE)}}.
\newblock
  \bibinfo{howpublished}{\url{https://www.gov.br/inep/pt-br/areas-de-atuacao/avaliacao-e-exames-educacionais/enade}}.
\newblock


\bibitem[Jalil et~al\mbox{.}(2023)]%
        {jalil2023chatgpt}
\bibfield{author}{\bibinfo{person}{Sajed Jalil}, \bibinfo{person}{Suzzana
  Rafi}, \bibinfo{person}{Thomas~D LaToza}, \bibinfo{person}{Kevin Moran},
  {and} \bibinfo{person}{Wing Lam}.} \bibinfo{year}{2023}\natexlab{}.
\newblock \showarticletitle{{ChatGPT and Software Testing Education: Promises
  \& Perils}}. In \bibinfo{booktitle}{\emph{2023 IEEE International Conference
  on Software Testing, Verification and Validation Workshops (ICSTW)}}. IEEE,
  \bibinfo{pages}{4130--4137}.
\newblock


\bibitem[Joshi et~al\mbox{.}(2024)]%
        {joshi2024chatgpt}
\bibfield{author}{\bibinfo{person}{Ishika Joshi}, \bibinfo{person}{Ritvik
  Budhiraja}, \bibinfo{person}{Harshal Dev}, \bibinfo{person}{Jahnvi Kadia},
  \bibinfo{person}{Mohammad~Osama Ataullah}, \bibinfo{person}{Sayan Mitra},
  \bibinfo{person}{Harshal~D Akolekar}, {and} \bibinfo{person}{Dhruv Kumar}.}
  \bibinfo{year}{2024}\natexlab{}.
\newblock \showarticletitle{{ChatGPT in the Classroom: An Analysis of Its
  Strengths and Weaknesses for Solving Undergraduate Computer Science
  Questions}}. In \bibinfo{booktitle}{\emph{55th ACM Technical Symposium on
  Computer Science Education (SIGCSE) V. 1}}. \bibinfo{pages}{625--631}.
\newblock


\bibitem[Joshi et~al\mbox{.}(2023)]%
        {joshi2023chatgpt}
\bibfield{author}{\bibinfo{person}{Ishika Joshi}, \bibinfo{person}{Ritvik
  Budhiraja}, \bibinfo{person}{Harshal Dev}, \bibinfo{person}{Jahnvi Kadia},
  \bibinfo{person}{M~Osama Ataullah}, \bibinfo{person}{Sayan Mitra},
  \bibinfo{person}{Dhruv Kumar}, {and} \bibinfo{person}{Harshal~D Akolekar}.}
  \bibinfo{year}{2023}\natexlab{}.
\newblock \showarticletitle{{ChatGPT in the Classroom: An Analysis of Its
  Strengths and Weaknesses for Solving Undergraduate Computer Science
  Questions}}.
\newblock \bibinfo{journal}{\emph{arXiv preprint arXiv:2304.14993}}
  (\bibinfo{year}{2023}).
\newblock


\bibitem[Kasneci et~al\mbox{.}(2023)]%
        {kasneci2023chatgpt}
\bibfield{author}{\bibinfo{person}{Enkelejda Kasneci}, \bibinfo{person}{Kathrin
  Se{\ss}ler}, \bibinfo{person}{Stefan K{\"u}chemann}, \bibinfo{person}{Maria
  Bannert}, \bibinfo{person}{Daryna Dementieva}, \bibinfo{person}{Frank
  Fischer}, \bibinfo{person}{Urs Gasser}, \bibinfo{person}{Georg Groh},
  \bibinfo{person}{Stephan G{\"u}nnemann}, \bibinfo{person}{Eyke
  H{\"u}llermeier}, {et~al\mbox{.}}} \bibinfo{year}{2023}\natexlab{}.
\newblock \showarticletitle{{ChatGPT for good? On opportunities and challenges
  of large language models for education}}.
\newblock \bibinfo{journal}{\emph{Learning and Individual Differences}}
  \bibinfo{volume}{103} (\bibinfo{year}{2023}), \bibinfo{pages}{102274}.
\newblock


\bibitem[Lewis et~al\mbox{.}(2020)]%
        {lewis2020retrieval}
\bibfield{author}{\bibinfo{person}{Patrick Lewis}, \bibinfo{person}{Ethan
  Perez}, \bibinfo{person}{Aleksandra Piktus}, \bibinfo{person}{Fabio Petroni},
  \bibinfo{person}{Vladimir Karpukhin}, \bibinfo{person}{Naman Goyal},
  \bibinfo{person}{Heinrich K{\"u}ttler}, \bibinfo{person}{Mike Lewis},
  \bibinfo{person}{Wen-tau Yih}, \bibinfo{person}{Tim Rockt{\"a}schel},
  {et~al\mbox{.}}} \bibinfo{year}{2020}\natexlab{}.
\newblock \showarticletitle{{Retrieval-Augmented Generation for
  Knowledge-Intensive NLP Tasks}}.
\newblock \bibinfo{journal}{\emph{Advances in Neural Information Processing
  Systems (NeurIPS 2020)}}  \bibinfo{volume}{33} (\bibinfo{year}{2020}),
  \bibinfo{pages}{9459--9474}.
\newblock


\bibitem[Li et~al\mbox{.}(2023)]%
        {li2023adapting}
\bibfield{author}{\bibinfo{person}{Qingyao Li}, \bibinfo{person}{Lingyue Fu},
  \bibinfo{person}{Weiming Zhang}, \bibinfo{person}{Xianyu Chen},
  \bibinfo{person}{Jingwei Yu}, \bibinfo{person}{Wei Xia},
  \bibinfo{person}{Weinan Zhang}, \bibinfo{person}{Ruiming Tang}, {and}
  \bibinfo{person}{Yong Yu}.} \bibinfo{year}{2023}\natexlab{}.
\newblock \showarticletitle{{Adapting Large Language Models for Education:
  Foundational Capabilities, Potentials, and Challenges}}.
\newblock \bibinfo{journal}{\emph{arXiv preprint arXiv:2401.08664}}
  (\bibinfo{year}{2023}).
\newblock


\bibitem[Liu et~al\mbox{.}(2023)]%
        {liu2023evaluating}
\bibfield{author}{\bibinfo{person}{Hanmeng Liu}, \bibinfo{person}{Ruoxi Ning},
  \bibinfo{person}{Zhiyang Teng}, \bibinfo{person}{Jian Liu},
  \bibinfo{person}{Qiji Zhou}, {and} \bibinfo{person}{Yue Zhang}.}
  \bibinfo{year}{2023}\natexlab{}.
\newblock \showarticletitle{{Evaluating the Logical Reasoning Ability of
  ChatGPT and GPT-4}}.
\newblock \bibinfo{journal}{\emph{arXiv preprint arXiv:2304.03439}}
  (\bibinfo{year}{2023}).
\newblock


\bibitem[Mahon et~al\mbox{.}(2023)]%
        {mahon2023no}
\bibfield{author}{\bibinfo{person}{Joyce Mahon}, \bibinfo{person}{Brian
  Mac~Namee}, {and} \bibinfo{person}{Brett~A Becker}.}
  \bibinfo{year}{2023}\natexlab{}.
\newblock \showarticletitle{{No More Pencils No More Books: Capabilities of
  Generative AI on Irish and UK Computer Science School Leaving Examinations}}.
  In \bibinfo{booktitle}{\emph{2023 Conference on United Kingdom \& Ireland
  Computing Education Research (UKICER)}}. \bibinfo{pages}{1--7}.
\newblock


\bibitem[Mao et~al\mbox{.}(2023)]%
        {mao2023gpteval}
\bibfield{author}{\bibinfo{person}{Rui Mao}, \bibinfo{person}{Guanyi Chen},
  \bibinfo{person}{Xulang Zhang}, \bibinfo{person}{Frank Guerin}, {and}
  \bibinfo{person}{Erik Cambria}.} \bibinfo{year}{2023}\natexlab{}.
\newblock \showarticletitle{{GPTEval: A Survey on Assessments of ChatGPT and
  GPT-4}}.
\newblock \bibinfo{journal}{\emph{arXiv preprint arXiv:2308.12488}}
  (\bibinfo{year}{2023}).
\newblock


\bibitem[Mao et~al\mbox{.}(2024)]%
        {mao2024gpteval}
\bibfield{author}{\bibinfo{person}{Rui Mao}, \bibinfo{person}{Guanyi Chen},
  \bibinfo{person}{Xulang Zhang}, \bibinfo{person}{Frank Guerin}, {and}
  \bibinfo{person}{Erik Cambria}.} \bibinfo{year}{2024}\natexlab{}.
\newblock \showarticletitle{{GPTEval: A Survey on Assessments of ChatGPT and
  GPT-4}}. In \bibinfo{booktitle}{\emph{2024 Joint International Conference on
  Computational Linguistics, Language Resources and Evaluation (LREC-COLING)}}.
\newblock


\bibitem[Mayne(2024)]%
        {mayne2024-improvingGPT}
\bibfield{author}{\bibinfo{person}{Andrew Mayne}.}
  \bibinfo{year}{2024}\natexlab{}.
\newblock \bibinfo{title}{{Improving GPT-4's Visual Reasoning with Prompting}}.
\newblock
  \bibinfo{howpublished}{\url{https://andrewmayne.com/2024/03/12/improving-gpt-4s-visual-reasoning-with-prompting/}}.
\newblock


\bibitem[Meyer et~al\mbox{.}(2023)]%
        {meyer2023chatgpt}
\bibfield{author}{\bibinfo{person}{Jesse~G Meyer}, \bibinfo{person}{Ryan~J
  Urbanowicz}, \bibinfo{person}{Patrick~CN Martin}, \bibinfo{person}{Karen
  O'Connor}, \bibinfo{person}{Ruowang Li}, \bibinfo{person}{Pei-Chen Peng},
  \bibinfo{person}{Tiffani~J Bright}, \bibinfo{person}{Nicholas Tatonetti},
  \bibinfo{person}{Kyoung~Jae Won}, \bibinfo{person}{Graciela
  Gonzalez-Hernandez}, {et~al\mbox{.}}} \bibinfo{year}{2023}\natexlab{}.
\newblock \showarticletitle{ChatGPT and large language models in academia:
  opportunities and challenges}.
\newblock \bibinfo{journal}{\emph{BioData Mining}} \bibinfo{volume}{16},
  \bibinfo{number}{1} (\bibinfo{year}{2023}), \bibinfo{pages}{20}.
\newblock


\bibitem[Nori et~al\mbox{.}(2023)]%
        {nori2023capabilities}
\bibfield{author}{\bibinfo{person}{Harsha Nori}, \bibinfo{person}{Nicholas
  King}, \bibinfo{person}{Scott~Mayer McKinney}, \bibinfo{person}{Dean
  Carignan}, {and} \bibinfo{person}{Eric Horvitz}.}
  \bibinfo{year}{2023}\natexlab{}.
\newblock \showarticletitle{{Capabilities of GPT-4 on Medical Challenge
  Problems}}.
\newblock \bibinfo{journal}{\emph{arXiv preprint arXiv:2303.13375}}
  (\bibinfo{year}{2023}).
\newblock


\bibitem[Nunes et~al\mbox{.}(2023)]%
        {nunes2023evaluating}
\bibfield{author}{\bibinfo{person}{Desnes Nunes}, \bibinfo{person}{Ricardo
  Primi}, \bibinfo{person}{Ramon Pires}, \bibinfo{person}{Roberto Lotufo},
  {and} \bibinfo{person}{Rodrigo Nogueira}.} \bibinfo{year}{2023}\natexlab{}.
\newblock \showarticletitle{{Evaluating GPT-3.5 and GPT-4 Models on Brazilian
  University Admission Exams}}.
\newblock \bibinfo{journal}{\emph{arXiv preprint arXiv:2303.17003}}
  (\bibinfo{year}{2023}).
\newblock


\bibitem[OpenAI(2023a)]%
        {chatgpt}
\bibfield{author}{\bibinfo{person}{OpenAI}.} \bibinfo{year}{2023}\natexlab{a}.
\newblock \bibinfo{title}{{ChatGPT}}.
\newblock \bibinfo{howpublished}{\url{https://chat.openai.com/}}.
\newblock


\bibitem[OpenAI(2023b)]%
        {openai2023chatgpt}
\bibfield{author}{\bibinfo{person}{OpenAI}.} \bibinfo{year}{2023}\natexlab{b}.
\newblock \bibinfo{title}{{ChatGPT can now see, hear, and speak}}.
\newblock
  \bibinfo{howpublished}{\url{https://openai.com/blog/chatgpt-can-now-see-hear-and-speak}}.
\newblock


\bibitem[OpenAI(2023c)]%
        {openai2023gpt4v}
\bibfield{author}{\bibinfo{person}{OpenAI}.} \bibinfo{year}{2023}\natexlab{c}.
\newblock \bibinfo{title}{{GPT-4V(ision) system card}}.
\newblock
  \bibinfo{howpublished}{\url{https://cdn.openai.com/papers/GPTV_System_Card.pdf}}.
\newblock


\bibitem[OpenAI(2023d)]%
        {openai2023chatgpt-turbo}
\bibfield{author}{\bibinfo{person}{OpenAI}.} \bibinfo{year}{2023}\natexlab{d}.
\newblock \bibinfo{title}{{New models and developer products announced at
  DevDay}}.
\newblock
  \bibinfo{howpublished}{\url{https://openai.com/blog/new-models-and-developer-products-announced-at-devday}}.
\newblock


\bibitem[Pires et~al\mbox{.}(2023)]%
        {pires2023evaluating}
\bibfield{author}{\bibinfo{person}{Ramon Pires}, \bibinfo{person}{Thales~Sales
  Almeida}, \bibinfo{person}{Hugo Abonizio}, {and} \bibinfo{person}{Rodrigo
  Nogueira}.} \bibinfo{year}{2023}\natexlab{}.
\newblock \showarticletitle{{Evaluating GPT-4's Vision Capabilities on
  Brazilian University Admission Exams}}.
\newblock \bibinfo{journal}{\emph{arXiv preprint arXiv:2311.14169}}
  (\bibinfo{year}{2023}).
\newblock


\bibitem[Pursnani et~al\mbox{.}(2023)]%
        {pursnani2023performance}
\bibfield{author}{\bibinfo{person}{Vinay Pursnani}, \bibinfo{person}{Yusuf
  Sermet}, \bibinfo{person}{Musa Kurt}, {and} \bibinfo{person}{Ibrahim Demir}.}
  \bibinfo{year}{2023}\natexlab{}.
\newblock \showarticletitle{{Performance of ChatGPT on the US fundamentals of
  engineering exam: Comprehensive assessment of proficiency and potential
  implications for professional environmental engineering practice}}.
\newblock \bibinfo{journal}{\emph{Computers and Education: Artificial
  Intelligence}}  \bibinfo{volume}{5} (\bibinfo{year}{2023}),
  \bibinfo{pages}{100183}.
\newblock


\bibitem[Qi et~al\mbox{.}(2023)]%
        {geminivsgpt4v}
\bibfield{author}{\bibinfo{person}{Zhangyang Qi}, \bibinfo{person}{Ye Fang},
  \bibinfo{person}{Mengchen Zhang}, \bibinfo{person}{Zeyi Sun},
  \bibinfo{person}{Tong Wu}, \bibinfo{person}{Ziwei Liu},
  \bibinfo{person}{Dahua Lin}, \bibinfo{person}{Jiaqi Wang}, {and}
  \bibinfo{person}{Hengshuang Zhao}.} \bibinfo{year}{2023}\natexlab{}.
\newblock \showarticletitle{{Gemini vs GPT-4V: A Preliminary Comparison and
  Combination of Vision-Language Models Through Qualitative Cases}}.
\newblock \bibinfo{journal}{\emph{arXiv preprint arXiv:2312.15011}}
  (\bibinfo{year}{2023}).
\newblock


\bibitem[Rezigalla(2022)]%
        {rezigalla2022item}
\bibfield{author}{\bibinfo{person}{Assad~Ali Rezigalla}.}
  \bibinfo{year}{2022}\natexlab{}.
\newblock \showarticletitle{{Item Analysis: Concept and Application}}.
\newblock \bibinfo{journal}{\emph{Medical Education for the 21st Century}}
  (\bibinfo{year}{2022}), \bibinfo{pages}{105--120}.
\newblock


\bibitem[Richards et~al\mbox{.}(2024)]%
        {richards2024bob}
\bibfield{author}{\bibinfo{person}{Mike Richards}, \bibinfo{person}{Kevin
  Waugh}, \bibinfo{person}{Mark Slaymaker}, \bibinfo{person}{Marian Petre},
  \bibinfo{person}{John Woodthorpe}, {and} \bibinfo{person}{Daniel Gooch}.}
  \bibinfo{year}{2024}\natexlab{}.
\newblock \showarticletitle{{Bob or Bot: Exploring ChatGPT's Answers to
  University Computer Science Assessment}}.
\newblock \bibinfo{journal}{\emph{ACM Transactions on Computing Education}}
  \bibinfo{volume}{24}, \bibinfo{number}{1} (\bibinfo{year}{2024}),
  \bibinfo{pages}{1--32}.
\newblock


\bibitem[Santos and Campelo(2023)]%
        {santos2023benchmarking-cbic}
\bibfield{author}{\bibinfo{person}{Matheus L.~O. Santos} {and}
  \bibinfo{person}{Cl{\'a}udio E.~C. Campelo}.}
  \bibinfo{year}{2023}\natexlab{}.
\newblock \showarticletitle{{Benchmarking quantized LLaMa-based models on the
  Brazilian Secondary School Exam}}. In \bibinfo{booktitle}{\emph{16th
  Brazilian Congress on Computational Intelligence ({CBIC})}}.
  \bibinfo{pages}{1--8}.
\newblock


\bibitem[Toyama et~al\mbox{.}(2023)]%
        {toyama2023performance}
\bibfield{author}{\bibinfo{person}{Yoshitaka Toyama}, \bibinfo{person}{Ayaka
  Harigai}, \bibinfo{person}{Mirei Abe}, \bibinfo{person}{Mitsutoshi Nagano},
  \bibinfo{person}{Masahiro Kawabata}, \bibinfo{person}{Yasuhiro Seki}, {and}
  \bibinfo{person}{Kei Takase}.} \bibinfo{year}{2023}\natexlab{}.
\newblock \showarticletitle{{Performance evaluation of ChatGPT, GPT-4, and Bard
  on the official board examination of the Japan Radiology Society}}.
\newblock \bibinfo{journal}{\emph{Japanese Journal of Radiology}}
  (\bibinfo{year}{2023}), \bibinfo{pages}{1--7}.
\newblock


\bibitem[Tyen et~al\mbox{.}(2024)]%
        {tyen2024llms}
\bibfield{author}{\bibinfo{person}{Gladys Tyen}, \bibinfo{person}{Hassan
  Mansoor}, \bibinfo{person}{Victor C\u{a}rbune}, \bibinfo{person}{Peter Chen},
  {and} \bibinfo{person}{Tony Mak}.} \bibinfo{year}{2024}\natexlab{}.
\newblock \showarticletitle{{LLMs cannot \textit{find} reasoning errors, but
  can \textit{correct} them!}}
\newblock \bibinfo{journal}{\emph{arXiv preprint arXiv:2311.08516}}
  (\bibinfo{year}{2024}).
\newblock


\bibitem[Wang et~al\mbox{.}(2023)]%
        {wang2023selfconsistency}
\bibfield{author}{\bibinfo{person}{Xuezhi Wang}, \bibinfo{person}{Jason Wei},
  \bibinfo{person}{Dale Schuurmans}, \bibinfo{person}{Quoc Le},
  \bibinfo{person}{Ed Chi}, \bibinfo{person}{Sharan Narang},
  \bibinfo{person}{Aakanksha Chowdhery}, {and} \bibinfo{person}{Denny Zhou}.}
  \bibinfo{year}{2023}\natexlab{}.
\newblock \showarticletitle{{Self-Consistency Improves Chain of Thought
  Reasoning in Language Models}}.
\newblock \bibinfo{journal}{\emph{arXiv preprint arXiv:2203.11171}}
  (\bibinfo{year}{2023}).
\newblock


\bibitem[Wei et~al\mbox{.}(2022)]%
        {wei2022chain}
\bibfield{author}{\bibinfo{person}{Jason Wei}, \bibinfo{person}{Xuezhi Wang},
  \bibinfo{person}{Dale Schuurmans}, \bibinfo{person}{Maarten Bosma},
  \bibinfo{person}{Fei Xia}, \bibinfo{person}{Ed Chi}, \bibinfo{person}{Quoc~V
  Le}, \bibinfo{person}{Denny Zhou}, {et~al\mbox{.}}}
  \bibinfo{year}{2022}\natexlab{}.
\newblock \showarticletitle{{Chain-of-Thought Prompting Elicits Reasoning in
  Large Language Models}}.
\newblock \bibinfo{journal}{\emph{Advances in Neural Information Processing
  Systems (NeurIPS 2022)}}  \bibinfo{volume}{35} (\bibinfo{year}{2022}),
  \bibinfo{pages}{24824--24837}.
\newblock


\bibitem[Wu et~al\mbox{.}(2024)]%
        {wu2024visualization}
\bibfield{author}{\bibinfo{person}{Wenshan Wu}, \bibinfo{person}{Shaoguang
  Mao}, \bibinfo{person}{Yadong Zhang}, \bibinfo{person}{Yan Xia},
  \bibinfo{person}{Li Dong}, \bibinfo{person}{Lei Cui}, {and}
  \bibinfo{person}{Furu Wei}.} \bibinfo{year}{2024}\natexlab{}.
\newblock \showarticletitle{{Visualization-of-Thought Elicits Spatial Reasoning
  in Large Language Models}}.
\newblock \bibinfo{journal}{\emph{arXiv preprint arXiv:2404.03622}}
  (\bibinfo{year}{2024}).
\newblock


\bibitem[Yao et~al\mbox{.}(2023)]%
        {yao2023tree}
\bibfield{author}{\bibinfo{person}{Shunyu Yao}, \bibinfo{person}{Dian Yu},
  \bibinfo{person}{Jeffrey Zhao}, \bibinfo{person}{Izhak Shafran},
  \bibinfo{person}{Thomas~L. Griffiths}, \bibinfo{person}{Yuan Cao}, {and}
  \bibinfo{person}{Karthik Narasimhan}.} \bibinfo{year}{2023}\natexlab{}.
\newblock \showarticletitle{{Tree of Thoughts: Deliberate Problem Solving with
  Large Language Models}}.
\newblock \bibinfo{journal}{\emph{arXiv preprint arXiv:2305.10601}}
  (\bibinfo{year}{2023}).
\newblock


\bibitem[Zar(2005)]%
        {zar2005spearman}
\bibfield{author}{\bibinfo{person}{Jerrold~H Zar}.}
  \bibinfo{year}{2005}\natexlab{}.
\newblock \showarticletitle{{Spearman Rank Correlation}}.
\newblock \bibinfo{journal}{\emph{Encyclopedia of Biostatistics}}
  \bibinfo{volume}{7} (\bibinfo{year}{2005}).
\newblock


\bibitem[Zhai et~al\mbox{.}(2024)]%
        {zhai2024generative}
\bibfield{author}{\bibinfo{person}{Xiaoming Zhai}, \bibinfo{person}{Matthew
  Nyaaba}, {and} \bibinfo{person}{Wenchao Ma}.}
  \bibinfo{year}{2024}\natexlab{}.
\newblock \showarticletitle{{Can Generative AI and ChatGPT Outperform Humans on
  Cognitive-demanding Problem-Solving Tasks in Science?}}
\newblock \bibinfo{journal}{\emph{arXiv preprint arXiv:2401.15081}}
  (\bibinfo{year}{2024}).
\newblock


\bibitem[Zhang et~al\mbox{.}(2024b)]%
        {zhang2024mathverse}
\bibfield{author}{\bibinfo{person}{Renrui Zhang}, \bibinfo{person}{Dongzhi
  Jiang}, \bibinfo{person}{Yichi Zhang}, \bibinfo{person}{Haokun Lin},
  \bibinfo{person}{Ziyu Guo}, \bibinfo{person}{Pengshuo Qiu},
  \bibinfo{person}{Aojun Zhou}, \bibinfo{person}{Pan Lu},
  \bibinfo{person}{Kai-Wei Chang}, \bibinfo{person}{Peng Gao}, {et~al\mbox{.}}}
  \bibinfo{year}{2024}\natexlab{b}.
\newblock \showarticletitle{{MathVerse: Does Your Multi-modal LLM Truly See the
  Diagrams in Visual Math Problems?}}
\newblock \bibinfo{journal}{\emph{arXiv preprint arXiv:2403.14624}}
  (\bibinfo{year}{2024}).
\newblock


\bibitem[Zhang et~al\mbox{.}(2024a)]%
        {zhang2024far}
\bibfield{author}{\bibinfo{person}{Yizhe Zhang}, \bibinfo{person}{He Bai},
  \bibinfo{person}{Ruixiang Zhang}, \bibinfo{person}{Jiatao Gu},
  \bibinfo{person}{Shuangfei Zhai}, \bibinfo{person}{Josh Susskind}, {and}
  \bibinfo{person}{Navdeep Jaitly}.} \bibinfo{year}{2024}\natexlab{a}.
\newblock \showarticletitle{{How Far Are We from Intelligent Visual Deductive
  Reasoning?}}
\newblock \bibinfo{journal}{\emph{arXiv preprint arXiv:2403.04732}}
  (\bibinfo{year}{2024}).
\newblock


\bibitem[Zhao et~al\mbox{.}(2023a)]%
        {zhao2023mmicl}
\bibfield{author}{\bibinfo{person}{Haozhe Zhao}, \bibinfo{person}{Zefan Cai},
  \bibinfo{person}{Shuzheng Si}, \bibinfo{person}{Xiaojian Ma},
  \bibinfo{person}{Kaikai An}, \bibinfo{person}{Liang Chen},
  \bibinfo{person}{Zixuan Liu}, \bibinfo{person}{Sheng Wang},
  \bibinfo{person}{Wenjuan Han}, {and} \bibinfo{person}{Baobao Chang}.}
  \bibinfo{year}{2023}\natexlab{a}.
\newblock \showarticletitle{{MMICL: Empowering Vision-language Model with
  Multi-Modal In-Context Learning}}.
\newblock \bibinfo{journal}{\emph{arXiv preprint arXiv:2309.07915}}
  (\bibinfo{year}{2023}).
\newblock


\bibitem[Zhao et~al\mbox{.}(2023b)]%
        {arxiv2023surveyLLMs}
\bibfield{author}{\bibinfo{person}{Wayne~Xin Zhao}, \bibinfo{person}{Kun Zhou},
  \bibinfo{person}{Junyi Li}, \bibinfo{person}{Tianyi Tang},
  \bibinfo{person}{Xiaolei Wang}, \bibinfo{person}{Yupeng Hou},
  \bibinfo{person}{Yingqian Min}, \bibinfo{person}{Beichen Zhang},
  \bibinfo{person}{Junjie Zhang}, \bibinfo{person}{Zican Dong},
  \bibinfo{person}{Yifan Du}, \bibinfo{person}{Chen Yang},
  \bibinfo{person}{Yushuo Chen}, \bibinfo{person}{Zhipeng Chen},
  \bibinfo{person}{Jinhao Jiang}, \bibinfo{person}{Ruiyang Ren},
  \bibinfo{person}{Yifan Li}, \bibinfo{person}{Xinyu Tang},
  \bibinfo{person}{Zikang Liu}, \bibinfo{person}{Peiyu Liu},
  \bibinfo{person}{Jian-Yun Nie}, {and} \bibinfo{person}{Ji-Rong Wen}.}
  \bibinfo{year}{2023}\natexlab{b}.
\newblock \showarticletitle{{A Survey of Large Language Models}}.
\newblock \bibinfo{journal}{\emph{arXiv preprint arXiv:2303.18223}}
  (\bibinfo{year}{2023}).
\newblock


\bibitem[Zhou et~al\mbox{.}(2024)]%
        {zhou2024the}
\bibfield{author}{\bibinfo{person}{Kyrie~Zhixuan Zhou},
  \bibinfo{person}{Zachary Kilhoffer}, \bibinfo{person}{Madelyn~Rose
  Sanfilippo}, \bibinfo{person}{Ted Underwood}, \bibinfo{person}{Ece Gumusel},
  \bibinfo{person}{Mengyi Wei}, \bibinfo{person}{Abhinav Choudhry}, {and}
  \bibinfo{person}{Jinjun Xiong}.} \bibinfo{year}{2024}\natexlab{}.
\newblock \showarticletitle{{``The teachers are confused as well'': A
  Multiple-Stakeholder Ethics Discussion on Large Language Models in Computing
  Education}}.
\newblock \bibinfo{journal}{\emph{arXiv preprint arXiv:2401.12453}}
  (\bibinfo{year}{2024}).
\newblock


\end{thebibliography}

\clearpage
\appendix

\section{Prompt Templates}
\label{app:prompt-templates}

The following are the prompt templates used to guide ChatGPT-4 Vision and ChatGPT-4 Turbo's responses during the ENADE 2021 BCS question-answering session. The templates are organized into three categories: initial prompts, reassessment prompts, and reflective prompts. Initial prompts are used to guide the model's first response to each question. Reassessment prompts are used to guide the model's self-reflective response when presented with a differing response standard or answer key. Finally, reflective prompts are used to guide ChatGPT-4 Turbo's analysis of ChatGPT-4 Vision's main challenges and difficulties in answering the questions.

\vspace{\parskip}
% \begin{multicols}{2}
\begin{modelinputbox}[title={Initial prompts}]
	\textbf{For open questions}: Answer the question in this image. Explain your reasoning step by step. Inform if the question is unclear or has no possible answer.\\

	[question image]\\

	\textbf{For multiple-choice questions}: Answer the question in this image. Explain your reasoning step by step. At the end, choose which of the five alternatives (A-E) is correct. Inform if there is more than one correct alternative or if all five alternatives are incorrect.\\

	[question image]
\end{modelinputbox}
\begin{modelinputbox}[title={Reassessment prompts}]
	\textbf{For open questions}: Your response is not entirely consistent with the response standard released for this question:\\

	[response standard]\\

	After analyzing the response standard, do you maintain or change your original response? If you decide to change your response, explain the possible reasons that led you to errors in the previous response. Try to identify the concrete causes of the errors instead of listing generic reasons, such as a failure to understand the question statement.\\

	\textbf{For multiple-choice questions}: Your response differs from the answer key released for this question: \\
	
	[answer key]\\

	After analyzing the answer key, do you maintain or change your response? If you decide to change your response, explain the possible reasons that led you to errors in the previous answer. Try to identify the concrete causes of the errors instead of listing generic reasons, such as a failure to understand the question statement.
\end{modelinputbox}
% \newcolumn
\begin{modelinputbox}[title={Reflective prompts (for ChatGPT-4 Turbo)}]
	The following are the prompts and responses of a generative AI for the attached question, along with possible assessments by two or more human experts on the generative AI's responses and the general quality of the question. Based on the provided information, summarize the main difficulties faced by the generative AI in attempting to answer the question, including possible difficulties resulting from incorrect recognition of some graphical or textual element present in the statement of the question. Consider only concrete difficulties whose manifestation can be evidenced by the information provided in generative AI's responses. At the end, discuss if these difficulties could be mitigated or eliminated by using prompt engineering techniques or if they would be inherent to the limitations of current large language models (LLMs).\\

	[question image]\\

	[ChatGPT-4 Vision's prompts and responses] \\

	[expert assessments (when available)]
\end{modelinputbox}	
% \end{multicols}
\vspace{\parskip}

Note that the main difference between the initial prompts for the two types of questions is that multiple-choice prompts explicitly request the model to select the correct answer from the five given alternatives. 
% In contrast, open-question prompts do not include this request, allowing the model to choose any appropriate answer. 
In both cases, the prompts ask the model to explain its reasoning and inform if the question is unclear or has multiple or no possible correct answers.

\vfill
\clearpage

\section{ChatGPT-4's Full Conversations and Experts' Assessments}
\label{app:conversations}

The following are ChatGPT-4's full conversations and the human experts' assessments---when available---for the three open questions and a representative sample of six valid multiple-choice questions.
% \footnote{The conversations and expert assessments for the remaining multiple-choice questions are available in the paper's companion repository.}

\subsection{Open Questions}
\label{app:open-questions}

% The following subsections present the ChatGPT-4 Vision's prompts and responses and expert evaluation and scoring for the three open questions of the ENADE 2021 BCS exam. 

% The following are the evaluation artifacts for the three open questions of the ENADE 2021 BCS exam. 

\subsubsection*{\myhref[Question 03]{03}}

This question asks the respondent to analyze a truth table for a set of formulas of propositional logic and answer four questions about the formulas' properties. ChatGPT-4 Vision gave correct answers to the first and fourth questions, a partially correct answer to the second question, and an incorrect answer to the third question. Upon being confronted with the response standard for the question, the model admitted it did a rush analysis of the truth table and confused the labels of some columns. The model also failed to cross-validate information between the formulas and the truth table, leading to the incorrect evaluation of one of the formulas. This question shows the model's understanding of complex mathematical concepts such as propositional logic and its limitations in correctly identifying and reasoning about structured visual information such as tables. 
% An expert's assessment of the model's initial response gave it 6.25 scores out of 10.
% \twocolumn

\vspace{\parskip}

\begin{questionbox}[title={Question}]
	% \textbf{Question image} \\[2pt]
	\includegraphics[width=\imagewidth,valign=t]{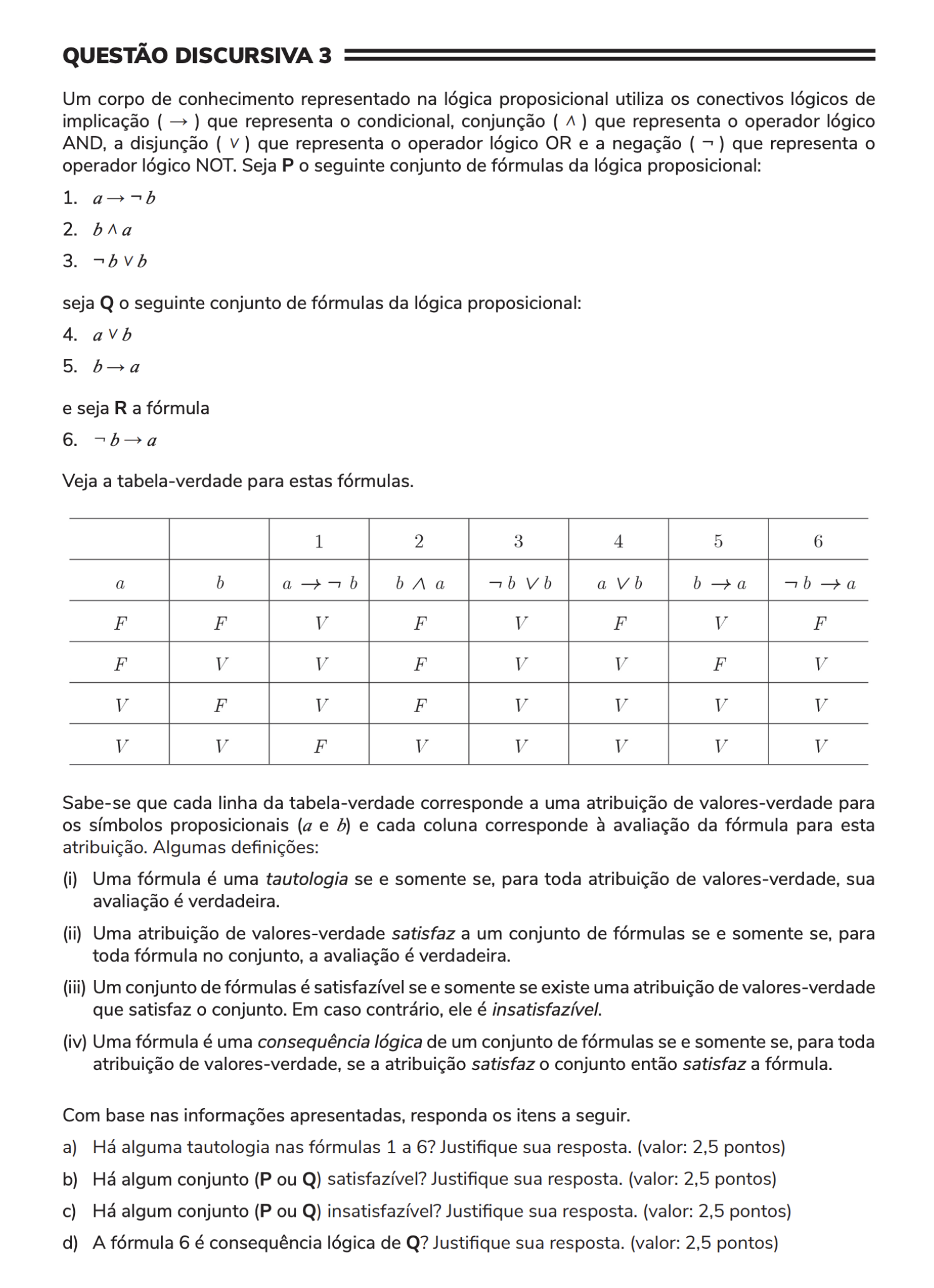}\\
	\vspace{8pt}

	\textbf{English transcription} \\[2pt]
	A body of knowledge represented in propositional logic uses logical connectives of implication (→) which represents the conditional, conjunction (∧) which represents the logical AND, the disjunction (∨) which represents the logical OR and the negation (¬) which represents the logical operator NOT. Let P be the following set of formulas of propositional logic:\\
	1. a → ¬b\\
	2. b ∧ a\\
	3. ¬b ∨ b\\

	let Q be the following set of formulas of propositional logic:\\
	4. a ∨ b\\
	5. b → ¬a\\

	and let R be the formula\\
	6. ¬b → a\\

	See the truth table for these formulas.\\

	[Table displaying truth values for propositions a and b, and corresponding truth values for formulas 1 through 6. The table has columns labeled "a", "b", "a → ¬b", "b ∧ a", "¬b ∨ b", "a ∨ b", "b → ¬a", "¬b → a", with rows that provide the truth values (True or False) for each formula based on the combination of truth values for a and b.]\\

	It is known that each row in the truth table corresponds to an attribution of truth values for the propositional symbols (a and b) and each column corresponds to the evaluation of the formula for that attribution. Some definitions:\\
	(i) A formula is a tautology if and only if, for every attribution of truth-values, its evaluation is true.\\
	(ii) An attribution of truth-values satisfies a set of formulas if and only if, for every formula in the set, the evaluation is true.\\
	(iii) A set of formulas is satisfiable if and only if there exists an attribution of truth-values that satisfies the set. Otherwise, it is unsatisfiable.\\
	(iv) A formula is a logical consequence of a set of formulas if and only if, for every truth-value attribution that satisfies the set, the attribution also satisfies the formula.\\

	Based on the information presented, answer the following items:\\
	a) Is there any tautology in formulas 1 to 6? Justify your answer.\\ (value: 2.5 points)\\
	b) Is there any satisfiable set (P or Q)? Justify your answer. \\(value: 2.5 points)\\
	c) Is there any unsatisfiable set (P or Q)? Justify your answer. \\(value: 2.5 points)\\
	d) Is formula 6 a logical consequence of Q? Justify your answer.\\ (value: 2.5 points)\\

	\textbf{Response standard}\\[2pt]
	a) The respondent must say that formula 3 contains a tautology, as it presents a true evaluation for every assignment (column of Vs).\\
	b) The respondent must indicate that the set Q is satisfiable, as all formulas of Q (4 and 5) are true for assignments 3 and 4 (one would be enough).\\
	c) The respondent must indicate that the set P is not satisfiable, as there is no assignment for which the formulas of P (1, 2, and 3) are all true.\\	
	d) The respondent must indicate that formula 6 is a logical consequence of Q, as all assignments that satisfy Q (3 and 4) also satisfy 6.		
\end{questionbox}
% \vfill
% \clearpage
% \begin{multicols}{2}
% \twocolumn
\begin{modelinputbox}[title={Initial prompt}]
	Answer the question in this image. Explain your reasoning step by step. Inform if the question is unclear or has no possible answer.\\

	[question image]
\end{modelinputbox}
% \vfill
% \clearpage
\begin{modeloutputbox}[title={ChatGPT-4 Vision's initial response}]
	Understood, let's address the questions.\\

	a) Is there any tautology in formulas 1 to 6?\\
	First, let's define what a tautology is: A formula in propositional logic is a tautology if, and only if, it is true for all possible truth-value assignments to its variables.\\
	
	Looking at the provided truth table:		
	- Formula 1 (a → b): Not a tautology, as it is false when a is true and b is false.\\
	- Formula 2 (b ∧ a): Not a tautology, as it is false for three out of the four possible combinations.\\
	- Formula 3 (¬b ∨ b): Is a tautology, as it is true in all assignments.\\
	- Formula 4 (a ∨ b): Not a tautology, as it is false when both a and b are false.\\
	- Formula 5 (b → a): Not a tautology, as it is false when b is true and a is false.\\
	- Formula 6 (¬b → a): Not a tautology, as it is false when ¬b is true and a is false.\\
	
	Answer: The only tautology is formula 3.\\
	
	b) Is there any satisfiable set (P or Q)?\\		
	A set of formulas is satisfiable if there is some truth-value assignment that makes all the formulas in the set true simultaneously.\\
	
	- Set P: We see that the second line of the table (where a is false and b is true) satisfies all three formulas in P.\\
	- Set Q: The third line of the table (where a is true and b is false) satisfies both formulas in Q.\\
	
	Answer: Both sets, P and Q, are satisfiable.\\
	
	c) Is there any unsatisfiable set (P or Q)?\\		
	Using the information from the previous item, we know that both sets are satisfiable, so none of them are unsatisfiable.\\
	
	Answer: No set is unsatisfiable.\\
	
	d) Is formula 6 a logical consequence of Q?\\		
	A formula is a logical consequence of a set if, whenever all the formulas in the set are true, the formula is also true.\\
	To check this, we observe any line where all formulas of Q are true and verify if formula 6 is true on the same line.\\	
	Formula 6 is true on lines 2, 3, and 4. The formulas in Q are both true on lines 3 and 4. In both these lines, formula 6 is true.\\
	
	Answer: Yes, formula 6 is a logical consequence of Q.
\end{modeloutputbox}
\begin{modelinputbox}[title={Reassessment prompt}]
	Your answers are not completely consistent with the response standard released for this question: [response standard]\\

	After analyzing the response standard, do you maintain or change your original answers? If you decide to change your answers, explain the possible reasons that led you to errors in the previous answers. Try to identify the concrete causes of errors, rather than listing generic reasons like misunderstanding the statement of the question.
\end{modelinputbox}
% \vfill
% \newpage
\begin{modeloutputbox}[title={ChatGPT-4 Vision's revised response}]
	I change my original answers to align with the released response standard:\\
	a) The only tautology is formula 3.\\
	b) The set Q is satisfiable, as all formulas of Q (4 and 5) are true for assignments 3 and 4.\\
	c) The set P is not satisfiable, as there is no assignment for which the formulas of P (1, 2, and 3) are all true simultaneously.\\	
	d) Formula 6 is a logical consequence of Q, as all assignments that satisfy Q (3 and 4) also satisfy 6.\\
	
	Upon examining the truth table and previous answers, we can identify some more specific errors:\\	
	1.~Satisfiable Set P: By stating that set P was satisfiable, it seems I confused the columns. Looking at the truth table, no single line makes all three formulas in P true simultaneously. The error here was clearly a failure in analyzing the truth table, perhaps giving too much attention to one column or line and losing sight of the overall picture.\\	
	2.~Evaluation of Tautologies: The tautology in formula 3 was correctly identified, but errors in other items may have arisen from a rushed analysis or not properly cross-referencing information between the formulas and the truth table.\\	
	3.~About Logical Consequence: The original answer to item d was correct. However, a mistake in understanding set P may have led to misunderstandings in the relationships between the formulas and their sets.\\	
	4.~Failure in the Verification Process: In stating something about satisfiability or logical consequence, an additional check of the truth table would have helped avoid errors. It seems the verification process was not rigorous enough.\\
	
	In summary, the errors seem to center on a combination of rushed analysis of the truth table and failure to follow a rigorous verification process. The complexity of the problem is not an excuse, as propositional logic and truth tables are something the model should be able to handle accurately. The most concrete failure was in meticulous analysis and cross-validation of information from the truth table with the given formulas and sets.
\end{modeloutputbox}
% \vfill
% \clearpage
\begin{expertbox}[title={Expert's scoring of ChatGPT-4 Vision's initial response}]
	a) Correct answer: the model correctly identified that formule 3 contains a tautology. (2.5/2.5 scores)\\
	b) Partially correct answer: the model correctly identified that set Q is satisfiable, but incorrectly identified that set P is also satisfiable. (1.25/2.5 scores)\\	
	c) Incorrect answer: the model incorrectly identified that set P is satisfiable. (0/2.5 scores)\\	
	d) Correct answer: the model correctly identified that formula 6 is a logical consequence of set Q. (2.5/2.5 scores)\\
	
	\textbf{Total scores: 2.5 + 1.25 + 0 + 2.5 = 6.25/10.0 (62.5/100)}
\end{expertbox}
\begin{expertbox}[title={Challenge categories / error types}]
	\textbf{Visual Acuity / Misidentification of Visual Elements}\\[2pt] 
	Upon self-reflection, the model admitted it did a rushed analysis of the truth table and confused the labels of some columns.\\

	\textbf{Logical Reasoning / Incorrect Multi-Step Reasoning}\\[2pt] 
	The model failed to cross-validate information between the formulas and the truth table, perhaps due to the error described above, leading to the incorrect evaluation of the satisfiability of set P.
\end{expertbox}
% \end{multicols}
% \onecolumn

\vfill
\clearpage

\subsubsection*{\myhref[Question 04]{04}}

This question asks the respondent to describe the truth table and draw the circuit diagram of a 1-bit full adder. ChatGPT-4 Vision correctly built the truth table but failed to draw the circuit diagram. This question highlights the model's capacity to understand complex computer architecture concepts such as logic circuits but also the model's current inability to generate domain-specific visual outputs such as a circuit diagram. 
% An expert's assessment of the model's initial response gave it 5.0 scores out of 10.

\vspace{\parskip}
\begin{questionbox}[title={Question}]
	% \textbf{Question image} \\[2pt]
	\includegraphics[width=\imagewidth,valign=t]{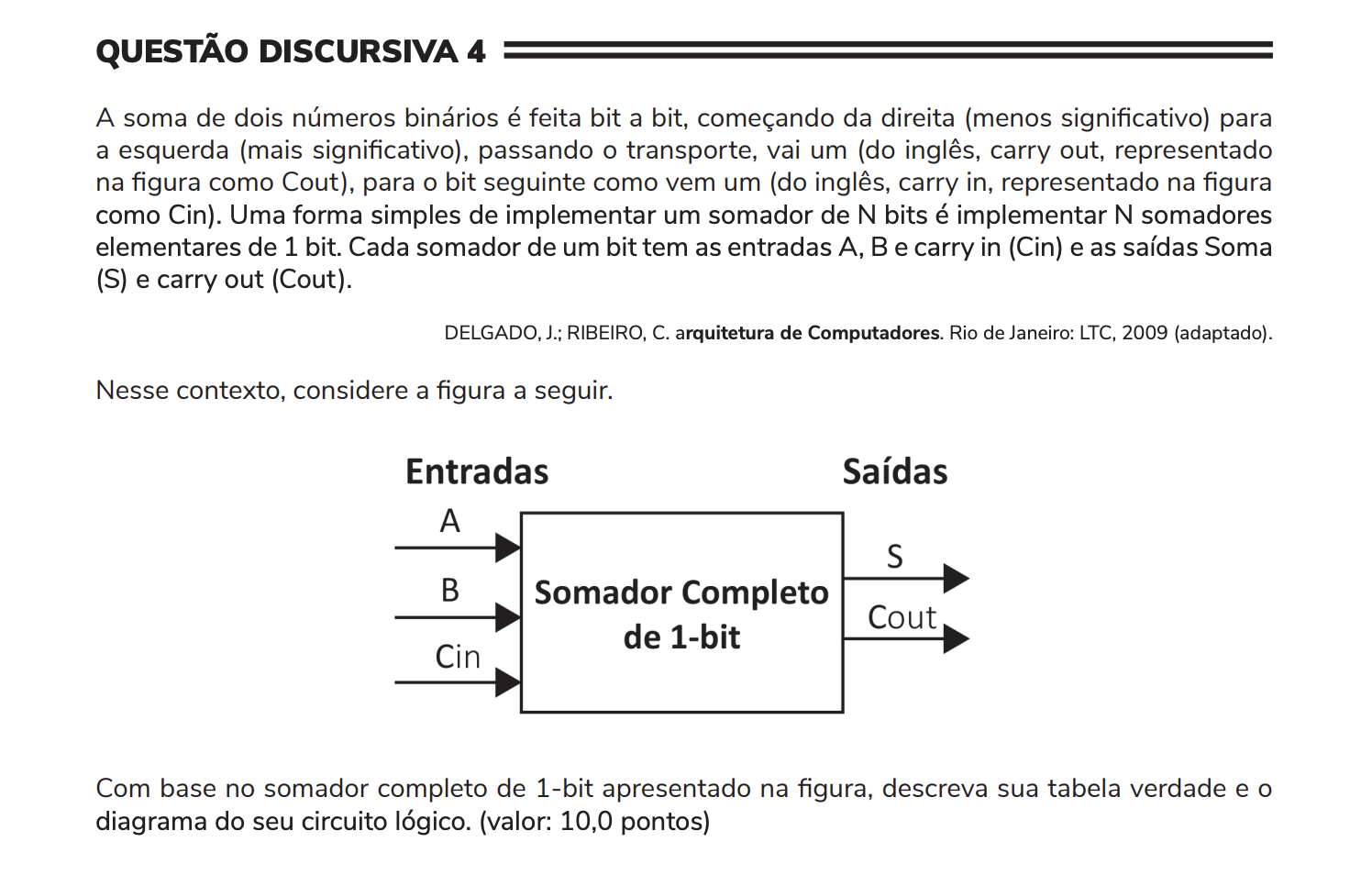}\\
	\vspace{8pt}

	\textbf{English transcription} \\[2pt]
	The sum of two binary numbers is done bit by bit, starting from the right (less significant) to the left (more significant), and the carryover is forwarded, that is, carry out (represented in the figure as \textit{Cout}), to the next bit as carry in (represented in the figure as \textit{Cin}). A simple way to implement an N-bit adder is to implement N elementary 1-bit adders. Each 1-bit adder has inputs A, B, and carry in (Cin) and the outputs Sum (S) and carry out (Cout).\\

	DELGADO, J.; RIBEIRO, C. Computer Architecture. Rio de Janeiro: LTC, 2009 (adapted).\\

	In this context, consider the following figure.\\

	[The image shows a block diagram of a "1-bit Full Adder" with three inputs labeled A, B, and Cin and two outputs labeled S and Cout.]\\

	Based on the 1-bit full adder depicted in the figure, describe its truth table and the diagram of its logic circuit. (value: 10.0 points)\\

	\textbf{Response standard} \\[2pt]
	The respondent should describe the truth table and draw the diagram, as below.\\
	
	\includegraphics[width=.6\linewidth]{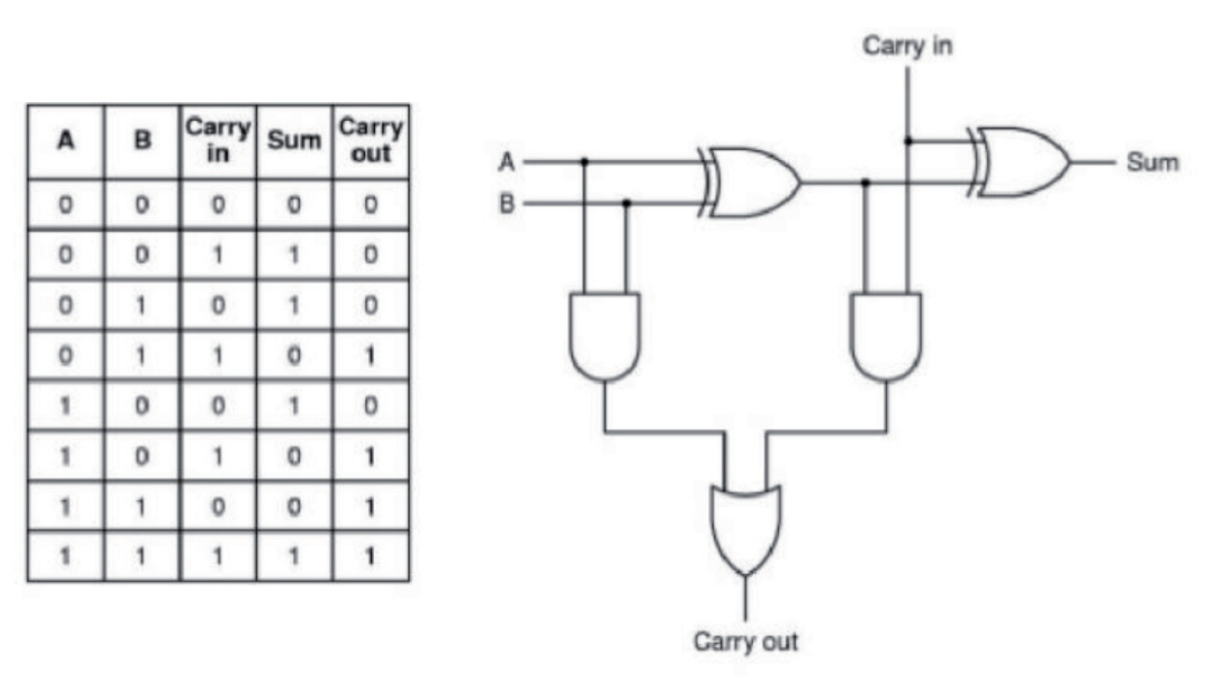}\\

	The truth table can be created in different ways, swapping "1"s and "0"s for "V" and "F", or for "T" and "F". Moreover, the rows can appear in any arbitrary order.\\

	The logic circuit can be solved in various ways, with variations in notation, and more than one circuit performing the same operation. The proposed circuit, in fact, combines the two outputs into a single circuit.\\
	
	Possible notations are presented in the following image, from the IEC 60617-12 standard, and also from ANSI IEEE.\\

	\includegraphics[width=.3\linewidth]{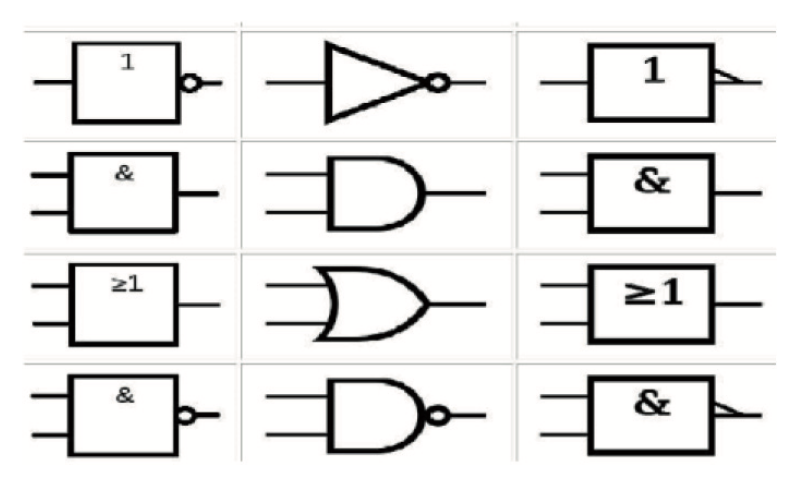}\\
	
	Other possible circuits, without being an exhaustive list, are:\\
	
	% \newcolumn

	Examples of possible answers for S:\\
	
	\includegraphics[width=.4\linewidth]{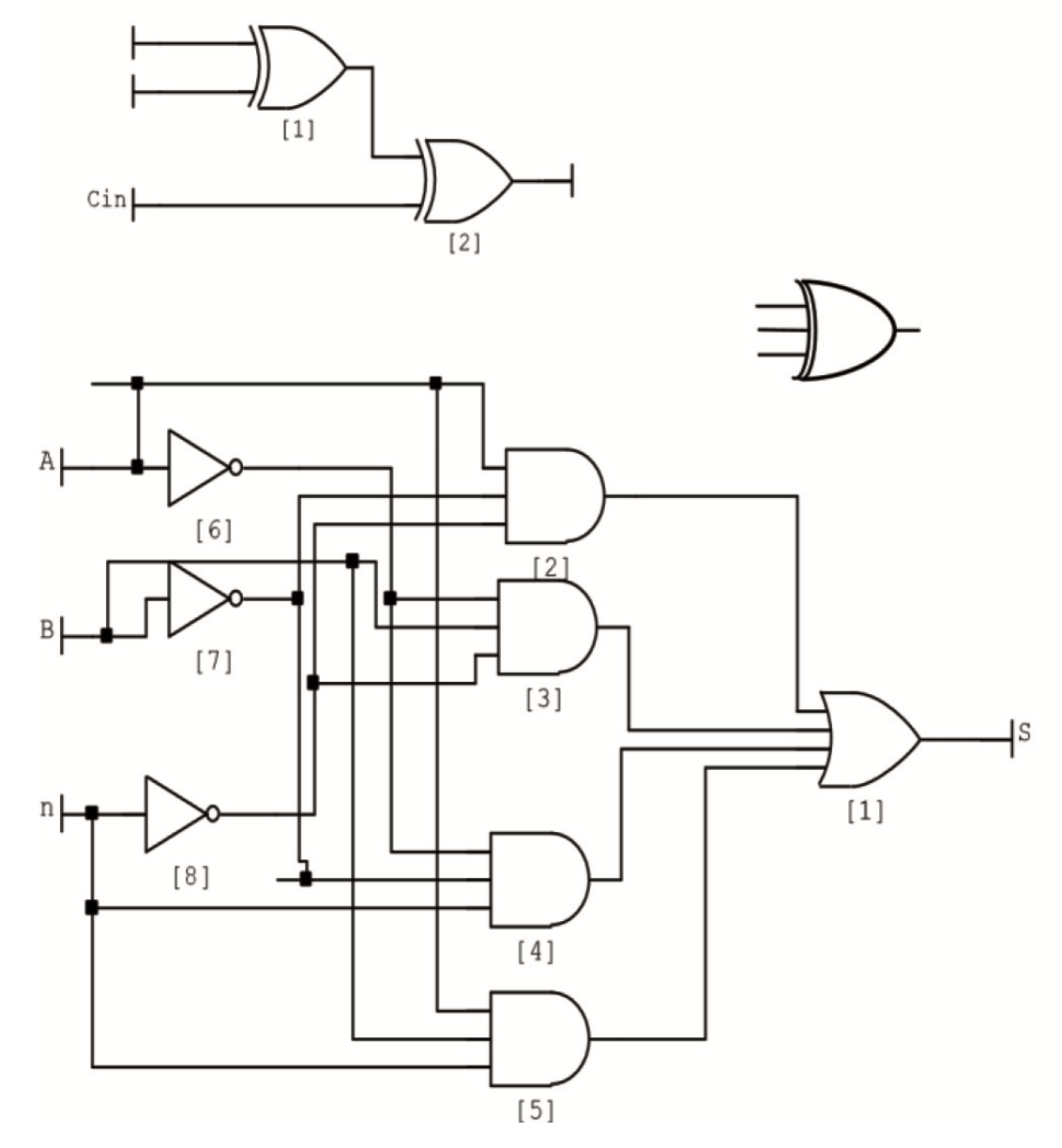}\\

	Examples of possible answers for Cout:\\

	\includegraphics[width=.6\linewidth]{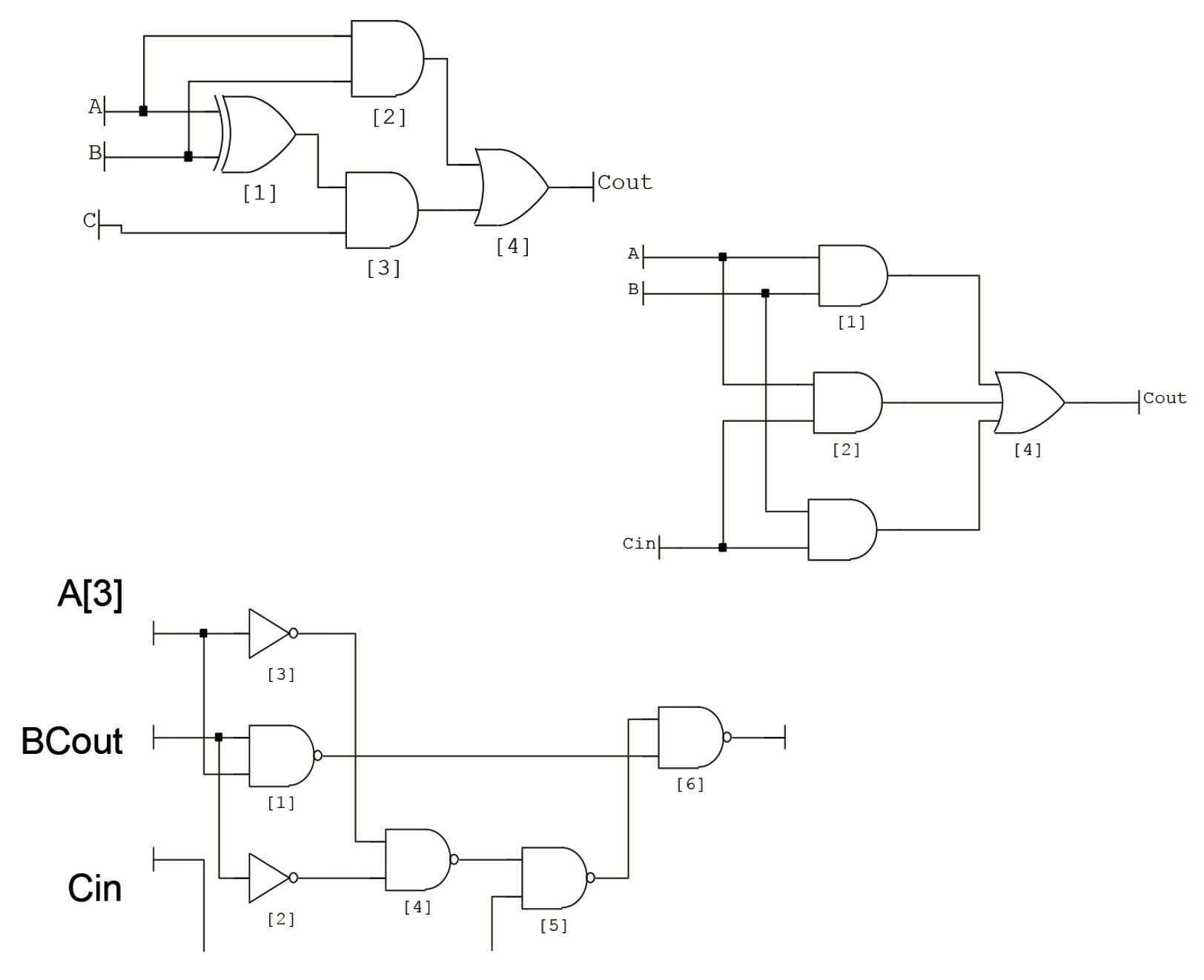}\\
\end{questionbox}
% \vfill
% \clearpage
% \begin{multicols}{2}
% \twocolumn
\begin{modelinputbox}[title={Initial prompt}]
	Answer the question in this image. Explain your reasoning step by step. Inform if the question is unclear or has no possible answer.\\

	[question image]
\end{modelinputbox}
\begin{modeloutputbox}[title={ChatGPT-4 Vision's initial response}]
	Understood. Let's address the problem of the 1-bit full adder.\\

	A 1-bit full adder has three inputs: A, B, and Cin (carry-in) and two outputs: S (sum) and 	Cout (carry-out). The aim of this adder is to add the three input bits and produce the corresponding sum bit and carry bit.\\
	
	Let's start with the truth table:

	\begin{verbatim}	
	| A | B | Cin | S | Cout |
	|---|---|-----|---|------|
	| 0 | 0 | 0   | 0 | 0    |
	| 0 | 0 | 1   | 1 | 0    |
	| 0 | 1 | 0   | 1 | 0    |
	| 0 | 1 | 1   | 0 | 1    |
	| 1 | 0 | 0   | 1 | 0    |
	| 1 | 0 | 1   | 0 | 1    |
	| 1 | 1 | 0   | 0 | 1    |
	| 1 | 1 | 1   | 1 | 1    |
	\end{verbatim}
	~\\
	Now, let's deduce the logical expressions for S and Cout:\\
	
	For the S output (sum):\\
	S = A XOR B XOR Cin\\
	
	For the Cout output (carry-out):\\
	There are three scenarios where Cout is 1:\\
	1. A = 1, B = 1\\
	2. A = 1, Cin = 1\\
	3. B = 1, Cin = 1\\
	
	Therefore:
	Cout = (A AND B) OR (A AND Cin) OR (B AND Cin)\\
	
	Finally, the logical circuit diagram consists of:\\
	1. Three XOR gates to calculate the sum S.\\
	2. Two AND gates and one OR gate to calculate the carry-out Cout.\\
	
	However, visualizing a logical circuit diagram is not possible here, but you can use the
	description above to draw it on paper or use a circuit drawing tool.\\
	
	In summary, a 1-bit full adder adds three input bits and produces a 2-bit result (sum and carry). 
	The logic to calculate these bits is determined by the expressions for S and Cout we just deduced.
\end{modeloutputbox}
% \vfill
% \newpage
\begin{expertbox}[title={Expert's scoring of ChatGPT-4 Vision's initial response}]
	The truth table produced by ChatGPT-4 Vision is correct according to the response standard. However, since the model cannot currently generate domain-specific diagrams, and was not explicitly prompted to "draw" a circuit using some alternative textual notation, it gets only half of the question's total scores.\\

	\textbf{Total scores: 5.0/10.0 (50/100)}
\end{expertbox}
\begin{expertbox}[title={Challenge categories / error types}]
	\textbf{Visual Acuity / Lack of	Domain-Specific Visual Output}\\[2pt] 
	The model failed to draw the diagram for the logic circuit, as requested by the question.
\end{expertbox}

\vfill
\clearpage

\subsubsection*{\myhref[Question 05]{05}}

This question asks the respondent to answer questions regarding the behavior and time complexity of a binary heap implementation in C whose code is provided as part of the question. ChatGPT-4 Vision gave an incorrect answer to the first question, where it failed to calculate the final state of a given array after the execution of the binary heap implementation with certain parameters, but correctly informed the binary heap's time complexity as the answer to the second question. This question showcases the model's capacity to understand complex algorithmic concepts such as a sorting algorithm's time complexity but exposes the model's limitations to correctly simulate the execution of a piece of code in a step-by-step manner.
% An expert's assessment of the model's initial response gave it 5.0 scores out of 10.

\vspace{\parskip}
\begin{questionbox}[title={Question}]
	% \textbf{Question image} \\[2pt]
	\includegraphics[width=\imagewidth,valign=t]{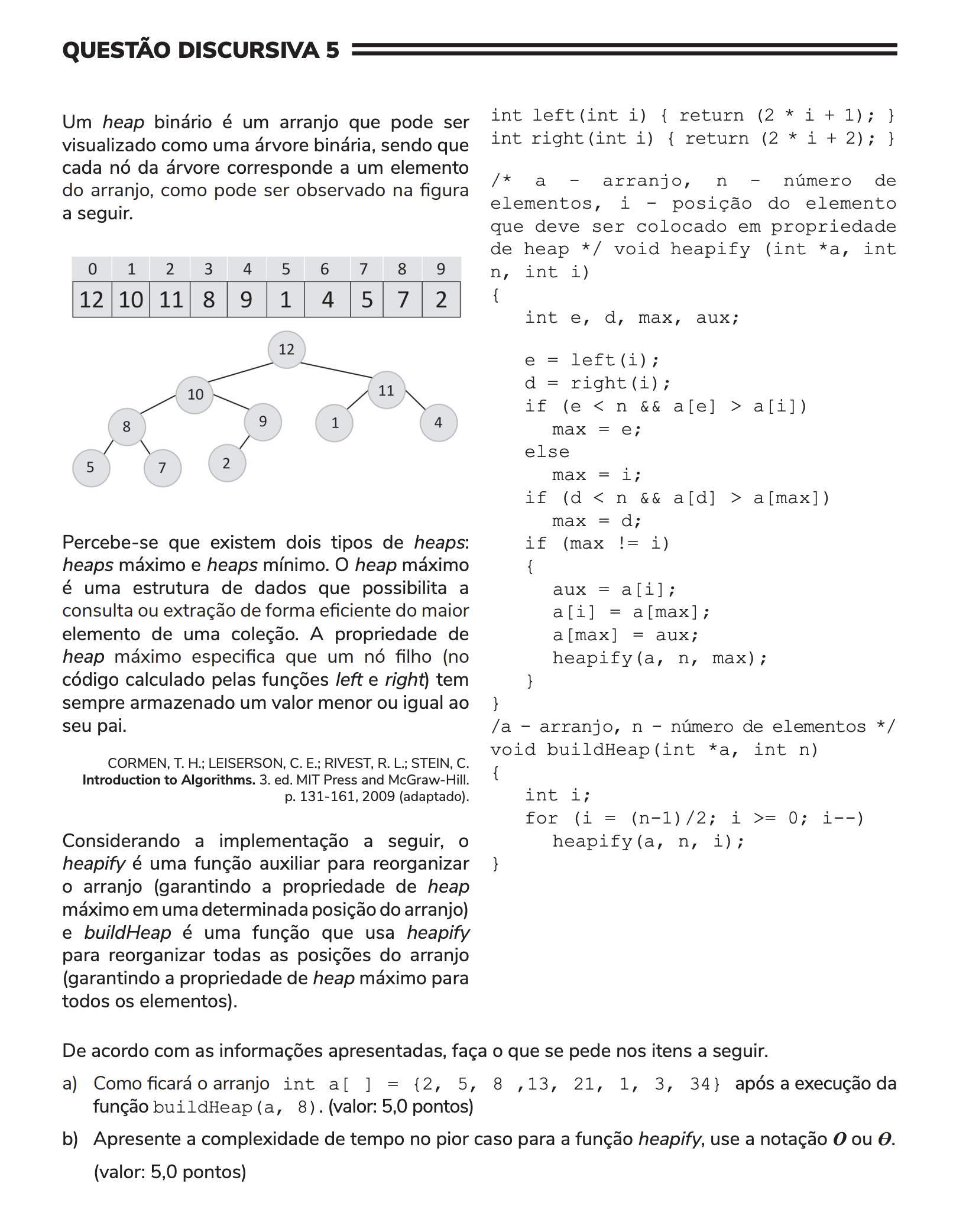}\\

	\vspace{8pt}
	\textbf{English transcription} \\[2pt]
	A binary heap is an array that can be visualized as a binary tree, where each tree node corresponds to an array element, as can be observed in the figure below.\\

	[An image of a binary tree containing 12 nodes with corresponding array indexes and values above it.]\\
	% The binary tree shows a tree structure with the number 12 at the root and subsequent child nodes displaying the numbers 10, 11, 8, 9, 1, 4, 5, 7 and 2. The corresponding array is shown above the tree with indices ranging from 0 to 9 and containing the values 12, 10, 11, 8, 9, 1, 4, 5, 7, 2.

	There exist two types of heaps: maximum heaps and minimum heaps. The maximum heap is a data structure that allows efficient query or removal of the maximum element from the collection. The maximum heap property specifies that a node (calculated by the functions \textit{left} and \textit{right} in the companying code) should store a value smaller or equal to its parent.\\
	
	CORMEN, T. H.; LEISERSON, C. E.; RIVEST, R. L.; STEIN, C. Introduction to Algorithms. 3. ed. MIT Press and McGraw-Hill, p. 131-161, 2009 (adapted).\\
	
	Considering the following implementation, \textit{heapify} is a helper function to organize the array (ensuring the heap property for a certain position of the array) and \textit{buildHeap} is a function that uses \textit{heapify} to reorganize all the positions of the array (ensuring the heap property of maximum heap for all its elements).

	\begin{lstlisting}[language=C]
	int left(int i) { return (2 * i + 1); }
	int right(int i) { return (2 * i + 2); }
	
	/* a - array, n - number of elements, i - position of the 
	element that must be placed in proper heap order */
	void heapify(int *a, int n, int i)
	{
		int e, d, max, aux;
	
		e = left(i);
		d = right(i);
		if (e < n && a[e] > a[i])
			max = e;
		else
			max = i;
		if (d < n && a[d] > a[max])
			max = d;
		if (max != i)
		{
			aux = a[i];
			a[i] = a[max];
			a[max] = aux;
			heapify(a, n, max);
		}
	}
	
	/* a - array, n - number of elements */
	void buildHeap(int *a, int n)
	{
		int i;
		for (i = (n-1)/2; i >= 0; i--)
			heapify(a, n, i);
	}
	\end{lstlisting}

	Based on the information presented, do what is requested in the items below.\\

	a) How will the array int a[] = \{2, 5, 8, 13, 21, 1, 3, 34\} look like after the execution of the function \textit{buildHeap(a, 8)}. (value: 5.0 points)\\		
	b) Present the time complexity in the worst case for the function \textit{heapify}, using the notation \textit{O} or \textit{Θ}. (value: 5.0 points)\\

	\textbf{Response standard} \\[2pt]
	a) The respondent should show that after executing the function \textit{buildHeap} the arrangement will be as follows: \{34, 21, 8, 13, 2, 1, 3, 5\}.\\
	b) The respondent should present that in the worst case for the \textit{heapify} function the time complexity will be as follows: $O(log n)$, being $n$ the number of elements in the heap.	
\end{questionbox}
% \vfill
% \clearpage
\begin{modelinputbox}[title={Initial prompt}]
	Answer the question in this image. Explain your reasoning step by step. Inform if the question is unclear or has no possible answer.\\

	[question image]
\end{modelinputbox}
\begin{modeloutputbox}[title={ChatGPT-4 Vision's initial response}]
	Let's address the presented questions:\\

	a) What will be the arrangement of int a[] = \{2, 5, 8, 13, 21, 1, 3, 34\} after executing the function \textit{buildHeap(a, 8)}?\\	

	The function \textit{buildHeap} begins by calling the \textit{heapify} function at the penultimate level of the tree (or heap). In terms of the array, this translates to starting from the middle of the array and going backwards.\\
	
	The \textit{heapify} method is responsible for maintaining the max heap property. If an element is smaller than its children, it will be swapped with the larger of its children.\\
	
	The initial arrangement is: 2, 5, 8, 13, 21, 1, 3, 34\\
	
	Now, let's go through the array using \textit{buildHeap}:\\	
	1. We start at i = 3 (remembering that indexing starts at 0), which is the value 13. Its children are 21 and 1. 21 is larger, and therefore, 13 and 21 swap places. Result: 2, 5, 8, 21, 13, 1, 3, 34\\
	2. i = 2, which is the value 8. Its children are 1 and 3. 8 is already larger than both, so nothing changes. Result: 2, 5, 8, 21, 13, 1, 3, 34\\
	3. i = 1, which is the value 5. Its children are 21 and 13. 21 is larger, so 5 and 21 swap places. Moreover, after this swap, 5 will be compared again with its new children (i.e., the heapify method is called again). Result after first swap: 2, 21, 8, 5, 13, 1, 3, 34 After the recursive call, 5 and 13 swap places. Result: 2, 21, 8, 13, 5, 1, 3, 34\\
	4. i = 0, which is the value 2. Its children are 21 and 8. 21 is larger, so 2 and 21 swap places. Result: 21, 2, 8, 13, 5, 1, 3, 34
	Thus, after the execution of the function buildHeap, the arrangement will be: 21, 2, 8, 13, 5, 1, 3, 34\\
	
	b) Present the worst-case time complexity for the \textit{heapify} function, using the $O$ or $Θ$ notation.\\	
	The complexity of the \textit{heapify} function is determined by the height of the tree, because in the worst case, it may be necessary to swap an element with its child until it reaches the bottom of the tree. The height of a complete binary tree is $log(n)$. Therefore, the worst-case time complexity for the \textit{heapify} function is $O(log n)$.\\
	
	Answers: a) 21, 2, 8, 13, 5, 1, 3, 34 b) $O(log n)$
\end{modeloutputbox}
\begin{modelinputbox}[title={Reassessment prompt}]
	Your answers are not completely consistent with the response standard released for this question: [response standard]\\

	After analyzing the response standard, do you maintain or change your original answers? If you decide to change your answers, explain the possible reasons that led you to errors in the previous answers. Try to identify the concrete causes of errors, rather than listing generic reasons like misunderstanding the statement of the question.
\end{modelinputbox}
% \vfill
% \newpage
\begin{modeloutputbox}[title={ChatGPT-4 Vision's revised response}]
	After reviewing the response standard, I would change my original answers as follows:\\

	a) The arrangement after executing the function \textit{buildHeap} is \{34, 21, 8, 13, 2, 1, 3, 5\}.\\	
	b) The worst-case time complexity of the \textit{heapify} function is $O(log n)$.\\
	
	Reasons for Errors:\\
	
	a) Error in Executing \textit{buildHeap}: When executing the function \textit{buildHeap}, I did not correctly apply the max heap property at some stages. For example, the value 34, which is the highest in the arrangement, should have risen to the root after the complete execution of \textit{buildHeap}, but this did not happen in my original answer. This indicates a failure in my simulation of the algorithm's execution.\\
	
	b) For the complexity, my answer was correct, but it could have been explained more clearly and directly.\\
	
	Concrete Causes of Errors:\\
	
	1.~Algorithm Simulation: In simulating the algorithm, I did not apply all possible swaps that \textit{heapify} could make. For example, the value 34 should have been promoted higher in the heap, but I missed this step. Algorithm simulation requires careful attention at each step, and it seems there was haste or lack of attention to details in my simulation.\\
	
	2.~Review: At the end of the simulation, I did not review the resulting structure to ensure it met the max heap property, which would have indicated that something was wrong with the original response.\\
	
	In summary, the errors mainly stem from failing to simulate the algorithm correctly and not reviewing the resulting structure to ensure it adhered to the desired property.
\end{modeloutputbox}
\begin{expertbox}[title={Expert's scoring of ChatGPT-4 Vision's initial response}]
	a) Incorrect answer: the model produced an incorrect arrangement according to the response standard. (0.0/5.0 scores)\\
	b) Correct answer: the model correctly informed the algorithm complexity in line with the response standard. (5.0/5.0 scores)\\
	
	\textbf{Total scores: 0.0 + 5.0 = 5.0/10.0 (50/100)}
\end{expertbox}
\begin{expertbox}[title={Challenge categories / error types}]
	\textbf{Logical Reasoning / Incorrect Algorithmic Reasoning}\\[2pt] 
	The model failed to correctly simulate the execution of the function \textit{buildHeap}, leading to an incorrect array arrangement.
\end{expertbox}
% \end{multicols}

\vfill
\clearpage

\subsection{Multiple-Choice Questions}
\label{app:mc-questions}

% The following six multiple-choice questions were selected to illustrate some representative capabilities and limitations of ChatGPT-4 Vision in answering the ENADE 2021 BCS exam. The conversations and expert assessments for the remaining multiple-choice questions are available in the paper's companion repository.

% \vfill
% \clearpage

\subsubsection*{\myhref[Question 11]{11}}

This question asks the respondent to identify the correct statements regarding the improvement of a deep neural network architecture for a specific image classification task and then select the alternative that refers to those correct statements. ChatGPT-4 Vision accurately identified the correct statements and their corresponding alternative while providing a detailed explanation for its choice. This question demonstrates the model's capacity to understand complex multimodal content in artificial intelligence and reason about it in a multi-step process.

\vspace{\parskip}
\begin{questionbox}[title={Question}]
	% \textbf{Question image} \\[2pt]
	\includegraphics[width=\imagewidth,valign=t]{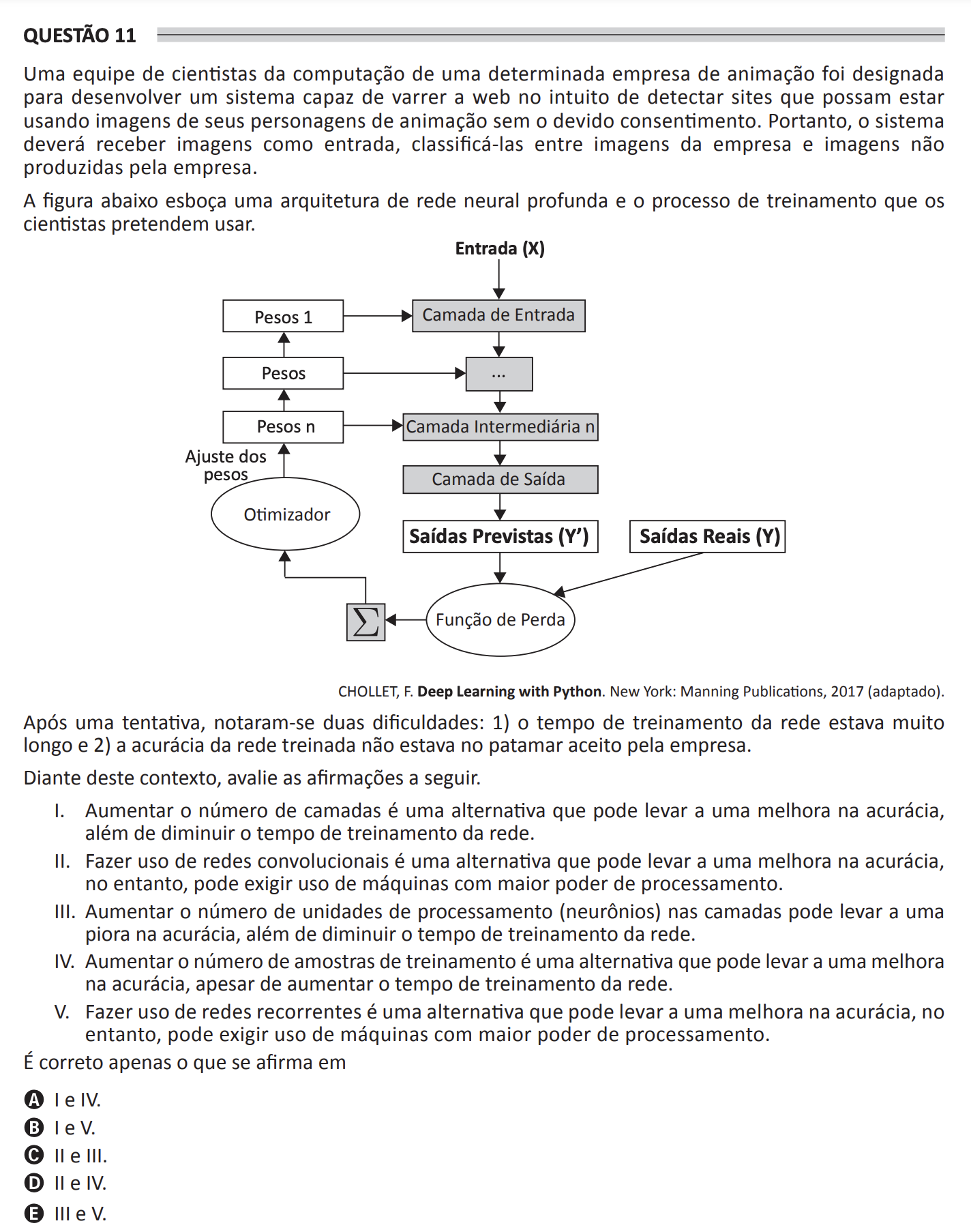}\\

	\vspace{8pt}
	\textbf{English transcription} \\[2pt]
	A team of computer scientists from a certain animation company has been tasked with developing a system capable of scanning the web to detect sites that might be using the company's animation character images without proper consent. Therefore, the system should receive images as input, and classify them between company images and images not produced by the company.\\

	The figure below outlines a deep neural network architecture and the training process the scientists intend to use.\\
	
	[The image shows a flow diagram for a deep neural network architecture, consisting of an input layer (Entrada (X)), a series of intermediary layers (Pesos 1, ..., Pesos n, with Camada de Entrada, ..., and Camada Intermediária n between the weight layers), leading to an output layer (Camada de Saída). This is followed by Predicted Outputs (Saídas Previstas (Y')) being compared with Real Outputs (Saídas Reais (Y)) using a Loss Function (Função de Perda), which feeds into an Optimizer (Otimizador) that adjusts the weights (Ajuste dos pesos).]\\
	
	CHOULLET, F. Deep Learning with Python. New York: Manning Publications, 2017 (adapted).\\
	
	After one attempt, two difficulties were noticed: 1) the network training time was too long, and 2) the network's accuracy was not at an acceptable level for the company.\\
	
	Given these findings, consider the following statements.\\
	
	I. Increasing the number of layers is an alternative that can lead to an improvement in accuracy, as well as decreasing the network training time.\\
	
	II. Making use of convolutional networks is an alternative that can lead to an improvement in accuracy, however, it may require the use of machines with greater processing power.\\
	
	III. Increasing the number of processing units (neurons) in the layers may lead to an improvement in accuracy, as well as decreasing the network training time.\\
	
	IV. Increasing the number of training samples is an alternative that can lead to an improvement in accuracy, as well as decreasing the network training time.\\
	
	V. Making use of recurrent networks is an alternative that can lead to an improvement in accuracy, however, it may require the use of machines with greater processing power.\\
	
	It is correct only what is stated in\\
	
	A) I and IV.\\
	B) I and V.\\
	C) II and III.\\
	D) II and IV.\\
	E) III and V.\\

	\textbf{Answer key}: D
\end{questionbox}
% \vfill
% \clearpage
% \begin{multicols}{2}
% \twocolumn
\begin{modelinputbox}[title={Initial prompt}]
	Answer the question in this image. Explain your reasoning step by step. At the end, choose which of the five alternatives (A-E) is correct. Inform if there is more than one correct alternative or if all five alternatives are incorrect.\\

	[question image]
\end{modelinputbox}
% \vfill
% \clearpage
\begin{modeloutputbox}[title={ChatGPT-4 Vision's initial response}]
	Let’s analyze the statements:\\

	I. Increasing the number of layers is an alternative that can lead to an improvement in accuracy, as well as decreasing the network training time.\\
	   - This is partially correct. Increasing the number of layers can improve accuracy, as it makes the network deeper and able to capture more complex features. However, it generally does not decrease training time, in fact, it may increase it since there are more parameters to be adjusted.\\
	
	II. Making use of convolutional networks is an alternative that can lead to an improvement in accuracy, however, it may require the use of machines with greater processing power.\\
		- This is true. Convolutional Neural Networks (CNNs) are widely used for image-related tasks and generally perform better on such tasks compared to fully connected networks. However, due to their nature and the number of operations involved, they may require more computing power, especially for large data volumes.\\
	
	III. Increasing the number of processing units (neurons) in the layers may lead to an improvement in accuracy, as well as decreasing the network training time.\\
	   - Increasing the number of neurons can improve accuracy since it allows the network to capture more features. However, similar to the first statement, increasing the number of neurons will likely not decrease the training time; in fact, it might increase it because there are more parameters to be trained.\\
	
	IV. Increasing the number of training samples is an alternative that can lead to an improvement in accuracy, as well as decreasing the network training time.\\
	   - This is true. More data generally lead to more robust and accurate models. However, with more data, the training time will increase because there are more examples for the network to process.\\
	
	V. Making use of recurrent networks is an alternative that can lead to an improvement in accuracy, however, it may require the use of machines with greater processing power.\\
	   - Recurrent Neural Networks (RNNs) are commonly used for sequences or temporal data. They are not typically the first choice for image classification tasks. Therefore, this statement is potentially misleading in the context of this question.\\
	
	Given the above analyses, statements II and IV are the most accurate.\\
	
	Answer: D) II and IV.
\end{modeloutputbox}
% \vfill
% \newpage
% \begin{expertbox}[title={Conclusion from ChatGPT-4 Vision's initial response}]
% 	ChatGPT-4 Vision's response is consistent with the answer key.
% \end{expertbox}
% \end{multicols}
% \onecolumn

\vfill
\clearpage

\subsubsection*{\myhref[Question 13]{13}}

This question asks the respondent to select the correct statement regarding iterative and evolutionary software development considering a given reference text. ChatGPT-4 Vision chose a different alternative from the answer key based on an inconsistent reasoning process where it sometimes followed a strict interpretation of the question and sometimes applied implicit domain knowledge not directly mentioned in the reference text. This question showcases the model's capacity to understand complex software development concepts but also its struggles with questions requiring implicit knowledge or inferences beyond the provided text. The fact that the three experts did not reach a consensus on the correct answer or the validity of the question also highlights the potential ambiguity of the question and the critical need for better quality control in the exam elaboration. Note: this question was one of the six questions excluded from the final score due to a Weak discrimination level (see Section~\ref{sec:questions-stats}).

\vspace{\parskip}
\begin{questionbox}[title={Question}]
	% \textbf{Question image} \\[2pt]
	\includegraphics[width=\imagewidth,valign=t]{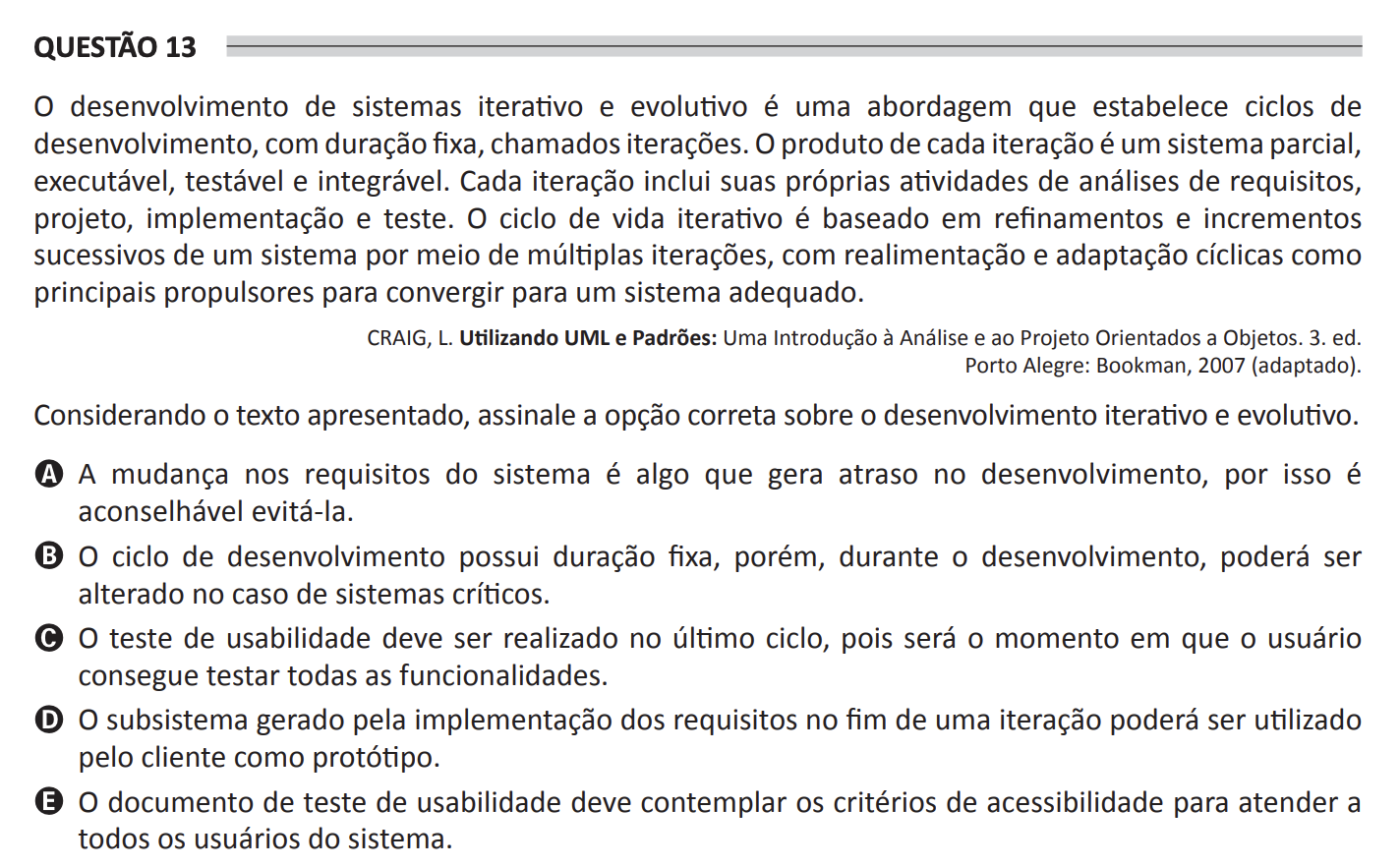}\\

	\vspace{8pt}
	\textbf{English transcription} \\[2pt]
	Iterative and evolutionary systems development is an approach that establishes cycles of development with a fixed duration, called iterations. The product of each iteration is a partial, executable, testable, and integrable system. Each iteration includes its own requirement analysis, project, implementation, and test activities. The iterative lifecycle is based on refinements and increments of a system through multiple iterations, with cyclical feedback and adaptation as the main drivers to converge to a suitable system.\\

	CRAIG, L. Using UML and Patterns: An Introduction to Object-Oriented Analysis and Design. 3rd ed. Porto Alegre: Bookman, 2007 (adapted).\\
	
	Considering the presented text, select the correct option about iterative and evolutionary development.\\
	
	A) Changing system requirements is something that causes delays in development, so it is advisable to avoid doing it.\\	
	B) The development cycle has a fixed duration but may be changed in the case of critical systems.\\	
	C) Usability testing should be carried out in the last cycle because it is the moment when the user can test all functionalities.\\	
	D) The subsystem generated by the implementation of requirements at the end of an iteration may be used by the customer as a prototype.\\	
	E) The usability testing document must contemplate accessibility criteria to serve all system users.\\

	\textbf{Answer key}: E
\end{questionbox}
% \begin{multicols}{2}
% \twocolumn
\begin{modelinputbox}[title={Initial prompt}]
	Answer the question in this image. Explain your reasoning step by step. At the end, choose which of the five alternatives (A-E) is correct. Inform if there is more than one correct alternative or if all five alternatives are incorrect.\\

	[question image]
\end{modelinputbox}
% \vfill
% \clearpage
\begin{modeloutputbox}[title={ChatGPT-4 Vision's initial response}]
	Let's analyze the text and the options provided step by step:\\

	Text: Iterative and evolutionary systems development is an approach that establishes cycles of development with a fixed duration, called iterations. The product of each iteration is a partial, executable, testable, and integrable system. Each iteration includes its own requirement analysis, project, implementation, and test activities. The iterative lifecycle is based on refinements and increments of a system through multiple iterations, with cyclical feedback and adaptation as the main drivers to converge to a suitable system.\\
	
	Based on the text, let's analyze the alternatives:\\
	
	A) Changing system requirements is something that causes delays in development, so it is advisable to avoid doing it.\\
	- The text does not mention that changes in requirements cause delays, and iterative and evolutionary systems, in general, are designed to accommodate changes in requirements. Therefore, this option seems incorrect.\\
	
	B) The development cycle has a fixed duration but may be changed in the case of critical systems.\\
	- The text mentions that the cycle has a "fixed duration," but it does not make any reference to changes in the case of "critical systems." Therefore, this alternative is partially true, but not entirely based on text.\\
	
	C) Usability testing should be carried out in the last cycle because it is the moment when the user can test all functionalities.\\
	- The text does not specifically mention "usability testing," nor does it state that it should be performed in the last cycle. Therefore, this option seems incorrect.\\
	
	D) The subsystem generated by the implementation of requirements at the end of an iteration may be used by the customer as a prototype.\\
	- The text mentions that the "product of each iteration is a partial system, executable, testable, and integrable," which can be interpreted as a "prototype" that the client can view and test. This alternative seems to be aligned with the text.\\
	
	E) The usability testing document must contemplate accessibility criteria to serve all system users.\\
	- Again, the text does not specifically mention "usability testing" or "accessibility" criteria. Therefore, this option seems incorrect.\\
	
	Conclusion:\\
	Alternative D "The subsystem generated by the implementation of requirements at the end of an iteration may be used by the customer as a prototype" is the most aligned with the information presented in the text. Hence, the correct alternative is D.
\end{modeloutputbox}
\begin{modelinputbox}[title={Reassessment prompt}]
	Your response differs from the answer key released for this question: [answer key]\\
	
	After analyzing the answer key, do you maintain or change your response? If you decide to change your response, explain the possible reasons that led you to errors in the previous answer. Try to identify the concrete causes of the errors instead of listing generic reasons, such as a failure to understand the question statement.
\end{modelinputbox}
\begin{modeloutputbox}[title={ChatGPT-4 Vision's revised response}]
	I maintain response D as correct based on the analysis of the text provided. Option D aligns with the concept of iterative and evolutionary development, where each iteration results in a product or subsystem that is "executable, testable, and integrable". This subsystem can indeed be used by the client as a prototype or an incremental version of the final system.\\

	Regarding alternative E, which was chosen in the answer key, I can highlight the following points:\\
	- The text does not mention "accessibility" criteria.\\
	- "Usability testing" is not mentioned in the text.\\
	
	The previous error may have been caused by the attempt to interpret the statement too broadly, trying to infer information that was not explicitly mentioned in the text. Option E may have been chosen based on the idea that usability tests usually consider accessibility, but this is not indicated in the presented text. In objective assessments, it is crucial to base oneself strictly on the information given in the statement and the text, avoiding assumptions or inferences that go beyond what is provided.\\
	
	In conclusion, the previous error was caused by the attempt to infer information not explicitly mentioned in the text, leading to the choice of an option that was not directly supported by the content presented.
\end{modeloutputbox}
\begin{modelinputbox}[title={Clarification prompt}]
	To be clear: if you maintain the choice of alternative D as correct, what do you mean by "previous error"? Who would have committed such an error? The one who elaborated the answer key?
\end{modelinputbox}
\begin{modeloutputbox}[title={ChatGPT-4 Vision's clarifying response}]
	I apologize for the confusion. By mentioning "previous error", I was referring to the discrepancy between the choice in the answer key and the analysis made here. If response D is the one I consider correct based on the analysis of the text, then the "error" would be in the creation of the answer key that indicated alternative E as correct. That is, the possible error would be from the part of those who elaborated the answer key or, alternatively, there could be additional information or contexts that were not considered in the analysis made here. Again, I apologize for the confusion and hope to have clarified the point.
\end{modeloutputbox}
% \vfill
% \clearpage
\begin{expertbox}[title={Experts' assessments}]
	\textbf{Expert \#1}\\

	ChatGPT is correct.\\

	If the question statement asks to consider the reference text, the correct choice is indeed D. I agree with ChatGPT's analysis of the alternatives.\\

	\textbf{Expert \#2}\\

	A) Changing system requirements is something that causes delays in development, so it is advisable to avoid doing it.\\
	- The text does not mention that changes in requirements cause delays, and iterative and evolutionary systems, in general, are designed to accommodate changes in requirements. Therefore, this option seems incorrect.\\
	
	ChatGPT's answer is dubious, as it is common knowledge in software engineering (SE) that changes in requirements cause delays. There is an incompatibility with the text. To avoid doubt, the question should say: it is something that CAN cause delays ... Therefore, the alternative is not entirely incorrect.\\
	
	B) The development cycle has a fixed duration but may be changed in the case of critical systems.\\
	- The text mentions that the cycle has a "fixed duration," but it does not make any reference to changes in the case of "critical systems." Therefore, this alternative is partially true, but not entirely based on text.\\
	
	The text does not mention it, but we know that the cycles must be followed, although their duration varies according to the system or size, so the statement is not entirely incorrect.\\
	
	C) Usability testing should be carried out in the last cycle because it is the moment when the user can test all functionalities.\\
	- The text does not specifically mention "usability testing," nor does it state that it should be performed in the last cycle. Therefore, this option seems incorrect.\\
	
	Neither the text mentions nor is it common sense in SE, I agree that the alternative is incorrect.\\
	
	D) The subsystem generated by the implementation of requirements at the end of an iteration may be used by the customer as a prototype.\\
	- The text mentions that the "product of each iteration is a partial system, executable, testable, and integrable," which can be interpreted as a "prototype" that the client can view and test. This alternative seems to be aligned with the text.\\
	
	It is common sense in software engineering that we should not use the prototype as the final system. The text mentions a partial system, so the alternative is incorrect.\\
	
	E) The usability testing document must contemplate accessibility criteria to serve all system users.\\
	- Again, the text does not specifically mention "usability testing" or "accessibility" criteria. Therefore, this option seems incorrect.\\
	
	The text does not mention it, but it is common sense that usability tests should include non-functional requirements like accessibility. So the alternative is correct.\\
	
	Conclusion:\\
	
	The question is ambiguous and should be invalidated, as alternatives A, B, C, and E are not contained in the text. It seems to me that the text was used as an inducer of the questions, but the intention is to put general alternatives about the iterative and incremental life cycle. Thinking this way, I arrived at two possible answers A and E. However, as there was no CAN in A, then E would be more appropriate, so the answer key would be correct. However, the candidate has to make a reflection outside the text, so I consider the question ambiguous. The text hinders the interpretation of the alternatives.\\

	\textbf{Expert \#3}\\

	Analyzing all the material, I conclude that the official answer key is correct.\\

	Justification: options A-D contain errors based on my knowledge of Software Engineering, knowledge that is not limited to the statement of the question. It seems to me that ChatGPT's choice and justification are based solely on the statement because it repeatedly says that the text does not mention certain concepts. But at the same time, ChatGPT chooses option D, which mentions a subsystem that, in turn, was not defined in the text of the statement. It seems to me that it inferred what a subsystem would be, and in my opinion, its inference is incorrect (i.e., a subsystem is not a partial system).	
\end{expertbox}
\begin{expertbox}[title={Conclusion from the experts' assessments}]	
	The response from ChatGPT is inconsistent with the answer key and the three experts did not reach a consensus on what the correct answer would be or whether the question should be invalidated. 
\end{expertbox}
\begin{modelinputbox}[title={Reflectice prompt to ChatGPT-4 Turbo}]
	The following are the prompts and responses of a generative AI for the attached question, along with possible assessments by two or more human experts on the generative AI's responses and the general quality of the question. Based on the provided information, summarize the main difficulties faced by the generative AI in attempting to answer the question, including possible difficulties resulting from incorrect recognition of some graphical or textual element present in the statement of the question. Consider only concrete difficulties whose manifestation can be evidenced by the information provided in generative AI's responses. At the end, discuss if these difficulties could be mitigated or eliminated by using prompt engineering techniques or if they would be inherent to the limitations of current large language models (LLMs).\\

	[question image]\\

	[ChatGPT-4 Vision's prompts and responses] \\

	[expert assessments]
\end{modelinputbox}
% \vfill
% \clearpage
\begin{modeloutputbox}[title={ChatGPT-4 Turbo's response},colbacktitle=gray!35!white]
	The generative AI faced difficulty in attempting to answer the question on iterative and evolutionary systems development, resulting in a discrepancy between its answer and the official key. The main difficulties identified include:\\

	1. Strict Text Interpretation: The AI chose alternative D based on a literal interpretation of the text, which mentions a "partial system" as the outcome of each iteration, interpreting it as a prototype. However, the official answer key indicated alternative E as correct, which the AI did not consider based on the provided text.\\
	
	2. Focus on Explicit Details: The AI maintained its answer, even after the discrepancy with the answer key was pointed out, arguing that its choice was closely in line with the information given in the text and that criteria for "accessibility" could not be inferred to justify alternative E.\\
	
	3. Expert Feedback: The human experts provided conflicting feedback. One agreed with the AI, deeming alternative D as correct. Others considered the question ambiguous or suggested that the correct alternative should be E, arguing the question required knowledge beyond the text.\\
	
	These difficulties suggest that while the AI can analyze text and choose answers based on explicit information, it may struggle with interpretations requiring implicit knowledge or inferences beyond the provided text. Prompt engineering could be used to guide the AI to consider knowledge beyond the text, but this remains a limitation of current LLMs, which rely heavily on the input text and may not fully understand context or the expectations for implicit knowledge in specific domains, such as software engineering. It indicates a continued need for the advancement of AI models to better integrate textual comprehension with contextualized knowledge.\\	
\end{modeloutputbox}
\begin{modelinputbox}[title={Clarification prompt to ChatGPT-4 Turbo}]
	And as for the fact that the experts did not come to a consensus on the question, any additional comments?
\end{modelinputbox}
\begin{modeloutputbox}[title={ChatGPT-4 Turbo's clarifying response},colbacktitle=gray!35!white]
	The lack of consensus among human experts on the correct response to the question underscores the complexity and inherent ambiguity in some exam questions. This may arise for several reasons:\\

	1. Ambiguity in the Statement: If the statement of the question is ambiguous or insufficiently detailed, it may lead to multiple valid interpretations.\\
	
	2. Interpretation Differences: Experts can have different perspectives based on their experiences, contextual knowledge, and understanding of the subject. What may appear clear to one might be debatable to another.\\
	
	3. Implicit Knowledge Expectations: Some questions may presuppose knowledge that is not explicitly declared in the text of the question. Experts might disagree on how much of this implicit knowledge should be applied when interpreting the question.\\
	
	4. Variable Evaluation Criteria: Experts might value different criteria when assessing the correct response, such as practical relevance versus theoretical accuracy.\\
	
	This challenge is not only for AI but also for question design and assessments in general. In the context of AI, such ambiguity and the discrepancy of human opinions expose the limitations of AI systems in dealing with uncertainty and making judgments when there is no clear answer or defined consensus.\\
	
	Prompt engineering might help AI to ask for clarifications or to consider a broader range of interpretations, but the limitations of current LLMs mean there is still no reliable way for AI to resolve ambiguities without a clear consensus or well-defined criteria. This underscores the importance of continuing to develop LLM capacities for navigating the complexities and nuances of human communication and information interpretation.		
\end{modeloutputbox}
\begin{expertbox}[title={Challenge categories / error types}]
	\textbf{Question Interpretation / Inconsistent Responses}\\[2pt] 
	The model produced inconsistent responses where it sometimes followed a strict interpretation of the question and sometimes applied implicit domain knowledge not directly mentioned in the question statement.
\end{expertbox}

\vfill
\clearpage

\subsubsection*{\myhref[Question 22]{22}}

This question asks the respondent to select the most appropriate relational schema to be generated from a given Entity-Relationship Diagram (ERD). ChatGPT-4 Vision correctly selected the most appropriate alternative as the answer to the question. This question shows the model's ability to understand complex system modeling concepts such as entities and relationships from a purely visual representation and reason about them based on external domain knowledge.

\vspace{\parskip}

% \vfill
% \clearpage

\begin{questionbox}[title={Question}]
	% \textbf{Question image} \\[2pt]
	\includegraphics[width=\imagewidth,valign=t]{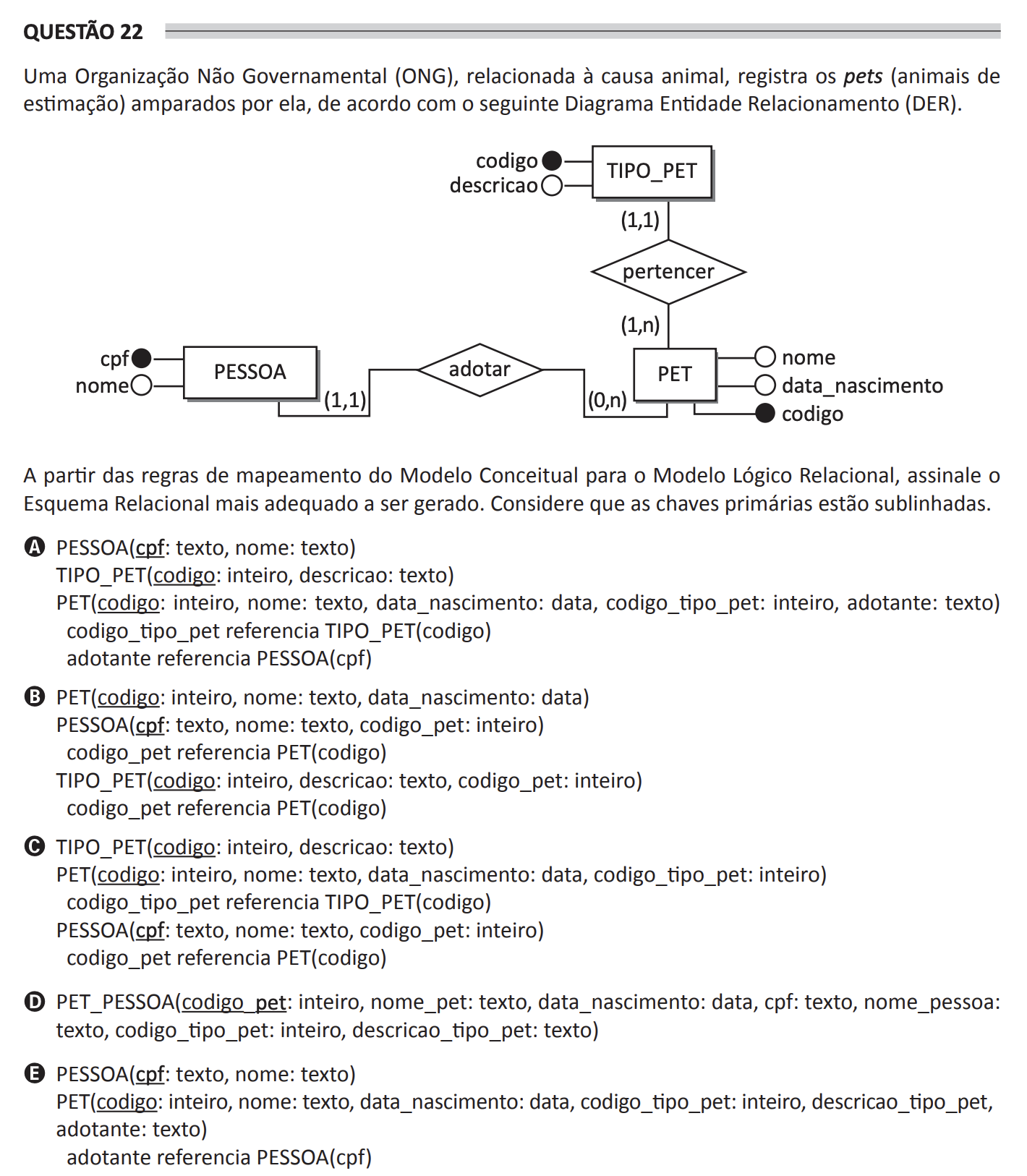}\\

	\vspace{8pt}
	\textbf{English transcription} \\[2pt]
	A Non-Governmental Organization (NGO), related to the animal cause, registers the pets (companion animals) supported by it, according to the following Entity-Relationship Diagram (ERD).\\

	[The image shows an Entity-Relationship Diagram (ERD) consisting of three entities connected by relationships: PERSON connected to PET by an "adopt" relationship, and PET connected to PET\_TYPE by a "belong" relationship. Attributes for each entity are as follows: PERSON has 'cpf' and 'name', PET has 'name', 'birth\_date', and 'code', and PET\_TYPE has 'code' and 'description'. The relationships also indicate cardinalities: (1,1) for PERSON to PET, (0,n) for PET to PERSON, (1,n) for PET to PET\_TYPE, and (1,1) for PET\_TYPE to PET.]\\
	
	Based on the mapping rules from the Conceptual Model to the Relational Logical Model, select the most appropriate Relational Schema to be generated. Consider that primary keys are underlined.\\
	
	A)\\
	PERSON(\underline{cpf}: text, name: text)\\
	PET\_TYPE(\underline{code}: integer, description: text)\\
	PET(\underline{code}: integer, name: text, birth\_date: date, pet\_type\_code: integer, adopter: text)\\
	pet\_type\_code references PET\_TYPE(code) \\
	adopter references PERSON(cpf)\\	
	B)\\
	PET(\underline{code}: integer, name: text, birth\_date: date) \\
	PERSON(\underline{cpf}: text, name: text, pet\_code: integer)\\
	pet\_code references PET(code)\\
	PET\_TYPE(\underline{code}: integer, description: text, pet\_code: integer)\\
	pet\_code references PET(code)\\	
	C)\\
	PET\_TYPE(\underline{code}: integer, description: text)\\
	PET(\underline{code}: integer, name: text, birth\_date: date, pet\_type\_code: integer)\\
	pet\_type\_code references PET\_TYPE(code) \\
	PERSON(\underline{cpf}: text, name: text, pet\_code: integer)\\
	pet\_code references PET(code)\\	
	D)\\
	PET\_PERSON(\underline{pet\_code}: integer, pet\_name: text, birth\_date: date, cpf: text, person\_name: text, pet\_type\_code: integer, pet\_type\_description: text)\\	
	E) \\
	PERSON(\underline{cpf}: text, name: text)\\
	PET(\underline{code}: integer, name: text, birth\_date: date, pet\_type\_code: integer, pet\_type\_description, adopter: text)\\
	adopter references PERSON(cpf)\\

	\textbf{Answer key}: A
\end{questionbox}
% \vfill
% \clearpage
% \begin{multicols}{2}
% \twocolumn
\begin{modelinputbox}[title={Initial prompt}]
	Answer the question in this image. Explain your reasoning step by step. At the end, choose which of the five alternatives (A-E) is correct. Inform if there is more than one correct alternative or if all five alternatives are incorrect.\\

	[question image]
\end{modelinputbox}
% \vfill
% \clearpage
\begin{modeloutputbox}[title={ChatGPT-4 Vision's initial response}]
	Let's start by analyzing the Entity Relationship Diagram (ERD) presented and understand each of the entities and their relationships:\\
	1.~PET\_TYPE: Represents the type of animal, with attributes code and description.\\
	2.~PERSON: Represents a person, with attributes cpf and name.\\
	3.~PET: Represents an animal supported by the NGO, with attributes name, birth\_date, code and has relationships with PET\_TYPE and PERSON.\\
	
	Observed relationships:\\
	1.~Each PET belongs to a type, indicated by the "belong" relationship between PET and PET\_TYPE.\\
	2.~Each PET can be adopted by a PERSON, as per the "adopt" relationship between PET and PERSON.\\
	
	Based on this, let's examine the alternatives:\\	

	A)\\
	- PERSON(cpf: text, name: text): Seems correct.\\
	- PET\_TYPE(code: integer, description: text): Seems correct.\\
	- PET(code: integer, name: text, birth\_date: date, code\_pet\_type: integer, adopter: text): Seems correct, since there is a reference to PET\_TYPE by code and a reference to PERSON by cpf (adopter).\\
	
	B)\\
	- PET(code: integer, name: text, birth\_date: date): Seems correct.\\
	- PERSON(cpf: text, name: text, code\_pet: integer): This suggests that a person can only adopt one pet, which contradicts the ERD that allows a person to adopt multiple pets (0, n).\\
	- PET\_TYPE(code: integer, description: text, code\_pet: integer): This suggests that an animal type is linked to a single pet, which is not true.\\ 

	C)\\
	- PET\_TYPE(code: integer, description: text): Seems correct.\\
	- PET(code: integer, name: text, birth\_date: date, code\_pet\_type: integer): Seems correct.\\
	- PERSON(cpf: text, name: text, code\_pet: integer): Again, this suggests that a person can only adopt one pet.\\
	
	D)\\
	- PET\_PERSON(code\_pet: integer, name\_pet: text, birth\_date: date, cpf: text, name\_person: text, code\_pet\_type: integer, description\_pet\_type: text): This looks like a combination of all entities into a single table, which is not the appropriate relational model.\\

	E)\\
	- PERSON(cpf: text, name: text): Seems correct.\\
	- PET(code: integer, name: text, birth\_date: date, code\_pet\_type: integer, description\_pet\_type, adopter: text): Here, we are including the "description\_pet\_type" directly in the PET table, which is an unnecessary redundancy and not an appropriate practice.\\
	
	The most suitable alternative that correctly translates the ERD into the relational schema is Alternative A.
\end{modeloutputbox}
% \vfill
% \newpage
% \begin{expertbox}[title={Conclusion from ChatGPT-4 Vision's initial response}]
% 	ChatGPT-4 Vision's response is consistent with the answer key.
% \end{expertbox}
% \end{multicols}
% \onecolumn

\vfill
\clearpage

\subsubsection*{\myhref[Question 28]{28}}

This question asks the respondent to assess the validity of four statements about a theoretical model of an instruction pipeline, visually illustrated in two accompanying figures, and then select the alternative that contains only valid statements. ChatGPT-4 Vision accurately evaluated the four statements, providing detailed explanations for each of them, and selected the correct alternative as the answer to the question. This question demonstrates the model's ability to understand complex computer architecture concepts such as pipeline stages and their interactions over time and reason about them based on external domain knowledge.

\vspace{\parskip}

\begin{questionbox}[title={Question}]
	% \textbf{Question image} \\[2pt]
	\includegraphics[width=\imagewidth,valign=t]{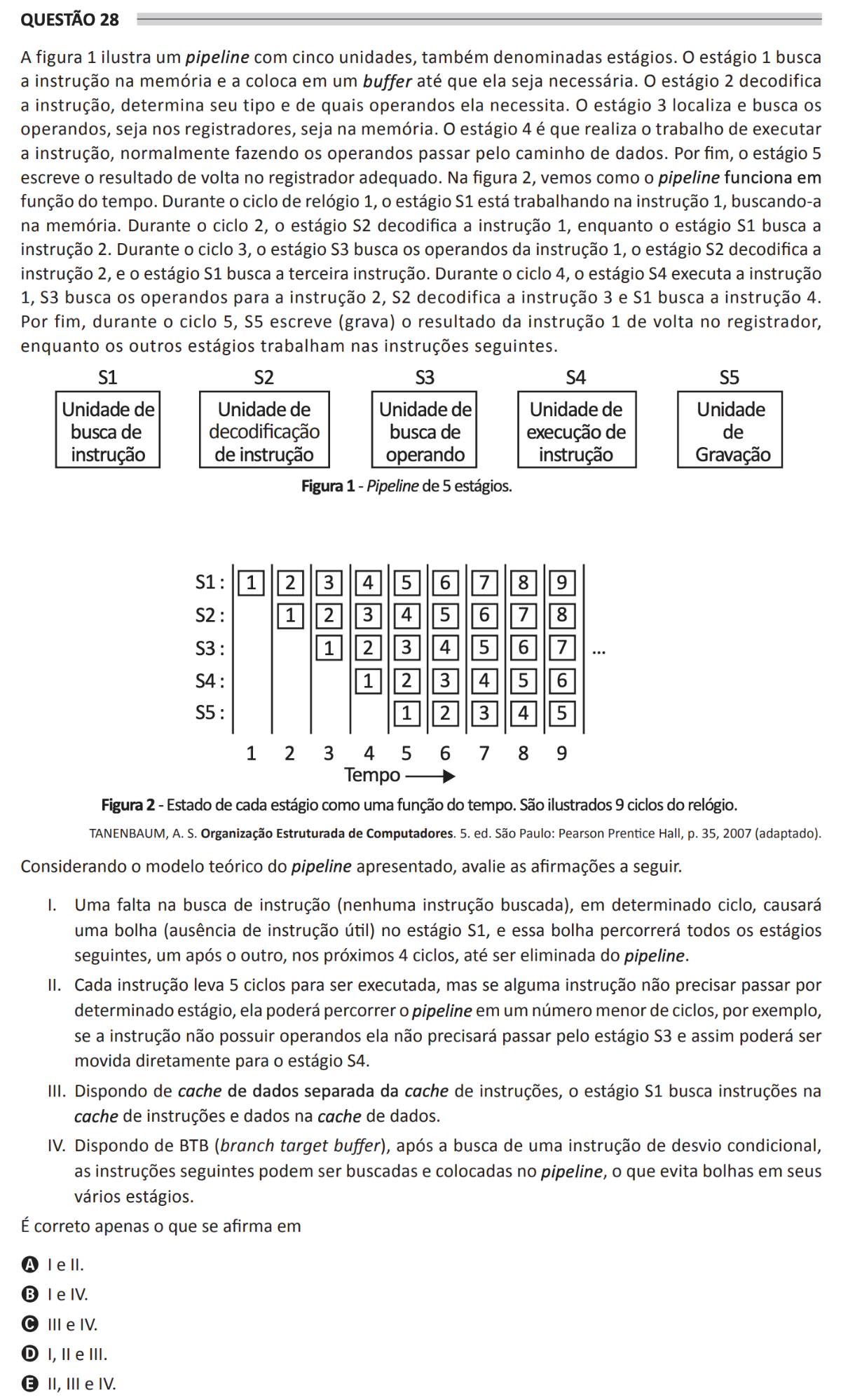}\\

	\vspace{8pt}
	\textbf{English transcription} \\[2pt]
	Figure 1 illustrates a pipeline with five units, also called stages. Stage 1 fetches the instruction from memory and places it into a buffer until it is needed. Stage 2 decodes the instruction, determines its type, and which operands it requires. Stage 3 locates and fetches the operands, whether from registers or memory. Stage 4 is the one that performs the work of executing the instruction, typically by passing the operands through the data path. Finally, stage 5 writes the result back into the appropriate register. In Figure 2, we see how the pipeline functions over time. During clock cycle 1, stage S1 is working on instruction 1, fetching it from memory. During cycle 2, stage S2 decodes instruction 1, while stage S1 fetches instruction 2. During cycle 3, stage S3 fetches the operands for instruction 1, stage S2 decodes instruction 2, and stage S1 fetches the third instruction. During cycle 4, stage S4 executes instruction 1, S3 fetches operands for instruction 2, S2 decodes instruction 3, and S1 fetches instruction 4. Finally, during cycle 5, S5 writes (records) the result of instruction 1 back into the register, while the other stages work on the following instructions.\\

	[Figure 1 - Pipeline with 5 stages. The figure shows five rectangles aligned horizontally representing stages S1 to S5, as explained above.]\\
	
	[Figure 2 - State of each stage as a function of time.]\\
	%  The figure shows 9 clock cycles. Rows labeled S1 through S5 indicate the stage of the pipeline (such as instruction fetch, decode, operand fetch, execute, and write back) and the columns numbered 1 through 9 represent the time (clock cycles). Each cell contains a number indicating which instruction is being processed by a specific stage at a given time.]\\
	
	TANENBAUM, A. S. Structured Computer Organization. 5th ed. São Paulo: Pearson Prentice Hall, p. 35, 2007 (adapted).\\
	
	Considering the theoretical model of the pipeline presented, evaluate the following statements.\\
	
	I. A miss in the instruction fetch (no instruction fetched) during a certain cycle will cause a bubble (absence of a useful instruction) in stage S1, and this bubble will travel through all the following stages, one after the other, in the next 4 cycles, until it is eliminated from the pipeline.\\
	
	II. Each instruction takes 5 cycles to execute, but if some instruction does not need to pass through a certain stage, it can move through the pipeline in fewer cycles, for example, if the instruction does not have operands it will not need to pass through stage S3 and thus can be moved directly to stage S4.\\
	
	III. With a separate data cache from the instruction cache, stage S1 fetches instructions in the instruction cache and data in the data cache.\\
	
	IV. With a BTB (branch target buffer), after the fetch of a conditional branch instruction, subsequent instructions can be fetched and placed into the pipeline, which avoids bubbles in its various stages.\\
	
	It is correct only what is stated in\\
	
	A) I and II.\\	
	B) I and IV.\\	
	C) II and IV.\\	
	D) I, II and III.\\	
	E) II, III and IV.\\

	\textbf{Answer key}: B
\end{questionbox}
% \vfill
% \clearpage
% \begin{multicols}{2}
% \twocolumn
\begin{modelinputbox}[title={Initial prompt}]
	Answer the question in this image. Explain your reasoning step by step. At the end, choose which of the five alternatives (A-E) is correct. Inform if there is more than one correct alternative or if all five alternatives are incorrect.\\

	[question image]
\end{modelinputbox}
% \vfill
% \clearpage
\begin{modeloutputbox}[title={ChatGPT-4 Vision's initial response}]
	Let's begin by analyzing each of the statements:\\

	I. A miss in the instruction fetch (no instruction fetched) during a certain cycle will cause a bubble (absence of a useful instruction) in stage S1, and this bubble will travel through all the following stages, one after the other, in the next 4 cycles, until it is eliminated from the pipeline.\\
	   - This is true. If we observe the pipeline model provided, we can notice that a miss in the instruction fetch stage (S1) will cause this miss (or bubble) to pass through all subsequent stages, taking one cycle per stage. Since there are 5 stages, it will take 4 additional cycles (beyond the cycle in which the miss occurred) for the bubble to be eliminated.\\
	
	II. Each instruction takes 5 cycles to execute, but if some instruction does not need to pass through a certain stage, it can move through the pipeline in fewer cycles, for example, if the instruction does not have operands it will not need to pass through stage S3 and thus can be moved directly to stage S4.\\
	   - This is inaccurate. Even if an instruction does not require operands, it still occupies a slot in stage S3 and simply will not perform any operation during that stage. The pipeline is designed to allow the simultaneous processing of multiple instructions at different stages, but each instruction still passes through every stage in sequence. Therefore, this statement is false.\\
	
	III. With a separate data cache from the instruction cache, stage S1 fetches instructions in the instruction cache and data in the data cache.\\
	   - This is true. In systems that have separate caches for instructions and data (Harvard architecture caches), the instruction fetch will occur in the instruction cache and the data fetch (e.g., operands) will occur in the data cache.\\
	
	IV. With a BTB (branch target buffer), after the fetch of a conditional branch instruction, subsequent instructions can be fetched and placed into the pipeline, which avoids bubbles in its various stages.\\
	   - This is true. The BTB is a hardware structure used to predict the outcome of conditional branch instructions. If a branch is predicted as taken, the BTB can provide the target address of the branch, allowing the next instruction to be fetched without waiting for the current branch to be resolved, thus helping to avoid bubbles in the pipeline.\\
	
	% Analyzing the options:\\	
	% A) I and II.\\
	% B) I and IV.\\
	% C) II and IV.\\
	% D) I, II and III.\\
	% E) II, III and IV.\\
	
	The correct option is alternative B) I and IV, as both statements I and IV are true.
\end{modeloutputbox}
% \vfill
% \newpage
% \begin{expertbox}[title={Conclusion from ChatGPT-4 Vision's initial response}]
% 	ChatGPT-4 Vision's response is consistent with the answer key.
% \end{expertbox}
% \end{multicols}
% \onecolumn

\vfill
\clearpage

\subsubsection*{\myhref[Question 31]{31}}

This question asks the respondent to evaluate four statements about the behavior of a given Turing machine diagram. ChatGPT-4 Vision made several mistakes in attempting to evaluate these statements. First, the model failed to exclude the ``B'' symbols from the tape content when evaluating statements I and II, as instructed by the question. Second, the model failed to correctly interpret some labels in the Turing machine diagram, leading to an incorrect analysis of the machine's operations. Third, perhaps due to these previous two errors, the model made an incorrect interpretation of the machine's actions, especially regarding the substitution and movement of the reading head, leading to an erroneous conclusion about the content of the tape after the requested number of moves. Upon being confronted with the answer key and challenged with multiple clarification prompts, the model made several revisions to its initial response but ultimately concluded that all four statements were true and thus the question had no correct answer. The full conversion with the model was then escalated to two human experts who unanimously agreed that the answer key was correct. This question shows that ChatGPT-4 Vision's performance can significantly be affected by its inability to correctly interpret key diagram labels and to accurately simulate an abstract machine's operations. Moreover, the model tends to overlook its limitations due to its typical overconfidence in question-answering tasks.

\vspace{\parskip}
\begin{questionbox}[title={Question}]
	% \textbf{Question image} \\[2pt]
	\includegraphics[width=\imagewidth,valign=t]{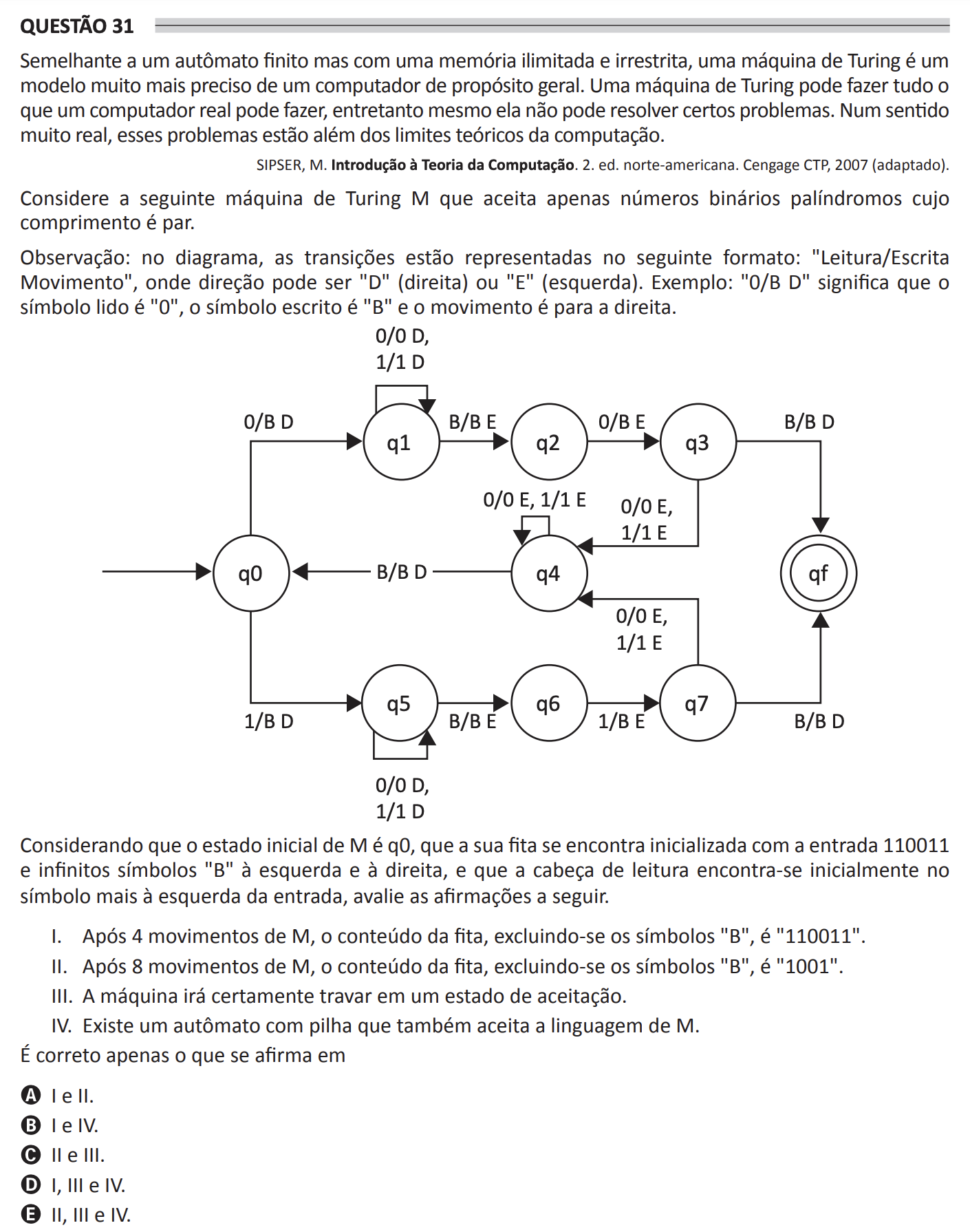}\\

	\vspace{8pt}
	\textbf{English transcription} \\[2pt]
	Similar to a finite automaton but with unlimited and unrestricted memory, a Turing machine is a much more precise model of a general-purpose computer. A Turing machine can do everything that a real computer can do; however, even it cannot solve certain problems. In a very real sense, these problems are beyond the theoretical limits of computation.\\

	SIPSER, M. Introduction to the Theory of Computation. 2nd ed. North American. Cengage CTP, 2007 (adapted).\\
	
	Consider the following Turing machine M that accepts only even-length binary palindromes.\\
	
	Observation: in the diagram, transitions are represented in the following format: "Read/Write Movement", where the direction can be "D" (right) or "E" (left). Example: "0/B D" indicates that the read symbol is "0", the written symbol is "B", and the movement is to the right.\\
	
	[In the image, there is a diagram of the Turing machine with states q0 through q7, along with the final state qf, and the transitions between them are represented with arrows indicating the condition/action pairs for each state transition.]\\
	
	Considering that the initial state of M is q0, that its tape is initially loaded with the input 110011 and there are infinite "B" symbols to the left and right, and that the reading head is initially at the leftmost symbol of the entry, evaluate the following statements.\\
	
	I. After 4 movements of M, the tape content, excluding the "B" symbols, is "110011".\\		
	II. After 8 movements of M, the tape content, excluding the "B" symbols, is "1001".\\		
	III. The machine will certainly halt in an acceptance state.\\		
	IV. There is a stack automaton that also accepts the language of M.\\
	
	It is correct only what is stated in\\		
	A) I and II.\\		
	B) I and IV.\\		
	C) II and III.\\		
	D) I, III, and IV.\\		
	E) II, III, and IV.\\

	\textbf{Answer key}: E
\end{questionbox}
% \begin{multicols}{2}
% \twocolumn
\begin{modelinputbox}[title={Initial prompt}]
	Answer the question in this image. Explain your reasoning step by step. At the end, choose which of the five alternatives (A-E) is correct. Inform if there is more than one correct alternative or if all five alternatives are incorrect.\\

	[question image]
\end{modelinputbox}
% \vfill
% \clearpage
\begin{modeloutputbox}[title={ChatGPT-4 Vision's initial response}]
	Let's analyze the operations of the given Turing machine and then evaluate the statements:\\

	The input is 110011 and the machine starts in state q0.\\
	
	1. From q0, '1' is read and the machine moves to the right, going to state q1.\\
	2. From q1, '1' is read and the machine moves to the right, staying in q1.\\
	3. From q1, '0' is read and the machine moves to the right, going to state q2.\\
	4. From q2, '0' is read and the machine moves to the right, staying in q2.\\
	
	After 4 moves, the tape content is still 110011, but the reading head is after the first two '0's.\\
	
	5. From q2, '1' is read and the machine moves to the left, going to state q5.\\
	6. From q5, '0' is read and replaced with 'B', moving to the left, staying in q5.\\
	7. From q5, '0' is read and replaced with 'B', moving to the left, staying in q5.\\
	8. From q5, '1' is read and the machine moves to the left, going to state q6.\\
	
	After 8 moves, the tape content is 11B0B1 and the reading head is after the first '1'.\\
	
	Let's evaluate the statements:\\
	
	I. After 4 moves of M, the tape content, excluding the "B" symbols, is "110011".\\
	Correct. As seen above.\\
	
	II. After 8 moves of M, the tape content, excluding the "B" symbols, is "11001".\\
	Incorrect. The content is 11B0B1.\\
	
	III. The machine will certainly halt in an accepting state.\\
	Correct. The Turing machine diagram shows that, after correctly identifying an even-length binary palindrome, the machine will move to the final state qf.\\
	
	IV. There is a pushdown automaton that also accepts the language of M.\\
	Correct. A pushdown automaton can be used to recognize palindromes, as the nature of the stack allows for the verification of symmetry. The Turing machine is more general and can simulate a pushdown automaton.\\
	
	Given the analyses, the correct statements are I, III, and IV.\\
	
	The correct answer is D) I, III, and IV.
\end{modeloutputbox}
% \vfill
% \clearpage
\begin{modelinputbox}[title={Reassessment prompt}]
	Your response differs from the answer key released for this question: [answer key]\\
	
	After analyzing the answer key, do you maintain or change your response? If you decide to change your response, explain the possible reasons that led you to errors in the previous answer. Try to identify the concrete causes of the errors instead of listing generic reasons, such as a failure to understand the question statement.
\end{modelinputbox}
\begin{modeloutputbox}[title={ChatGPT-4 Vision's revised response}]
	Let's revisit the analysis and see if I identify any discrepancy or error in my previous interpretation:\\

	From the description given and the Turing machine diagram:\\

	1. It is correct that, after 4 moves, the tape content is "110011".\\
	2. It is correct that, after 8 moves, the tape content is "11B0B1".\\
	3. The machine, upon receiving an even-length binary palindrome, will reach the final state qf, indicating acceptance.\\
	4. The nature of even-length binary palindromes can be depicted by a pushdown automaton. A pushdown automaton can be used to recognize palindromes because it can push symbols until it reaches the middle of the input and then pop matching symbols for the remaining input symbols.\\

	Based on the review, my earlier conclusions seem to be correct. The correct answer is D) I, III, and IV.\\

	Reasons for any discrepancy with the official key:\\
	- There may be an error in the key released.\\
	- The interpretation of the statement or the operations of the Turing machine may differ among different evaluators.\\

	After this review, I maintain my original answer as D) I, III, and IV. If the released key is indeed at odds, I recommend consulting the entity that administered the exam for clarifications or possible corrections.
\end{modeloutputbox}
% \vfill
% \clearpage
\begin{modelinputbox}[title={First clarification prompt}]
	A question: does statement II not explicitly mention that the tape content should be evaluated excluding the "B" symbols?
\end{modelinputbox}
\begin{modeloutputbox}[title={ChatGPT-4 Vision's first clarifying response}]
	You are correct. Let's re-evaluate statement II.\\

	II. After 8 moves of M, the tape content, excluding the "B" symbols, is "11001".\\
	
	Given that after 8 moves the tape content is "11B0B1" and if we exclude the "B" symbols, we get "11001", statement II is, in fact, correct.\\
	
	I apologize for my previous error. This means the correct answer is E) II, III, and IV.\\
	
	I appreciate you pointing out the discrepancy, and I regret the confusion caused.
\end{modeloutputbox}
\begin{modelinputbox}[title={Second clarification prompt}]
	I don't understand. How can the correct answer be E if you reaffirmed that statement I is also correct?
\end{modelinputbox}
\begin{modeloutputbox}[title={ChatGPT-4 Vision's second clarifying response}]
	I apologize for my error. You are correct. Let us correct that.\\

	The correct statements, as per our analysis, are:\\
	
	I. After 4 moves of M, the tape content, excluding the "B" symbols, is "110011".\\
	Correct.\\	
	II. After 8 moves of M, the tape content, excluding the "B" symbols, is "11001".\\
	Correct.\\	
	III. The machine will certainly halt in an accepting state.\\
	Correct.\\	
	IV. There is a pushdown automaton that also accepts the language of M.\\
	Correct.\\
	
	Given this, the correct statements are I, II, III, and IV. Thus, the correct answer should be an option that includes all these statements, which is not present among the given options.\\
	
	I regret my previous error and thank you for your patience.
\end{modeloutputbox}
\begin{expertbox}[title={Experts' assessments}]
	\textbf{Expert \#1}\\

	I follow the official answer key.\\
	
	Assertion I is not true because after 4 movements of M, the content of the tape, ignoring the B symbols, is "10011". ChatGPT is not taking into account the action of writing B on the tape, when reading 1 and transitioning from q0 to q5.\\
	
	Assertion II is correct, after 8 movements, the content of the tape is 1001. ChatGPT seems to be having some problem with image recognition, as in the responses, it is mentioned that\\
	\textit{II. After 8 movements of M, the content of the tape, excluding the "B" symbols, is "11001".} \\
	There is already a problem there, but even so, the execution of the machine is incorrect by ChatGPT, as there is no way the content can be "11B0B1". Moreover, when asked to review the answer and eliminate the B symbols, there is again inaccuracy as it claims equivalence between "11B0B1" and "11001".\\
	
	Assertion III is correct because the input is a palindrome, which leads the machine to "converge" to the acceptance state qf.\\
	
	Assertion IV is also correct, as palindromes can be represented with pushdown automata.\\

	\textbf{Expert \#2}\\

	Let's analyze the operations of the given Turing machine and then evaluate the assertions:\\
	
	The input is 110011 and the machine starts in state q0.\\
	
	1. From q0, '1' is read, replaced by 'B', and moved to the right, going to state q5.\\
	2. From q5, '1' is read and moved to the right, remaining in q5.\\
	3. From q5, '0' is read and moved to the right, remaining in q5.\\
	4. From q5, '0' is read and moved to the right, remaining in q5.\\
	
	After 4 movements, the content of the tape is 10011, and the read head is after the first two '0's.\\
	
	5. From q5, '1' is read and moved to the right, remaining in q5.\\
	6. From q5, '1' is read and moved to the right, remaining in q5.\\
	7. From q5, 'B' is read and moved to the left, going to state q6.\\
	8. From q6, '1' is read, replaced by 'B', and moved to the left, going to state q7.\\
	
	After 8 movements, the content of the tape is 1001, and the read head is after the last '0'.\\
	
	Let's evaluate the assertions:\\	
	I. After 4 movements of M, the content of the tape, excluding the "B" symbols, is "110011".\\
	Incorrect. The content is 10011.\\	
	II. After 8 movements of M, the content of the tape, excluding the "B" symbols, is "1001".\\
	Correct. As seen above.\\	
	III. The machine will certainly lock in an acceptance state.\\
	Correct. The Turing machine's diagram shows that, after correctly identifying an even-numbered binary palindrome, the machine will move to the final state qf.\\	
	IV. There exists a pushdown automaton that also accepts the language of M.\\
	Correct. A pushdown automaton can be used to recognize palindromes, as the nature of the stack allows for symmetry verification. The Turing machine is more general and can simulate a pushdown automaton.\\
	
	Given the analyses, the correct assertions are II, III, and IV.\\
	
	The correct answer is E) II, III, and IV.
\end{expertbox}
\begin{expertbox}[title={Conclusion from the experts' assessments}]	
	According to the two experts, the answer key is correct and ChatGPT's response is incorrect.
\end{expertbox}
\begin{modelinputbox}[title={Reflectice prompt to ChatGPT-4 Turbo}]
	The following are the prompts and responses of a generative AI for the attached question, along with possible assessments by two or more human experts on the generative AI's responses and the general quality of the question. Based on the provided information, summarize the main difficulties faced by the generative AI in attempting to answer the question, including possible difficulties resulting from incorrect recognition of some graphical or textual element present in the statement of the question. Consider only concrete difficulties whose manifestation can be evidenced by the information provided in generative AI's responses. At the end, discuss if these difficulties could be mitigated or eliminated by using prompt engineering techniques or if they would be inherent to the limitations of current large language models (LLMs).\\

	[question image]\\

	[ChatGPT-4 Vision's prompts and responses] \\

	[expert assessments]
\end{modelinputbox}
% \vfill
% \clearpage
\begin{modeloutputbox}[title={ChatGPT-4 Turbo's response},colbacktitle=gray!35!white]
	The generative AI, when attempting to answer the question on the Turing machine, encountered significant difficulties related to the interpretation of the Turing machine diagram and the verification of the proposed assertions. The main concrete difficulties identified were:\\

	1.~Interpretation and Analysis of Operations: The AI analyzed step by step the operations of the Turing machine, but made an incorrect interpretation of the machine's actions, especially regarding the substitution and movement of the reading head, which led to an erroneous conclusion about the content of the tape after four and eight moves.\\	 

	2.~Error in Evaluating Assertions: Based on its erroneous analysis, the AI initially claimed that assertions I and III were correct and assertion II was incorrect, contradicting the official key. After being confronted with the discrepancy, the AI revisited its analysis but continued to maintain an incorrect interpretation, even after revisiting the problem and receiving feedback to exclude the "B" symbols.\\	   

	3.~Feedback from Experts: Human experts highlighted that the AI's analysis was incorrect, specifically pointing out errors in the interpretation of the Turing machine diagram and in reasoning about the operations of the machine. This indicates that the AI failed to understand the Turing machine's logic as represented in the diagram properly.\\
	
	These difficulties reflect the current LLMs' limitations in correctly interpreting complex diagrams and performing logical analyses based on visual representations. Prompt engineering could potentially help the AI clarify ambiguities and better understand visual instructions, but it would not completely solve the problem of incorrect comprehension of a diagram, which is a visual representation that LLMs cannot directly process. This suggests that LLMs require significant advancements to better interpret visual representations and apply logical reasoning to complex technical problems.
\end{modeloutputbox}
% \vfill
% \clearpage
\begin{expertbox}[title={Challenge categories / error types}]
	\textbf{Logical Reasoning / Non-Compliance with Guidelines}\\[2pt] 
	The model failed to exclude the "B" symbols from the tape content when evaluating statements I and II, as instructed by the question.\\ 

	\textbf{Visual Acuity / Misidentification of Visual Elements}\\[2pt]
	The model failed to correctly interpret some labels in the Turing machine diagram, leading to an incorrect analysis of the machine's operations.\\

	\textbf{Logical Reasoning / Incorrect Algorithmic Reasoning}\\[2pt]
	Perhaps due to the previous two errors, the model made an incorrect interpretation of the machine's actions, especially regarding the substitution and movement of the reading head, leading to an erroneous conclusion about the content of the tape after the requested number of moves.
\end{expertbox}

\vfill
\clearpage

\subsubsection*{\myhref[Question 34]{34}}

This question asks the respondent to apply Dijkstra's shortest path algorithm to a given directed graph and then determine the cost estimate after two iterations of the algorithm starting from a given vertex. ChatGPT-4 Vision initially misidentified some weights of the edges in the provided graph, causing an erroneous calculation of the distances between some nodes and ultimately leading the model to conclude that no alternative was correct. After being challenged with the answer key and multiple clarification prompts, the model re-evaluated the graph's labels, correctly identified the weights of interest, and finally agreed with the answer key. This question demonstrates that ChatGPT-4 Vision can accurately apply complex algorithms to solve graph-based problems but may struggle to reach the correct solution on the first attempt due to the potential misidentification of important visual data. Moreover, the model's typical overconfidence in its initial response indicates a dependence on external stimuli to re-assess and ultimately change its conclusions.

\vspace{\parskip}
\begin{questionbox}[title={Question}]
	% \textbf{Question image} \\[2pt]
	\includegraphics[width=\imagewidth,valign=t]{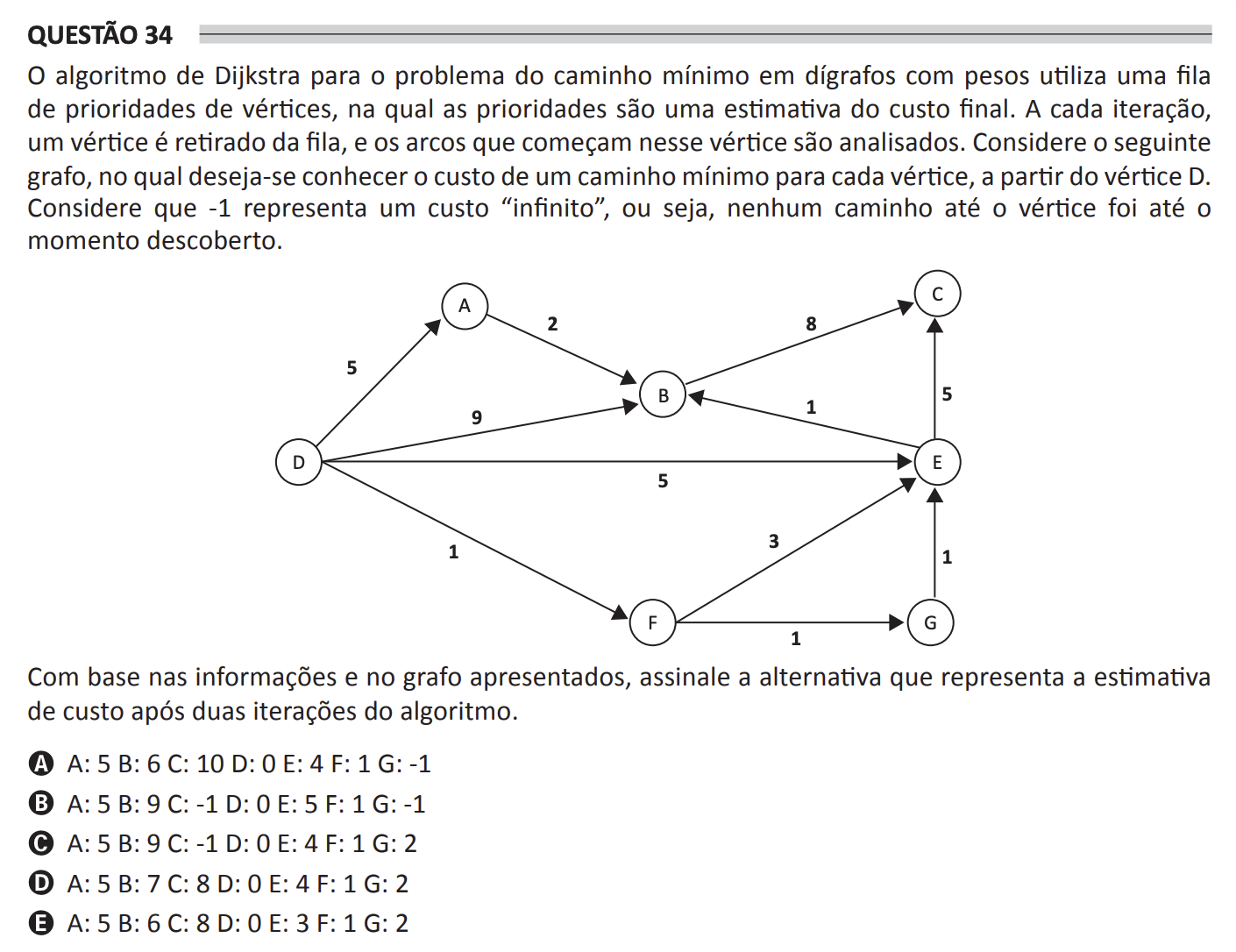}\\

	\vspace{8pt}
	\textbf{English transcription} \\[2pt]
	Dijkstra's algorithm for the shortest path problem in weighted digraphs uses a priority queue of vertices, where the priorities are an estimate of the final cost. At each iteration, a vertex is removed from the queue, and the arcs that start at that vertex are analyzed. Consider the following graph, in which the desire is to know the cost of the shortest path for each vertex, starting from vertex D. Assume that -1 represents an "infinite" cost, that is, no path to the vertex has been found up to that moment.\\

	[The image shows a weighted directed graph (digraph) with vertices labeled A to G and directed edges connecting them with assigned numerical weights.]\\
	
	Based on the information and the graph presented, select the alternative that represents the cost estimate after two iterations of the algorithm.\\
	
	A) A: 5 B: 6 C: 10 D: 0 E: 4 F: 1 G: -1\\
	B) A: 5 B: 9 C: -1 D: 0 E: 5 F: 1 G: -1\\	
	C) A: 5 B: 9 C: -1 D: 0 E: 4 F: 1 G: 2\\	
	D) A: 5 B: 7 C: 8 D: 0 E: 4 F: 1 G: 2\\	
	E) A: 5 B: 6 C: 8 D: 0 E: 3 F: 1 G: 2\\	

	\textbf{Answer key}: C
\end{questionbox}
% \begin{multicols}{2}
% \twocolumn
\begin{modelinputbox}[title={Initial prompt}]
	Answer the question in this image. Explain your reasoning step by step. At the end, choose which of the five alternatives (A-E) is correct. Inform if there is more than one correct alternative or if all five alternatives are incorrect.\\

	[question image]
\end{modelinputbox}
% \vfill
% \clearpage
\begin{modeloutputbox}[title={ChatGPT-4 Vision's initial response}]
	Let's apply Dijkstra's algorithm to the presented graph, considering D as the starting vertex. We need to calculate the cost estimates after two iterations of the algorithm.\\

	Initialization:\\
	D: 0 (since it is the starting point)\\
	A, B, C, E, F, G: -1 (indicating "infinity" or not yet discovered)\\
	
	First Iteration:\\
	We start with vertex D:\\
	D -> A = 5\\
	D -> B = 9\\  
	D -> E = 1\\  
	D -> F = 5\\ 
	
	At this point, the vertex with the smallest distance is E (1). Therefore, we select E for the next iteration.\\
	
	Update after the first iteration:\\
	D: 0 \\ 
	A: 5\\  
	B: 9 \\ 
	C: -1 \\ 
	E: 1 \\ 
	F: 5 \\ 
	G: -1 \\ 
	
	Second Iteration:\\
	From vertex E:\\  
	E -> C = 5 + 1 = 6 \\ 
	E -> G = 1 + 1 = 2 \\ 
	
	Update after the second iteration: \\
	D: 0  \\
	A: 5  \\
	B: 9  \\
	C: 6 (it's lower than the initial -1) \\ 
	E: 1  \\
	F: 5  \\
	G: 2 (it's lower than the initial -1) \\ 
	
	The cost estimates after two iterations are: \\ 
	A: 5 \\ 
	B: 9 \\ 
	C: 6 \\ 
	E: 1  \\
	F: 5  \\
	G: 2  \\
	
	Looking at the options, the alternative that corresponds to these estimates is:  \\
	E) A: 5 B: 6 C: 8 D: 0 E: 3 F: 1 G: 2\\
	
	However, this alternative seems to have errors, as B, C, E, and F do not match the values we calculated.\\
	
	Let's check the other alternatives:\\
	
	A) A: 5 B: 6 C: 10 D: 0 E: 4 F: 1 G: -1  \\
	B) A: 5 B: 9 C: -1 D: 0 E: 5 F: 1 G: -1 \\ 
	C) A: 5 B: 9 C: -1 D: 0 E: 4 F: 1 G: 2 \\ 
	D) A: 5 B: 7 C: 8 D: 0 E: 4 F: 1 G: 2 \\ 
	E) A: 5 B: 6 C: 8 D: 0 E: 3 F: 1 G: 2 \\ 
	
	Looking again at the alternatives, none of them exactly match the values we have calculated. Therefore, all five alternatives are incorrect.
\end{modeloutputbox}
\begin{modelinputbox}[title={Reassessment prompt}]
	Your response differs from the answer key released for this question: [answer key]\\
	
	After analyzing the answer key, do you maintain or change your response? If you decide to change your response, explain the possible reasons that led you to errors in the previous answer. Try to identify the concrete causes of the errors instead of listing generic reasons, such as a failure to understand the question statement.
\end{modelinputbox}
% \vfill
% \clearpage
\begin{modeloutputbox}[title={ChatGPT-4 Vision's revised response}]
	I maintain my original analysis. However, I apologize for the discrepancy with the answer key. The application of Dijkstra's algorithm that I presented is correct according to the traditional understanding of the algorithm and based on the information that was provided.\\

	Possible cause of the error in my previous response:\\	

	- Interpretation of the Statement: There may have been a different interpretation of what the question was asking. In some cases, the language of the statement or the way the problem is presented can create ambiguities.\\
	
	Dijkstra's algorithm is a standard method for finding the shortest paths in weighted graphs. The algorithm was applied correctly, and the estimated distances for each vertex after two iterations were correctly determined.\\
	
	It is possible that the question has some error or ambiguity that led to an incorrect answer key. I suggest that if there is any additional feedback or a detailed explanation of how the answer key arrived at the answer, this could help to clarify the discrepancy.	
\end{modeloutputbox}
% \vfill
% \clearpage
\begin{modelinputbox}[title={First clarification prompt}]
	Observing your first iteration, it appears you confused the weights of the edges leading to vertices E and F.
\end{modelinputbox}
\begin{modeloutputbox}[title={ChatGPT-4 Vision's first clarifying response}]
	Let's check it again:\\

	In the First Iteration, we start with vertex D:\\ 
	D -> A = 5 \\ 
	D -> B = 9  \\
	D -> E = 1 \\
	D -> F = 5 \\ 
	
	Indeed, looking at the graph again, I realize my mistake. Let's correct:\\
	
	D -> E = 5  \\
	D -> F = 3  \\
	
	Thus, the vertex with the smallest distance from D would be F, with a distance of 3. Therefore, we should select F for the next iteration, not E.\\
	
	Update after the first iteration:\\
	D: 0  \\
	A: 5  \\
	B: 9  \\
	C: -1  \\
	E: 5  \\
	F: 3  \\
	G: -1\\
	
	Second Iteration from vertex F:  \\
	F -> G = 3 + 1 = 4 \\ 
	
	*Update after the second iteration:
	D: 0 \\ 
	A: 5  \\
	B: 9  \\
	C: -1 \\ 
	E: 5  \\
	F: 3  \\
	G: 4  \\
	
	The cost estimates after two iterations are:\\  
	A: 5 \\
	B: 9 \\ 
	C: -1 \\ 
	E: 5 \\ 
	F: 3 \\ 
	G: 4 \\ 
	
	Looking at the options, the alternative that corresponds to these estimates is:\\  
	C) A: 5 B: 9 C: -1 D: 0 E: 4 F: 1 G: 2\\
	
	Thank you for pointing out the error. I apologize for the previous confusion and appreciate your patience.	
\end{modeloutputbox}
\begin{modelinputbox}[title={Second clarification prompt}]
	Ok, your answer may be correct, but you still got the weight of the edge D -> F wrong. I suggest re-examining the image.
\end{modelinputbox}
% \vfill
% \clearpage
\begin{modeloutputbox}[title={ChatGPT-4 Vision's second clarifying response}]
	I apologize for the previous error. Let's look at it again:\\

	In the First Iteration, starting from vertex D, we have:\\
	
	D -> A = 5  \\
	D -> B = 9  \\
	D -> E = 5  \\
	D -> F = 1 \textbf{(It was my previous error, and you are correct.)}\\
	
	At this point, the vertex with the smallest distance from D is F, at a distance of 1. Therefore, we select F for the next iteration.\\
	
	Update after the first iteration:\\
	D: 0  \\
	A: 5  \\
	B: 9  \\
	C: -1  \\
	E: 5  \\
	F: 1  \\
	G: -1 \\ 
	
	Second Iteration from vertex F: \\ 
	F -> G = 1 + 1 = 2 \\ 
	
	Update after the second iteration:\\
	D: 0 \\ 
	A: 5  \\
	B: 9 \\ 
	C: -1 \\ 
	E: 5 \\ 
	F: 1 \\ 
	G: 2 \\ 
	
	The cost estimates after two iterations are:\\  
	A: 5  \\
	B: 9  \\
	C: -1 \\ 
	E: 5 \\ 
	F: 1 \\ 
	G: 2 \\ 
	
	Looking at the options, the alternative that corresponds to these estimates is:  \\
	C) A: 5 B: 9 C: -1 D: 0 E: 4 F: 1 G: 2\\
	
	Thank you for pointing out the error. I apologize for the previous confusion and appreciate your patience.	
\end{modeloutputbox}
\begin{modelinputbox}[title={Reflectice prompt to ChatGPT-4 Turbo}]
	The following are the prompts and responses of a generative AI for the attached question, along with possible assessments by two or more human experts on the generative AI's responses and the general quality of the question. Based on the provided information, summarize the main difficulties faced by the generative AI in attempting to answer the question, including possible difficulties resulting from incorrect recognition of some graphical or textual element present in the statement of the question. Consider only concrete difficulties whose manifestation can be evidenced by the information provided in generative AI's responses. At the end, discuss if these difficulties could be mitigated or eliminated by using prompt engineering techniques or if they would be inherent to the limitations of current large language models (LLMs).\\

	[question image]\\

	[ChatGPT-4 Vision's prompts and responses] \\

	[expert assessments]
\end{modelinputbox}
\begin{modeloutputbox}[title={ChatGPT-4 Turbo's response},colbacktitle=gray!35!white]
	The difficulties faced by the generative AI when answering the question about Dijkstra's algorithm are evident in the analysis of the provided file:\\

	1. Error in Interpretation of the Image: The AI made a mistake in interpreting the weights of the edges in the provided graph. Initially, it incorrectly calculated the distances from D to E and F, leading to an erroneous response.\\
	
	2. Correction After Feedback: After being alerted to a possible discrepancy, the AI re-examined the graph and corrected the error, changing the distances from D to E and F, and consequently, the final response.\\
	
	3. Conflict Between Initial Analysis and Answer Key: Initially, the AI identified that none of the provided alternatives corresponded to its calculations. After revision and correction of the edge weights, the AI aligned its response with the official answer key.\\
	
	4. Confirmation with Answer Key: Even after adjusting the interpretation of the edge weights, the AI maintained its response until an answer key was provided, indicating a dependence on external confirmation to validate its conclusions.\\
	
	These difficulties could be partially mitigated by prompt engineering techniques, such as clarifying the interpretation of graphic elements or reinforcing the need to double-check responses with the visual information provided. However, accuracy in interpreting graphs and solving problems based on graphics are challenges inherent to the current LLMs. While prompt engineering can help guide the AI towards a more accurate interpretation, the fundamental limitations of LLMs in understanding and processing complex visual information still require advancements in integrating visual capabilities with natural language processing.	
\end{modeloutputbox}
\begin{expertbox}[title={Challenge categories / error types}]
	\textbf{Visual Acuity / Misidentification of Visual Elements}\\[2pt]
	The model failed to correctly identify some weights of the edges in the provided graph, leading to an erroneous calculation of the distances between some nodes and, consequently, an incorrect response.
\end{expertbox}

\end{document}